\documentclass{article}

\usepackage{PRIMEarxiv}
\usepackage{color}
\usepackage[utf8]{inputenc}
\usepackage[T1]{fontenc}
\usepackage{hyperref}
\usepackage{url}
\usepackage{booktabs}
\usepackage{amsfonts}
\usepackage{nicefrac}
\usepackage{microtype}
\usepackage{fancyhdr}
\usepackage{graphicx}
\graphicspath{{figs/}}
\usepackage{amsmath}
\usepackage{amssymb}
\usepackage{subfigure}
\usepackage[ruled,vlined,linesnumbered]{algorithm2e}
\usepackage{natbib}

\newcommand{\sd}{\sigma_d}
\newcommand{\xt}{\mathbf{z}_t}
\newcommand{\xz}{\hat{\mathbf{z}}_0}
\newcommand{\Dec}{\mathcal{D}}

\title{Multimodal spatiotemporal atmospheric data assimilation with latent video flow-matching}

\author{
  Dibyajyoti Chakraborty \\
  Information Sciences and Technology \\
  The Pennsylvania State University \\
  University Park, Pennsylvania\\
  \textit{d.chakraborty@psu.edu} \\
  \And
  Romit Maulik \\
  School of Mechanical Engineering \\
  Purdue University \\
  West Lafayette, Indiana\\
  \textit{rmaulik@purdue.edu} \\
}

\newcommand{\cmark}{\checkmark}

\begin{document}
\maketitle

\begin{abstract}
Data assimilation (DA) uses Bayesian inference to update the state of a numerical forecast model with observed data. In this study, we propose a fundamentally different, unified approach to atmospheric data assimilation. We use latent video flow-matching to sample temporally consistent trajectories from a prior trained using ERA5 reanalysis (69 variables over an 8-day window). We also use posterior sampling to assimilate real observation sources, such as those from the NOAA Integrated Global Radiosonde Archive and the Integrated Surface Database. Because the prior generates a continuous trajectory, it naturally propagates information between observed and unobserved frames. Therefore, we can perform various DA tasks, such as filtering and smoothing, simply by changing the observed frames. Moreover, we generate full-state ensemble forecasts directly from sparse observations, achieving performance competitive with state-of-the-art observation-to-forecast models.

\end{abstract}


\section{Introduction}

Earth's atmospheric dynamics span a wide range of spatial and temporal scales
\cite{arakawa2011toward,neugebauer2003dynamics,tao2009multiscale}. Forecasting such a system is difficult because it can only be simulated approximately and requires accurate initial conditions. However, atmospheric states can only be partially observed using specialized instruments \cite{thepaut2003satellite,bouttier2001observing}. This creates a need for data and model fusion, in which first-principles numerical models provide a prior (background) state that is updated given partial observations through a Bayesian formulation \cite{wikle2007bayesian,law2012evaluating}. This update, known as data assimilation (DA), is central to modern weather prediction and underpins reanalysis products such as ERA5 \cite{hersbach2020era5}. However, these classical variational and ensemble methods \cite{szunyogh2014applicable,wang2000data,reichle2008data} are computationally expensive and require several iterations of expensive numerical models. 

Generative machine learning \cite{salakhutdinov2015learning,gao2024generative} offers an alternative \cite{price2025probabilistic}. Instead of re-running a numerical model, one learns an implicit probability density of the atmospheric state from which new samples can be drawn, and conditions that density in real time on sparse observations. Diffusion models \cite{song2021maximum,song2020score,karras2022elucidating} learn such densities by corrupting data with noise through a forward stochastic process and learning the score (the gradient of the log-probability density), so that the reverse process can iteratively \textit{denoise} pure noise into samples from the data distribution. Flow-matching \cite{lipman2022flow} is an adjacent method that learns a velocity field, which is used to directly interpolate from noise to the data distribution. Recent advances in the computer vision community have shown that flow-matching techniques have an edge over previous methods in high-dimensional generative tasks \cite{esser2024scaling}. A trained generative model also enables \emph{zero-shot} conditioning: the pretrained unconditional prior can be guided by sparse observations without retraining \cite{song2020score,graikos2022diffusion,chung2022diffusion}. This is effectively equivalent to sampling from the posterior distribution.

Previous work has demonstrated this for ERA5 snapshots, combining coarse-grid fields, NOAA Integrated Global Radiosonde Archive (IGRA) data, and a climate emulator for super-resolution and multimodal reconstruction \cite{chakraborty2026multimodal}. There, temporal consistency came only from conditioning on the output of a separate emulator. In this work, temporal
propagation is internal to the prior itself. This has two consequences that are absent in a single snapshot generation setting. First, the prior inherently propagates information from observed frames to unobserved frames, removing the need for a separate emulator or numerical forecast model. Second, different classical DA regimes are accurately recovered simply by choosing which frames (or data) are observed. The frames here refer to the individual timesteps in the video. We summarize our contributions as follows.

\begin{enumerate}
  \item We train an unconditional latent \emph{video} flow-matching prior (TrigFlow \cite{lu2025simplifying} with a 3D diffusion-transformer backbone \cite{peebles2023scalable}) that generates a 32-frame (8-day, 6-hourly) window of the global atmosphere for 69 ERA5 variables. This serves as a background model for a wide range of data-assimilation experiments without any retraining or observation-specific conditioning.

  \item On the methods side, we adopt an adaptive noise-level weighting for the TrigFlow loss. We also explore a broad set of guided-sampler designs, utilizing low-noise-level correctors and gradient (and noise) rescaling techniques for both efficiency and accuracy.

  \item On the experimental side, we recover the full atmospheric state from both structured coarse observations and real observations from the NOAA
  Integrated Global Radiosonde Archive (IGRA) \cite{durre2006overview}, the NOAA
  Integrated Surface Database (ISD) \cite{smith2011integrated}, and the International
  Comprehensive Ocean-Atmosphere Data Set (ICOADS) \cite{freeman2017icoads}. We use these to realize the three classical DA regimes (filtering, smoothing, and fixed-interval reconstruction) from the same prior by changing only which frames are observed. We further use the prior for direct observation-to-forecast, producing a forecast competitive with some prior observation-to-forecast models (GraphDOP \cite{alexe2024graphdop}). Finally, we study its behavior in a particular case, Hurricane Laura (2020).
\end{enumerate}

Our results indicate that latent video flow-matching priors can perform accurate multimodal spatiotemporal DA from multiple real data sources, with uncertainty quantification through the spread of posterior draws. An
end-to-end comparison with operational assimilation is beyond our scope. The codes used in the work are in \hyperlink{https://github.com/ISCLPurdue/era5_videogen}{https://github.com/ISCLPurdue/era5\_videogen}.

\subsection{Related work}

Deep learning has produced strong results in reconstructing the state of high-dimensional multiscale dynamical systems
\cite{fukami2019super,fathi2020super,kelshaw2022physics,fukami2023super,gao2021super,ren2023physr,fukami2024single}. Specifically, Graph Neural Networks (GNNs) or
Voronoi-tessellation-assisted methods have been used for reconstruction from unstructured sensor measurements \cite{barwey2025mesh,fukami2021global}. Most of these demonstrations, however,
rely on coarse grids extracted from the fine state, so inputs and targets come
from the same dataset. Additionally, they require
retraining when the observation source changes and, being deterministic, lack
uncertainty quantification (UQ). Probabilistic networks \cite{maulik2020probabilistic},
deep ensembles \cite{maulik2023quantifying}, and stochastic weight averaging
\cite{izmailov2018averaging,morimoto2022assessments} provide UQ at the cost of restrictive assumptions or increased computation, which motivates generative priors.

Score-based diffusion models are now the dominant generative models for physical
systems, beginning with diffusion probabilistic models
\cite{sohl2015deep,ho2020denoising} and score-based generative models
\cite{song2019generative} (building on score matching \cite{hyvarinen2005estimation}),
unified through stochastic differential equations \cite{song2020score} and refined
by the EDM design space exploration \cite{karras2022elucidating}. Flow-matching \cite{lipman2022flow,liu2022flow,albergo2025stochastic} uses a different construction, prescribing an interpolation path from noise to the data distribution directly rather than reversing a noising process. These priors solve inverse problems in a zero-shot, plug-and-play manner through gradient guidance (DPS
\cite{chung2022diffusion}, $\Pi$GDM \cite{song2023pseudoinverse}), range-null-space methods
(DDRM \cite{kawar2022denoising}, DDNM \cite{wang2022zero}), variational and
variable-splitting solvers (RED-diff \cite{mardani2024variational}, DiffPIR
\cite{zhu2023denoising}, PnP-DM \cite{wu2024principled}, DAPS \cite{zhang2025improving}), and
latent-space variants \cite{rout2023solving,song2024solving}. The \textsc{InverseBench} benchmark \cite{zheng2025inversebench} compares many of these. We also re-implement some of these methods on our prior in App.~\ref{app:lit}. In atmospheric applications, they have been applied for downscaling and super-resolution
\cite{mardani2025residual,srivastava2024precipitation,watt2024generative},
probabilistic forecasting \cite{li2024generative,price2025probabilistic,andrae2025continuous}, and data assimilation
\cite{huang2024diffda,manshausen2025generative,yang2025generative,qu2024deep,wang2025phyda,bao2024score,liang2025ensemble}. Similar to us, Qu et al.~\cite{qu2024deep} target a multimodal setting, and Liang et al.~\cite{liang2025ensemble} and Bao et al.~\cite{bao2024score} pursue score-based filtering, but on single-time or lower-dimensional states rather than a latent video trajectory of the
full ERA5 state. SEEDS \cite{li2024generative} generates skillful forecast ensembles
cheaply but emulates an ensemble rather than assimilating observations.
Manshausen et al.~\cite{manshausen2025generative} assimilate sparse station data from a single source
on a single snapshot, while we assimilate several independently sourced observations over
an 8-day trajectory and infer the full vertical column of the atmosphere. 

Spatiotemporal diffusion priors for scientific video inverse problems
\cite{zhang2025step} and Appa Weather \cite{andry2025appa} are the
closest to our work, and we differ from each as follows. STeP
\cite{zhang2025step} performs spatiotemporal latent posterior sampling but is
evaluated on simulated observations. Similarly, Appa uses a latent diffusion prior and synthetic observations for global data assimilation and forecasting. Here, we use real observations to perform various data-assimilation tasks and show over three weeks of competitive forecast performance directly from observations.

\section{Methods}
\label{sec:methods}

\begin{figure}[t]
\centering
\includegraphics[width=0.95\linewidth]{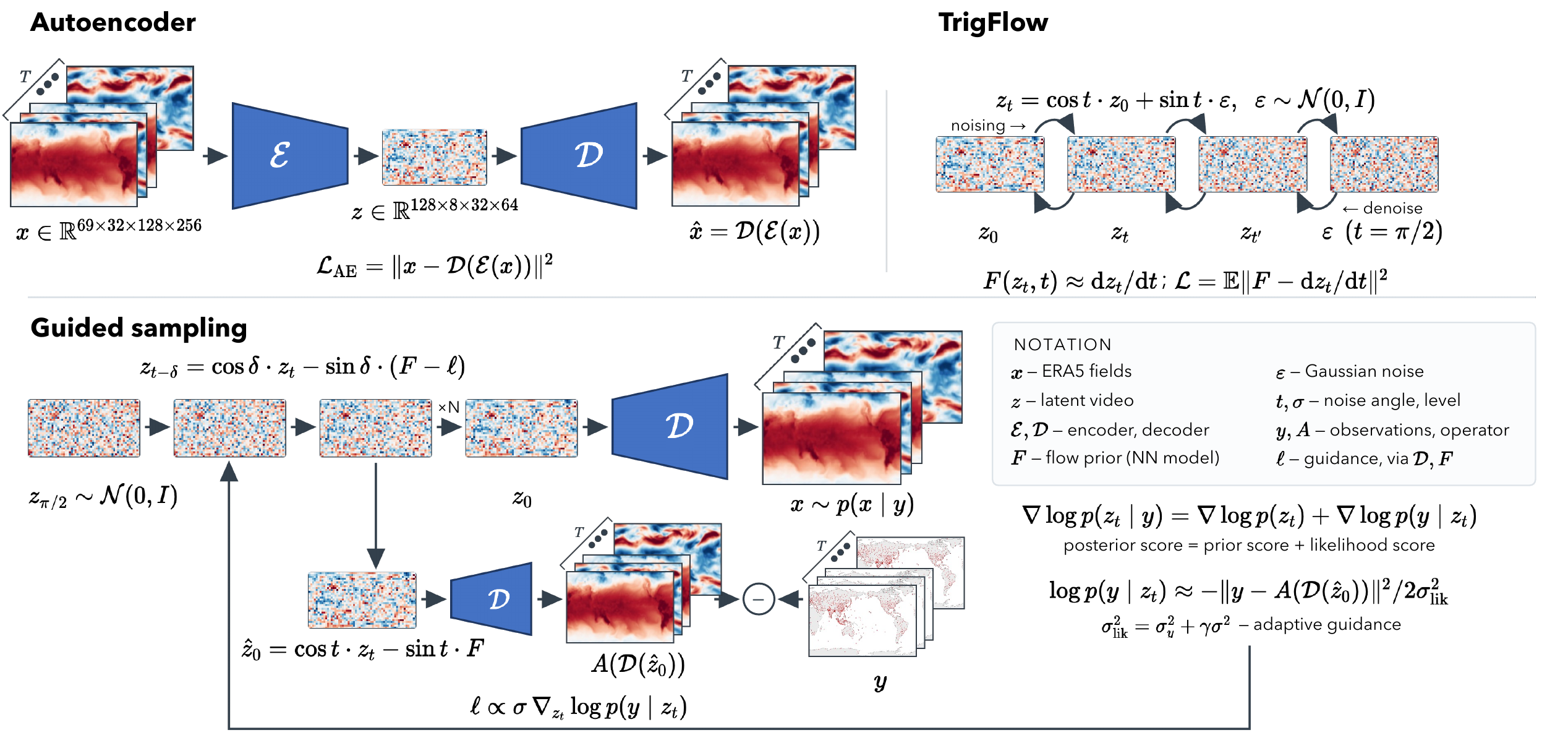}
\caption{Overview of the proposed method. An autoencoder
$(\mathcal{E},\mathcal{D})$ compresses 8-day ($32$ timesteps) ERA5 windows (69 variables at $128{\times}256$ grid) into a latent video
$\mathbf{z}\in\mathbb{R}^{128\times8\times32\times64}$. The TrigFlow prior $F$
is trained by velocity matching along the spherical noising path
$\mathbf{z}_t=\cos t\,\mathbf{z}_0+\sin t\,\boldsymbol{\varepsilon}$. From approximately pure
noise, $N$ guided steps sample the posterior: each step decodes the latent Tweedie estimate $\hat{\mathbf{z}}_0$, scores it against the observations
$\mathbf{y}$ through the measurement operator $A$, and injects the guidance term $\boldsymbol{\ell}$ into the update, with the annealed likelihood variance $\sigma_{\text{lik}}^2$ weakening guidance at high noise. $\mathcal{D}$ and $F$ stay frozen while sampling, so a single pretrained prior serves every assimilation
task in this paper zero-shot. Note: The latent video looks like Gaussian noise visually; however, it has some small structures that the decoder can decode successfully.}
\label{fig:schematic}
\end{figure}

A high-level view of the full methodology is given in the schematic of Fig.~\ref{fig:schematic}: an autoencoder compresses spatio-temporal ERA5 data into a latent video, a TrigFlow prior is trained on those latents, and at sampling time the learned prior is combined with an observation likelihood to draw posterior samples. The subsections below describe each ingredient in turn.

\subsection{Autoencoder.}
We train a 3D-convolutional autoencoder (AE) $(\mathcal{E},\Dec)$ to compress the ERA5 data $\mathbf{x}$ to a latent representation $\mathbf{z}$ for faster training and sampling. The decoder $\Dec$ maps the latent back to the 69 standardized ERA5 variables at $T{=}32,H{=}128,$ and $W{=}256$. The autoencoder has a $4\times$ temporal and $4\times$ spatial compression (each direction) for the latent grid. The generative prior is trained on the latents of the autoencoder (similar to a video) of shape $(C{=}128,T{=}8,H{=}32,W{=}64)$. Because likelihoods are evaluated in the physical field space, guidance passes through the decoder as in latent-diffusion inverse solvers \cite{rout2023solving,song2024solving}.

\subsection{Generative modeling}
\label{sec:genmod}

\paragraph{Flow-matching.}
Flow-matching \cite{lipman2022flow,albergo2025stochastic} prescribes an interpolation path $\mathbf{x}_t=\alpha_t\,\mathbf{x}_0+\beta_t\,\boldsymbol{\varepsilon}$ between data and Gaussian noise. The network is trained to regress the velocity $d\mathbf{x}_t/dt$ along that path. During sampling, new samples are generated by simulating the ordinary differential equation (ODE) $d\mathbf{x}_t = v_\theta(\mathbf{x}_t, t)dt$ from $t=T$ (pure noise) to $t=0$. Various flow models correspond to particular choices of schedule $(\alpha_t,\beta_t)$.

\paragraph{TrigFlow.}
The TrigFlow parameterization \cite{lu2025simplifying} is a flow-matching formulation whose interpolation between data and noise is spherical. Our prior is trained not on the ERA5 fields ($\mathbf{x}$) themselves but on the latent $\mathbf{z}$ of an autoencoder. The forward interpolation path for the noise is defined as:
\begin{equation}
\xt=\cos(t)\,\mathbf{z}_0+\sin(t)\,\boldsymbol{\varepsilon},\quad
\boldsymbol{\varepsilon}\sim\mathcal N(0,\sd^2 I),\;\; t\in(0,\tfrac\pi2).
\label{eq:forward}
\end{equation}

We fix $\sd{=}1$ throughout. Differentiating Eq.~\eqref{eq:forward} with respect to time gives the target velocity $d\xt/dt=-\sin(t)\,\mathbf{z}_0+\cos(t)\,\boldsymbol{\varepsilon}$. A neural network $F(\xt/\sd,t)$, here a 3D diffusion transformer (DiT3D \cite{peebles2023scalable}), is trained so that $\sd\,F(\xt/\sd,t)\approx d\xt/dt$. At the optimum of the training loss, the learned velocity yields an estimate of the clean latent, which coincides with the Tweedie (minimum mean squared error) estimate \cite{efron2011tweedie},
\begin{equation}
\xz=\cos(t)\,\xt-\sin(t)\,\sd\,F(\xt/\sd,\,t).
\label{eq:tweedie}
\end{equation}
Sampling integrates the ODE $d\mathbf{x}_t = v_\theta(\mathbf{x}_t, t)dt$ from $t_{\max}=\arctan(\sigma_{\max}/\sd)$ (approximately pure noise) to $t=0$ (data).

\paragraph{Adaptive noise-level weighting.}
The baseline flow model is trained to match the velocity by minimizing the expected squared error:
\begin{equation}
\mathcal L(\theta)=\mathbb E_{\mathbf{z}_0,\boldsymbol{\varepsilon},t}
\left[\big\|\sd\,F(\xt/\sd,\,t;\theta)-\tfrac{d\xt}{dt}\big\|^2\right].
\label{eq:vloss}
\end{equation}
The velocity target contains the noise $\boldsymbol{\varepsilon}$, so the regression has a $t$-dependent irreducible
variance. Since no network can predict the specific noise component, the loss floor is disproportionately large at certain noise levels. Under a fixed weighting, high noise levels dominate the gradient, which hampers learning. We therefore utilize an adaptive uncertainty weighting \cite{karras2024analyzing}, where the network additionally predicts a scalar log-variance $u(t)$ through a small linear head on its noise-level embedding. The loss is formulated as the maximum-likelihood objective for a Gaussian error model with level-dependent variance:
\begin{equation}
\mathcal L(\theta)=\mathbb E\big[\,e^{-u(t)}
\big\|\sd\,F(\xt/\sd,\,t;\theta)-\tfrac{d\xt}{dt}\big\|^2+u(t)\,\big].
\label{eq:adaptive}
\end{equation}
At the optimum, this objective sets $e^{u(t)}$ to the expected squared error at level $t$, ensuring that every noise level contributes gradients of comparable magnitude regardless of its floor. Training noise levels are drawn from a log-normal distribution centered on the data scale, $\ln\sigma\sim\mathcal N(0,1.5^2)$ with $\sigma=\tan t$. The $u$ head is
discarded at inference, so sampling is unaffected.

\paragraph{Posterior sampling and data fusion.}\label{sec:dps}
To sample the posterior $p(\mathbf{x}\mid\mathbf{y})$ we draw latents from $p(\mathbf{z}\mid\mathbf{y})$ and decode them, utilizing Bayes' theorem to combine the pretrained unconditional prior with a measurement likelihood. The posterior score is the sum of the prior and likelihood scores,
\begin{equation}
\nabla_{\xt}\log p(\xt\mid\mathbf{y})=
\nabla_{\xt}\log p(\xt)+\nabla_{\xt}\log p(\mathbf{y}\mid\xt),
\label{eq:posterior_score}
\end{equation}
structurally analogous to classifier guidance \cite{ho2022classifier}. For a differentiable measurement operator $A$ and Gaussian observation noise with standard deviation $\sigma_y$, DPS \cite{chung2022diffusion} evaluates the likelihood at the Tweedie estimate $\xz$,
\begin{equation}
\log p(\mathbf{y}\mid\xt)\approx
-\tfrac{1}{2\,\sigma_{\text{lik}}^2(\sigma)}\big\|\mathbf{y}-A(\Dec(\xz))\big\|^2,
\label{eq:lik}
\end{equation}
and score-based data assimilation (SDA) \cite{rozet2023score} accounts for the variance of $\xz$ through an annealed likelihood variance,
\begin{equation}
\sigma_{\text{lik}}^2(\sigma)=\sigma_y^2+\gamma\,(\sigma/\sd)^2 ,
\label{eq:sda}
\end{equation}
where $\gamma$ is the SDA annealing coefficient and $\sd$ is the latent data scale
(fixed to $1$ here, so $\sigma/\sd{=}\sigma$ is the running noise level while sampling). This is used to automatically weaken guidance at high noise levels.


\subsection{Guided TrigFlow sampler}\label{sec:sampler}
All sampling methods in this article are built on our guided TrigFlow sampler (Algorithm~\ref{alg:sampler}), which combines three base techniques. First, it uses the Elucidating Diffusion Model (EDM) noise schedule \cite{karras2022elucidating}, with each noise level mapped to a TrigFlow angle through $t=\arctan(\sigma/\sd)$. Second, it advances the state with the DPM-Solver++ update \cite{lu2025dpm}, which in TrigFlow coordinates is a rotation (the second-order coefficient is derived in Appendix~\ref{sec:math}e). Third, at every step it adds DPS-style measurement guidance with SDA annealing. 

The sampler has an option to follow a predictor-corrector structure \cite{song2020score}. The guided rotation acts as the \emph{predictor}: it transports the sample from one noise level to the next, with the guidance term bending the trajectory toward measurement consistency. However, since the error of this predictor step is carried to the next level, a \emph{Langevin corrector} (Algorithm~\ref{alg:effe3}) holds the noise level fixed and takes one Langevin step toward the intermediate posterior $p(\xt\mid\mathbf{y})\propto p(\xt)\,p(\mathbf{y}\mid\xt)$, using a fresh denoiser evaluation and a fresh likelihood gradient. Since each corrector step costs one extra denoiser evaluation, we apply it only below a threshold $\sigma_{\max}^c$, where the Tweedie estimate is sharp and measurement fidelity is actually established (Sec.~\ref{sec:design-choices}).

\begin{algorithm}[!ht]
\caption{Guided TrigFlow sampler. Tunable options and their
defaults are described in Sec.~\ref{sec:sampler}.}
\label{alg:sampler}
\SetKwInOut{Input}{input}\SetKwInOut{Output}{output}
\Input{measurement $\mathbf{y}$, steps $N$, $\sd{=}1$, \textit{scale}; angles $t_0{>}\dots{>}t_N{=}0$ at $\sigma$ levels, $t_k=\arctan(\sigma_k/\sd)$; $\mathrm{clip}(\cdot)$ is the elementwise clip to $[-1,1]$ (used in all runs)}
$\xt \leftarrow \sd\cdot\varepsilon,\ \varepsilon\sim\mathcal N(0,I)$;\ \ $\hat{\mathbf{z}}_0^{\text{prev}}\leftarrow\varnothing$;\ \ $\boldsymbol\ell^{\text{m}}\leftarrow\mathbf{0}$\;
\For{$k=0,\dots,N-1$}{
  $s\leftarrow t_k$,\ $t\leftarrow t_{k+1}$,\ $\delta\leftarrow s-t$,\ $\sigma\leftarrow\sd\tan s$\;
  $F_s \leftarrow F(\xt/\sd,\,s)$\tcp*{denoiser velocity}
  $\xz \leftarrow \cos(s)\,\xt-\sin(s)\,\sd\,F_s$\tcp*{Tweedie estimate, Eq.~\eqref{eq:tweedie}}
  \uIf{$\sigma_{\min}^g\le\sigma\le\sigma_{\max}^g$}{
    $\mathbf{g} \leftarrow \nabla_{\xt}\!\big[{-}\tfrac{1}{2\sigma_{\text{lik}}^2}\|\mathbf{y}-A(\Dec(\xz))\|^2\big]$\tcp*{backprop through $\Dec$ \emph{and} $F$}
    $\boldsymbol\ell \leftarrow \mathrm{scale}\cdot\sigma\cdot \mathrm{clip}(\mathbf{g})$\ \emph{or}\ $\mathrm{scale}\cdot\sigma\cdot\mathrm{clip}(\sqrt{n}\,\mathbf{g}/\|\mathbf{g}\|_2)$ under DSG\tcp*{$n{=}CTHW$}
    $\boldsymbol\ell\leftarrow\boldsymbol\ell+m\,\boldsymbol\ell^{\text{m}}$;\ \ $\boldsymbol\ell^{\text{m}}\leftarrow\boldsymbol\ell$\tcp*{momentum on the scaled score}
  }\Else{$\boldsymbol\ell\leftarrow \mathbf{0}$\tcp*{guidance band gate}}
  $\mathbf{z}^{(1)}\leftarrow\cos(\delta)\,\xt-\sin(\delta)\,\sd\,F_s$\tcp*{1st-order rotation}
  \uIf{$0<k<N-1$ \textbf{and} $\hat{\mathbf{z}}_0^{\text{prev}}\neq\varnothing$}{
    $r\leftarrow(\log\tan s-\log\tan t_{\text{prev}})/(\log\tan s-\log\tan t)$;\\
    $\mathbf{z}_{\text{next}}\leftarrow\mathbf{z}^{(1)}+\tfrac{\sin\delta}{2\,r\,\sin s}\,(\hat{\mathbf{z}}_0^{\text{prev}}-\xz)$\tcp*{2nd-order}
  }\Else{$\mathbf{z}_{\text{next}}\leftarrow\mathbf{z}^{(1)}$}
  $\mathbf{z}_{\text{next}}\leftarrow \mathbf{z}_{\text{next}}+\sin(\delta)\,\sd\,\boldsymbol\ell$\tcp*{inject guidance, $F\!\to\!F-\boldsymbol\ell$}
  $\hat{\mathbf{z}}_0^{\text{prev}}\leftarrow\xz$\;
  \If{corrector \textbf{and} $k<N-1$ \textbf{and} $\sd\tan t\le\sigma_{\max}^c$}{
     $\mathbf{z}_{\text{next}}\leftarrow\textsc{LangevinCorrector}(\mathbf{z}_{\text{next}},t)$\tcp*{Alg.~\ref{alg:effe3}}\ \
     $\hat{\mathbf{z}}_0^{\text{prev}}\leftarrow\varnothing$\tcp*{2nd-order unused for next step}
  }
  $t_{\text{prev}}\leftarrow s$;\ \ $\xt\leftarrow \mathbf{z}_{\text{next}}$\;
}
\Output{
$\xt$
}
\end{algorithm}

\begin{algorithm}[t]
\caption{$\textsc{LangevinCorrector}(\mathbf{z},s)$: one SDA-style Langevin step at angle $s$ toward the posterior score.}
\label{alg:effe3}
\SetKwInOut{Output}{output}
$F\leftarrow F(\mathbf{z}/\sd,s)$;\ \ $\xz\leftarrow\cos s\,\mathbf{z}-\sin s\,\sd\,F$\tcp*{$\sigma{=}\sd\tan s$}
$\mathbf{g}\leftarrow\nabla_{\mathbf{z}}\!\big[{-}\tfrac1{2\sigma_{\text{lik}}^2}\|\mathbf{y}-A(\Dec(\xz))\|^2\big]$;\ \ $\boldsymbol\ell\leftarrow\mathrm{scale}\cdot\sigma\,\mathrm{clip}(\mathbf{g})$\tcp*{clip/DSG as in Alg.~\ref{alg:sampler}; no momentum}
$\mathbf{s}_{\text{pri}}\leftarrow(\cos s\,\xz-\mathbf{z})/(\sin^2 s\,\sd^2)$\tcp*{prior score}
$\mathbf{s}_{\text{lik}}\leftarrow \boldsymbol\ell/(\sd\tan s)$\tcp*{likelihood score}
$\eta\leftarrow(\mathrm{snr}\cdot\sin s\,\sd)^2$\;
$\mathbf{z}\leftarrow \mathbf{z}+\eta(\mathbf{s}_{\text{pri}}{+}\mathbf{s}_{\text{lik}})+\lambda\,\sqrt{2\eta}\,\varepsilon,\ \varepsilon\sim\mathcal N(0,I)$\tcp*{$\lambda{=}1$: sample; $\lambda{\to}0$: mode-seeking}
\Output{$\mathbf{z}$}
\end{algorithm}

\subsubsection{Design choices}\label{sec:design-choices}
We have the following tunable options in our sampler.
\begin{itemize}
  \item \textit{scale}: guidance scale (the strength of the likelihood gradient);
  \item $[\sigma_{\min}^g,\sigma_{\max}^g]$: guidance band, outside which no
        guidance is applied;
  \item $m$: momentum decay on the guidance term;
  \item DSG: normalization of the guidance gradient;
  \item $\lambda$, $\sigma_{\max}^c$: noise scale and activation threshold of
        the Langevin corrector.
\end{itemize}

We highlight the following designs which are combinations of the above choices. Table~\ref{tab:variants} mentions them with their design choice settings and Number of Function Evaluations (NFE). Extended reasoning behind these designs is given in Appendix~\ref{sec:des-choices}. All the hyperparameters have been tuned using the ablation studies in Appendix~\ref{sec:samplers}. The DPS mentioned in the rest of the paper is DPS with the annealed likelihood variance of Eq. \ref{eq:sda}.

\paragraph{Low noise level correction.}
Every solver step and every corrector step costs one denoiser forward pass, plus one backward pass when guidance is active. At high $\sigma$ the Tweedie estimate $\xz$ is blurry, and SDA annealing further shrinks the
likelihood gradient, so guidance accomplishes little there. Measurement fidelity is only established near $t{\to}0$. The reduced-step corrector configurations (Table~\ref{tab:variants}) therefore reallocate the NFE budget. The Langevin corrector steps are restricted to the low-$\sigma$ band (Algorithm~\ref{alg:effe3}). The configurations we use further set the step count $N$ and corrector gate
$\sigma_{\max}^c$ to
\begin{itemize}
   \item \textit{DPS+corr\,(N25)}: $(N,\sigma_{\max}^c)=(25,3)$;
\item \textit{DPS+corr\,(N30)}: $(N,\sigma_{\max}^c)=(30,5)$.
\end{itemize}

\paragraph{Reshaping gradients and noise.}
We modify how the sampler uses the gradients and noise in the following ways.
\begin{itemize}
  \item \emph{$\lambda$ noise-scaling} \cite{wang2025traversing} multiplies the corrector's injected noise, $\lambda\sqrt{2\eta}\,\varepsilon$ in Algorithm~\ref{alg:effe3}, by a factor $\lambda\in[0,1]$. This interpolates between posterior samples at $\lambda{=}1$, the default, and a deterministic mode-seeking trajectory at $\lambda{=}0$, used by the mode-seeking configurations of Table~\ref{tab:variants}.
  \item \emph{Momentum guidance} accumulates the scaled likelihood score across steps, $\boldsymbol\ell_t=\mathrm{scale}\cdot\sigma_t\,\mathbf{g}_t+m\,\boldsymbol\ell_{t-1}$, with decay $m\in[0.3,0.5]$ (Algorithm~\ref{alg:sampler}). Unlike lazy gradient reuse \cite{castillo2025adaptive}, the gradient is still recomputed at every step; momentum only averages successive directions. The accumulator starts at $\boldsymbol\ell_{-1}{=}\mathbf 0$, so the first step is plain DPS. One variant uses $m{=}0.5$; the other momentum variants use $m{=}0.3$ (Table~\ref{tab:variants}).
  \item \emph{DSG} (diffusion with spherical-Gaussian constraint)
        \cite{yang2024guidance} unit-normalizes the guidance gradient per sample and rescales it to magnitude
        $\sqrt{n}$, where $n{=}CTHW$ is the number of latent elements (here $n{=}128{\cdot}8{\cdot}32{\cdot}64{=}2{,}097{,}152$),
        before the elementwise clip and the guidance scale are applied as
        in Algorithm~\ref{alg:sampler}. This fixes the guidance step size
        regardless of the raw gradient magnitude.
\end{itemize}

\begin{table*}[!ht]
\centering
\caption{Configuration of sampler variants used in the results, expressed as the switches of Algs.~\ref{alg:sampler} and \ref{alg:effe3} hyperparameters: solver steps $N$, corrector gate $\sigma_{\max}^c$, momentum $m$, corrector noise scale $\lambda$, and DSG normalization. NFE $=$ solver steps $+$ corrector evaluations. Tags are cumulative additions to the DPS baseline; (N25)/(N30) denote reduced-step variants and (all\,$\sigma$) an ungated corrector.}
\label{tab:variants}
\footnotesize
\setlength{\tabcolsep}{5pt}
\begin{tabular}{l c c c c c r}
\toprule
Sampler & steps $N$ & corrector & $m$ & $\lambda$ & DSG & NFE \\
\midrule
DPS  & 50 & - & 0 & - & - & 50 \\
+mom0.5      & 50 & - & 0.5 & - & - & 50 \\
+corr      & 50 & $\sigma{<}3$ & 0 & 1 & - & 76 \\
+corr+$\lambda$0  & 50 & $\sigma{<}3$ & 0 & 0 & - & 76 \\
+corr+$\lambda$0+mom & 50 & $\sigma{<}3$ & 0.3 & 0 & - & 76 \\
+corr+DSG   & 50 & $\sigma{<}3$ & 0 & 1 & \cmark & 76 \\
+corr+DSG+$\lambda$0   & 50 & $\sigma{<}3$ & 0 & 0 & \cmark & 76 \\
+corr(all\,$\sigma$)    & 50 & all $\sigma$ & 0 & 1 & - & 99 \\
+corr(all\,$\sigma$)+$\lambda$0    & 50 & all $\sigma$ & 0 & 0 & - & 99 \\
+corr(all\,$\sigma$)+$\lambda$0+mom  & 50 & all $\sigma$ & 0.3 & 0 & - & 99 \\
+corr\,(N25)         & 25 & $\sigma{<}3$ & 0 & 1 & - & 38 \\
+corr\,(N30)      & 30 & $\sigma{<}5$ & 0 & 1 & - & 47 \\
+corr\,(N30)+$\lambda$0      & 30 & $\sigma{<}5$ & 0 & 0 & - & 47 \\
\bottomrule
\end{tabular}
\end{table*}

\subsection{Measurement operators}
\label{sec:operators}
For our posterior sampling, we change only the measurement operator $A$ entering the likelihood. We broadly divide the measurement operators used here into these three categories.

\begin{itemize}
  \item \textbf{Structured (gridded) subsampling.} For SR-type tasks with a structured observation, $A$ selects the observed frames of the decoded video and spatially subsamples them by the coarsening factor. 

  \item \textbf{Sparse point interpolation.} For station datasets, $A$ bilinearly interpolates the decoded fields at the reported $[\mathrm{lat},\mathrm{lon}]$ locations on the observed frames. The observations are fused by taking their union.

  \item \textbf{Super-obbed point observations.} For the dense, strongly clustered observations (ISD and the ICOADS), we use grid-cell \textit{super-obbing}, the standard assimilation thinning step \cite{lorenc1981global,ochotta2005adaptive}. All raw observations in a cell are collapsed to a single super-observation weighted by their distance from the cell centers.
  
\end{itemize}

All likelihoods take the form of Eq.~\eqref{eq:lik}-\eqref{eq:sda}.

\section{Experimental configuration}

\subsection{Data}
\label{sec:obsdata}
We use ERA5 reanalysis \cite{hersbach2020era5} regridded to $1.40625^\circ$ ($128{\times}256$) from WeatherBench \cite{rasp2020weatherbench}, organized into 32-frame 6-hourly windows of 69 variables (surface fields plus geopotential, wind, temperature, and specific humidity on 13 pressure levels). The flow prior is trained on autoencoder latents of these windows. The years $1979-2018$ are used for training, $2019$ for validation, and $2020$ for testing.

For observations we use the NOAA Integrated Global Radiosonde Archive (IGRA) \cite{durre2006overview} and the Integrated Surface Database (ISD; the ISD-Lite subset) \cite{smith2011integrated}. IGRA provides on average ${\sim}300$ active stations per observed frame (ranging from ${\sim}600$ at the $00$/$12$\,UTC launch times down to a few tens at $06$/$18$\,UTC; ${\sim}900$ distinct sites), reporting 63 of the 69 variables: the three near-surface fields (2\,m temperature, $10$\,m $u,v$ wind) and, on every one of the thirteen pressure levels, geopotential, zonal and meridional wind, and temperature, together with specific humidity at $300$\,hPa and below. The soundings carry no mean sea-level pressure, and we discard radiosonde humidity above $300$\,hPa, where the sensors are unreliable and the reports correlate poorly with the reanalysis. ISD provides four \emph{surface} variables (2\,m temperature, 10\,m $u/v$ wind, mean sea-level pressure) from ${\sim}2500$ globally grid-thinned stations per step (${\sim}3100$ distinct sites, ${\sim}9400$ observations per 6\,h step): a dense but surface-only observation. For the observations-to-forecast experiments, we additionally ingest the International Comprehensive Ocean-Atmosphere Data Set (ICOADS) \cite{freeman2017icoads} ship and buoy reports (marine air temperature, wind, and pressure), which extend the surface coverage over the oceans. The ERA5 data closely match the observations for the temperature fields but less so for the winds, whose local behavior cannot be captured by ERA5 at ${\approx}1.4^\circ$. Although we note that these observations are closer to the ground truth at those stations (up to measurement accuracy), we use ERA5 data for the evaluation of global reconstructions.

\subsection{Autoencoder}
\label{sec:autoencoder}
The flow-matching prior operates in the latent space of a 3D-convolutional autoencoder trained separately on the same ERA5 windows. The encoder is a ResNet-style 3D network with four resolution levels (hidden widths $[96,192,384,768]$, three residual blocks per level) that maps a 69-variable, 32-timeframe window at $128{\times}256$ down to a latent of shape $(C{=}128,T{=}8,H{=}32,W{=}64)$ via the anisotropic strides
$[(1,2,2),(2,2,2),(2,1,1)]$, a $4\times$ temporal and $4\times$ spatial compression; the decoder mirrors the encoder with one extra residual block per level. The latent is passed through a smooth transformation $x\mapsto x/\sqrt{1+(x/10)^2}$ that bounds its range without hard clipping. Two static fields, the land-sea mask and the surface geopotential, are area-resampled to each decoder resolution and added to the decoder through zero-initialized $3{\times}3$ convolutions. This ensures that the decoder need not expend its capacity to learn the fixed surface structures.

The autoencoder is trained deterministically for $155$ epochs with Adam \cite{kingma2014adam}, a base learning rate $5{\times}10^{-4}$ cosine-annealed to $10^{-8}$, gradient
clipping at $5$, and a batch size of $128$. A small Gaussian latent noise (std $0.02$) is injected before the decoder during training as a regularizer. The reconstruction objective is a latitude-weighted and variance-normalized MSE, so high-variance fields do not dominate. We also have three terms to preserve small-scale details, each weighted $0.05$: a
spatial gradient, a Laplacian for second-order sharpness, and a temporal-difference penalty. The trained autoencoder is frozen for all posterior sampling, and every likelihood is evaluated in physical field space through its decoder. The complete architecture and training configuration are tabulated in Appendix~\ref{app:training}. We tune these hyperparameters empirically using a smaller subset of training data.

\subsection{Flow-matching training and sampling}
The prior $F$ is a DiT3D \cite{peebles2023scalable} trained with the modified TrigFlow loss Eq.~\ref{eq:adaptive} on $(128,8,32,64)$ latents. It uses a $[1,2,2]$ patch so the $(8,32,64)$ latent grid becomes a $4096$-token sequence. The DiT3D uses hidden size $1536$, depth $12$, $24$ attention heads, and MLP ratio $4$ with a size-$384$ sinusoidal timestep embedding, and is trained with log-noise $\ln\sigma\sim\mathcal N(0,\,1.5^2)$ and $\sd{=}1$ using AdamW \cite{loshchilov2017decoupled} and learning rate $2{\times}10^{-4}$, weight decay $0.01$, batch size $256$, with an exponential moving average of the weights with a half-life of $500$\,kilo-images (kimg) and a $500$-kimg learning-rate ramp-up. The prior is never fine-tuned or conditioned on observations; all results use a single EMA checkpoint at $19{,}901$\,kimg. The full architecture and training details are in Appendix~\ref{app:training}. The sampling hyperparameters are given in their respective subsections.

\subsection{Metric and cost}
\label{sec:metric}
The common metric used is the per-variable latitude-weighted RMSE in \emph{physical units}, following the WeatherBench evaluation convention \cite{rasp2020weatherbench}: for each variable we compute the area-weighted (cosine-latitude) RMSE of the \emph{ensemble-mean} prediction against ERA5, then average over frames. For variable $v$, with $\bar u=\tfrac1d\sum_{k=1}^{d}u^{(k)}$ the mean over $d$ posterior draws,
ERA5 truth $u^{\star}$, latitude $\phi_i$, and frame index $f$ running over the
$T{=}32$ decoded frames (or a stated frame subset),
\begin{equation}
\mathrm{RMSE}_v=\frac1T\sum_{f=1}^{T}\sqrt{\frac{\sum_{i,j}\cos\phi_i\,(\bar u_{v,f,i,j}-u^{\star}_{v,f,i,j})^2}{\sum_{i,j}\cos\phi_i}} .
\label{eq:rmse}
\end{equation}

The reported values in the results are mean\,$\pm$\, standard deviation over eight independent initial conditions (distinct, non-overlapping ERA5 windows), and every prediction is the mean of an 8-member ensemble, each member an independent posterior draw from a distinct seed. The sampler ablations are done in the validation split, and the forecast results are done in a disjoint test split of the data. For some tasks like forecasts, we have used more initial conditions and ensemble members, which are mentioned in those sections.

\section{Results}
\label{sec:results}

We first recover the full state from
structured coarse observations, which we call super-resolution (SR) in Sec.~\ref{sec:sr}, and from real
station observations, which we call data fusion in Sec.~\ref{sec:obs}. We then evaluate each case for posterior accuracy, ensemble spread, physical consistency, and performance against unseen observations. We perform extensive ablation studies on the different sampling methods (Appendix~\ref{sec:samplers}) and propose a set of recommended sampling hyperparameters in Table~\ref{tab:config}. Next, we explore the classical filtering, smoothing, and fixed-interval
regimes of traditional DA with the same fixed prior (Sec.~\ref{sec:filtering}). Finally, we perform forecasting directly from observations (Sec.~\ref{sec:forecast}), including a case study of Hurricane Laura.

\begin{table*}[!ht]
\centering
\caption{Recommended sampler configurations per task: the \emph{efficiency}
column gives the best low-NFE configuration and the \emph{accuracy} column the
best overall configuration in the corresponding ablation tables (key switches
as in Table~\ref{tab:variants}; NFE in parentheses).}
\label{tab:config}
\small
\setlength{\tabcolsep}{5pt}
\begin{tabular}{l l l l}
\toprule
task & efficiency recommendation & accuracy recommendation & Reference \\
\midrule
super-resolution & DPS+corr\,(N30) (47) & DPS+corr(all\,$\sigma$) (99) & Tables~\ref{tab:multisample},~\ref{tab:improved_sr} \\
stations (IGRA/ISD/joint) & DPS+corr\,(N30) (47) & DPS+corr+DSG+$\lambda$0 (76) & Tables~\ref{tab:igra},~\ref{tab:improved_igra},~\ref{tab:obs_bakeoff} \\
obs $\to$ forecast & - & DPS+corr+DSG+$\lambda$0 (76) & Fig.~\ref{fig:obs2forecast_chain} \\
filtering/smoothing study & DPS+corr\,(N30) (47) & DPS+corr+DSG+$\lambda$0 (76) & Table~\ref{tab:filtering}, Fig.~\ref{fig:filtering} \\
\bottomrule
\end{tabular}

\end{table*}

\subsection{Super-resolution}
\label{sec:sr}
\begin{figure}[!ht]
\centering
\includegraphics[width=0.95\linewidth]{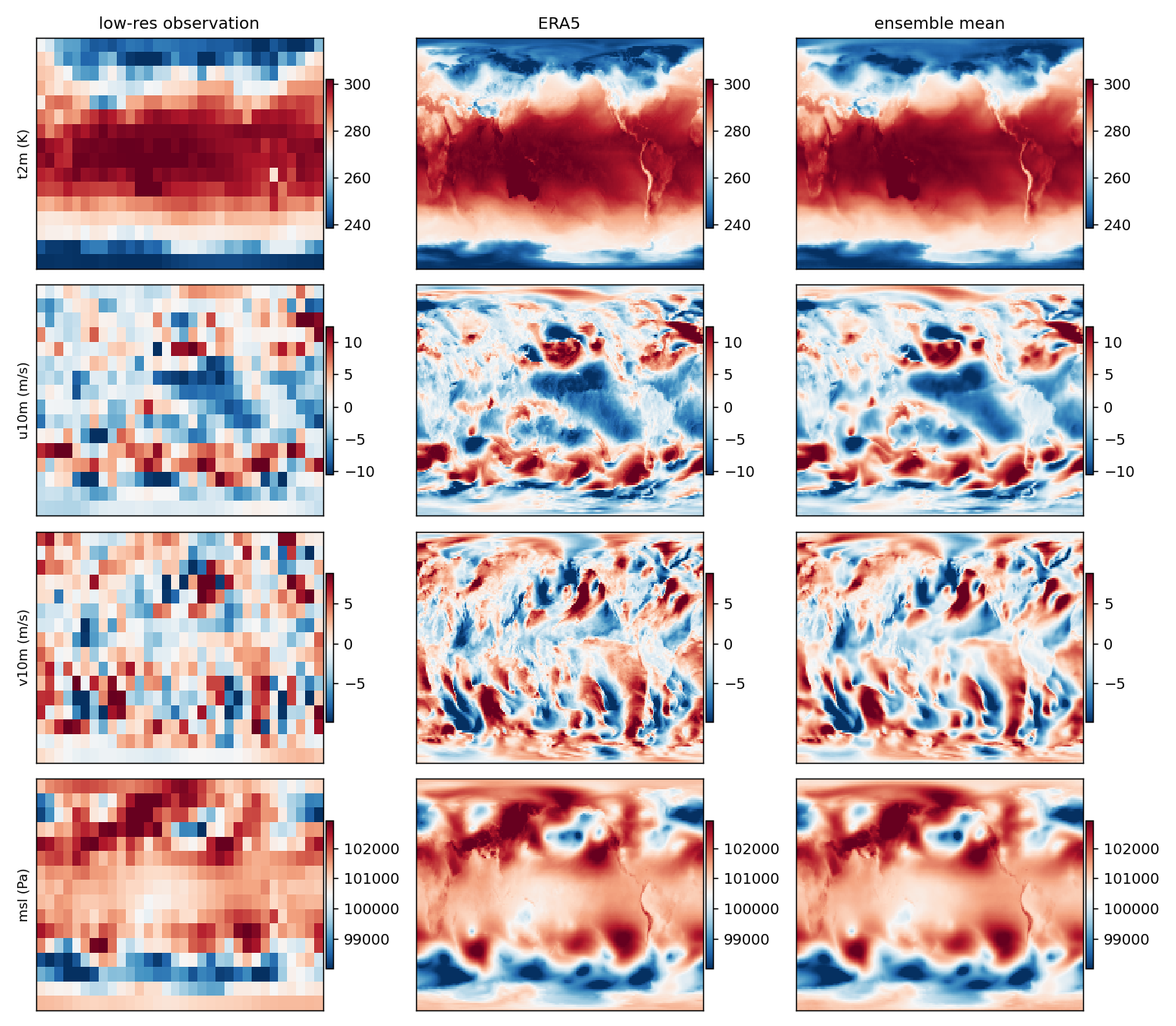}
\caption{Super-resolution reconstruction on an observed frame for the surface variables: the low-resolution observation (left), ERA5 truth (middle), and posterior ensemble mean (right). The prior recovers fine-scale structure absent from the coarsened observations. All 69 channels are shown in Appendix~\ref{app:allvars}.}
\label{fig:sr_fields}
\end{figure}

\begin{figure}[!ht]
\centering
\includegraphics[width=0.9\linewidth]{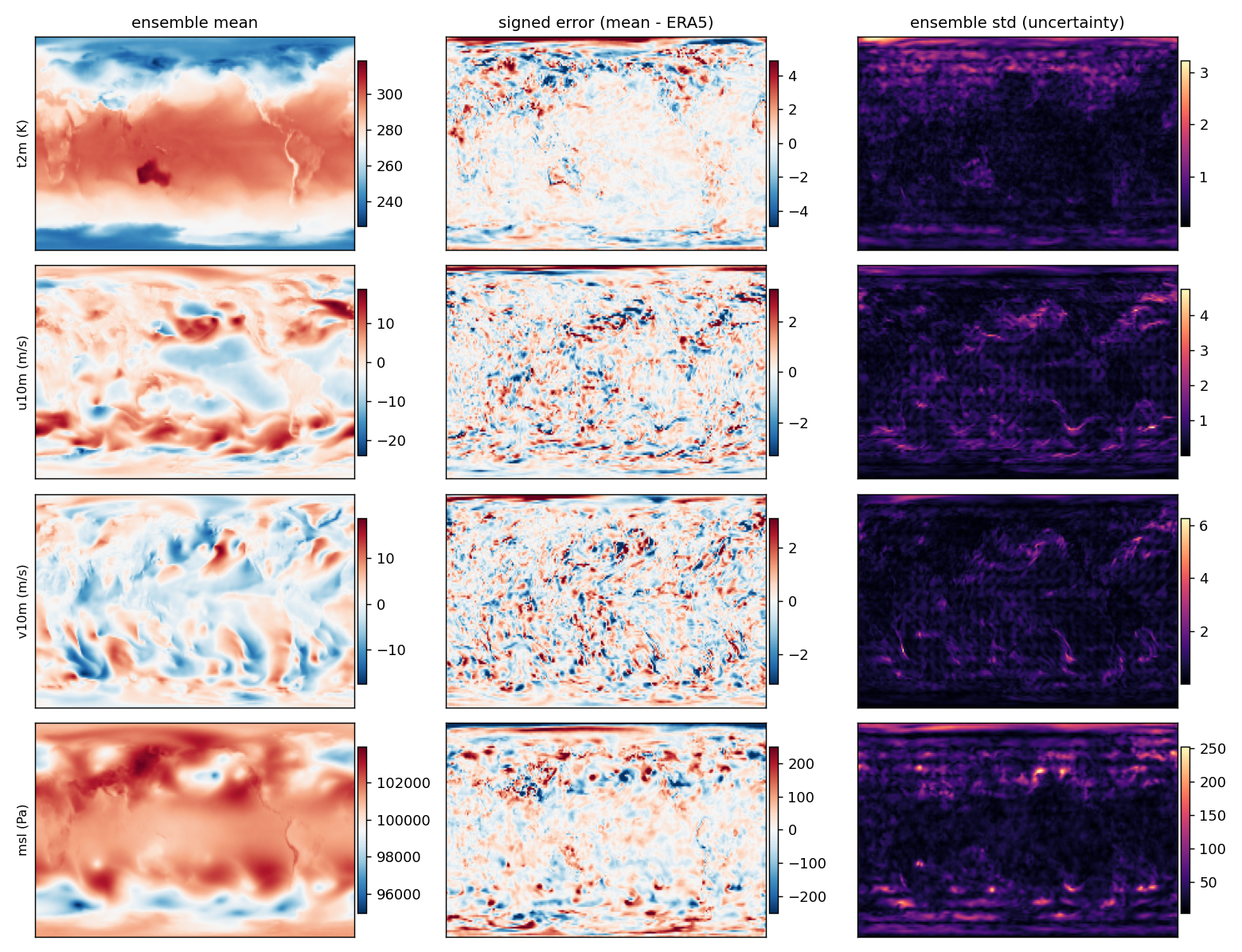}
\caption{Super-resolution uncertainty quantification: posterior ensemble mean (left),
signed error of the mean from ERA5 (middle), and ensemble
standard deviation (right) of an eight-seed ensemble. Spread is lowest over the
regular observation grid and highest over dynamically active regions, where the
error also concentrates.}
\label{fig:sr_uq}
\end{figure}

\begin{figure}[!ht]
\centering
\includegraphics[width=0.95\linewidth]{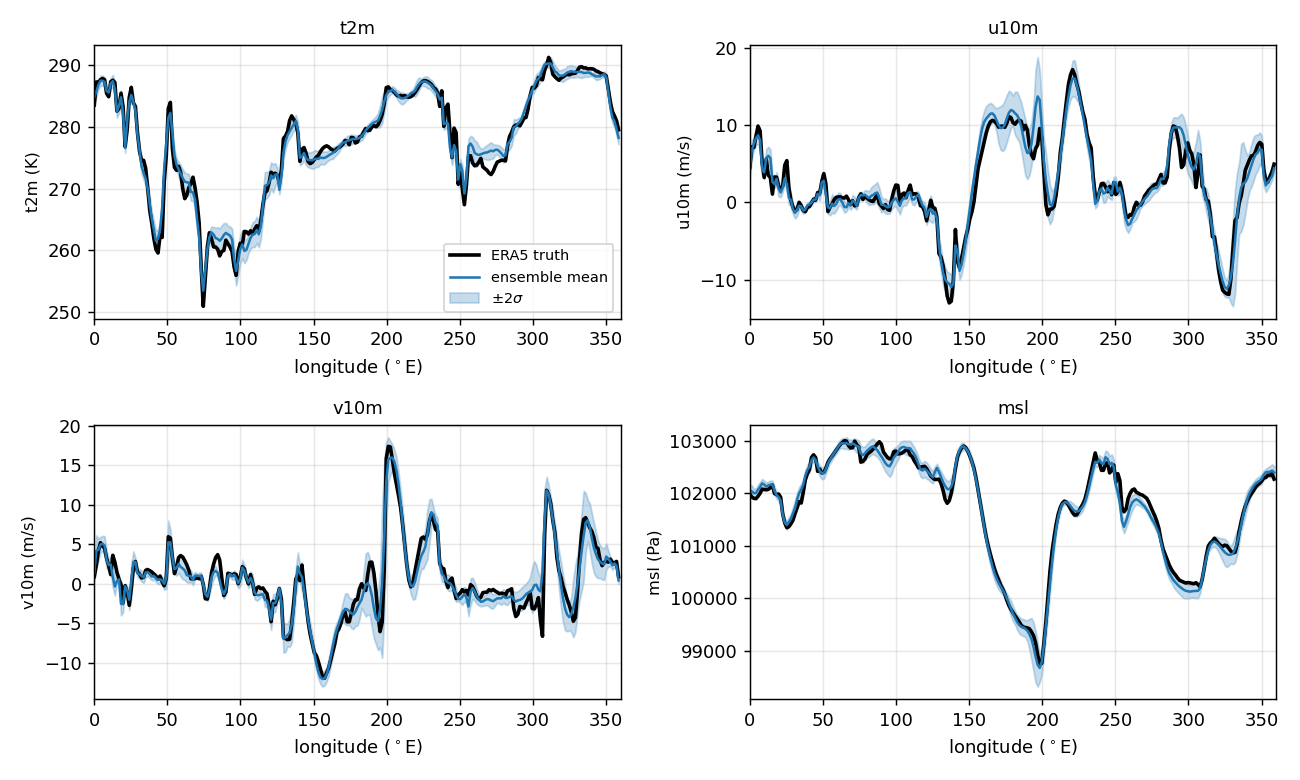}
\caption{Longitudinal traces at $40.4^\circ$N: ERA5 truth vs.\
the posterior ensemble mean with a $\pm2\sigma$ band.}
\label{fig:sr_traces}
\end{figure}

\begin{figure}[!ht]
\centering
\includegraphics[width=0.95\linewidth]{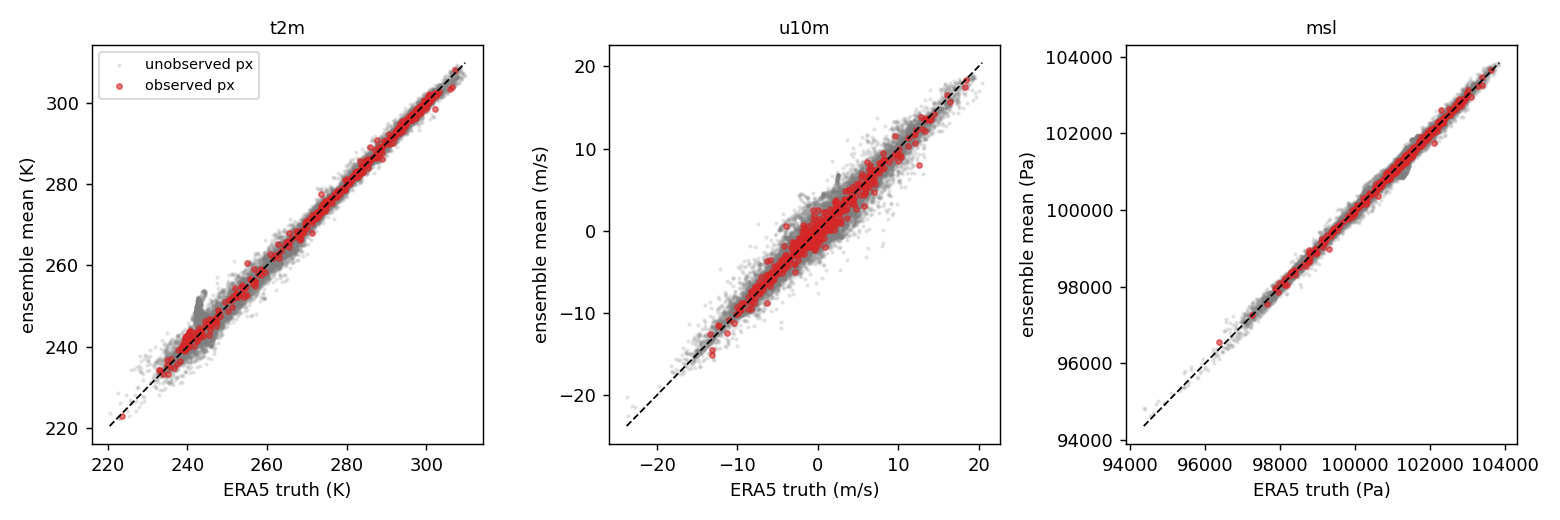}
\caption{Reconstruction scatter: ensemble mean vs.\ ERA5 truth at
observed (red) and unobserved (grey) pixels. Points on the diagonal indicate exact
recovery.}
\label{fig:sr_scatter}
\end{figure}

\begin{figure}[!ht]
\centering
\includegraphics[width=0.95\linewidth]{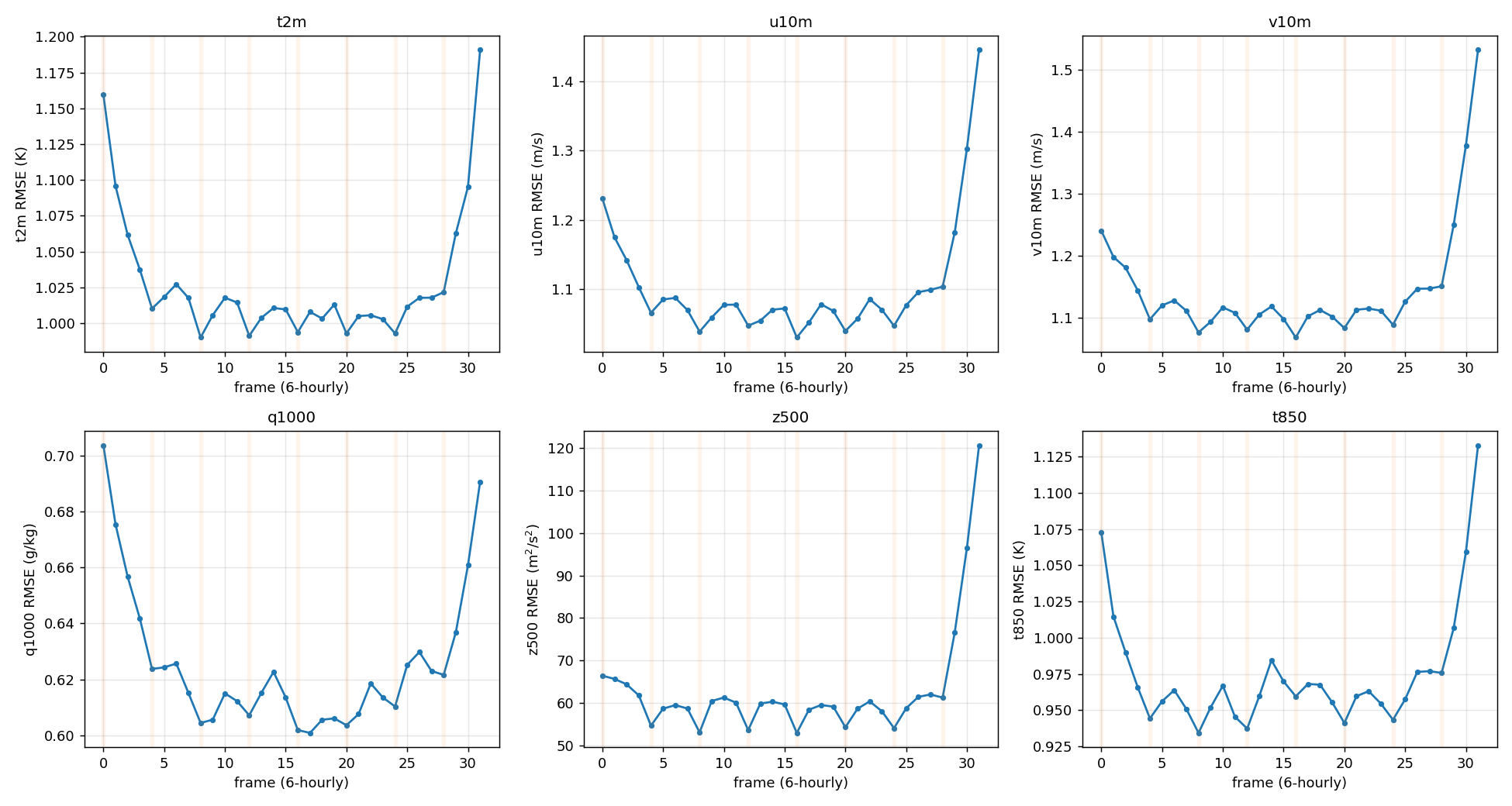}
\caption{SR per-frame physical RMSE per variable (six panels: t2m, u10m, v10m, q1000, z500, t850). Vertical shading marks the observed frames (every fourth).}
\label{fig:sr_rmse_time}
\end{figure}

We first perform a uniform spatio-temporal super-resolution (SR): a $4\times$ temporal and $8\times$-per-axis spatial down-sampling. From these observations the prior recovers 2\,m temperature to $1.02$\,K RMSE with the most accurate configuration (Table~\ref{tab:config}; the efficiency-tuned default gives $1.05$\,K at half the cost; Tables~\ref{tab:multisample},~\ref{tab:improved_sr}). For reference, the autoencoder's own reconstruction floor is $0.84$\,K for 2\,m temperature. Figure~\ref{fig:sr_fields} shows the reconstruction of the surface variables on
an observed frame. The ensemble mean recovers fine-scale structure absent from the coarsened observations. The ensemble spread (Fig.~\ref{fig:sr_uq}) quantifies uncertainty, which is low over the regularly sampled observation grid and higher over dynamically active regions. The corresponding reconstructions for all 69 output channels (the four surface fields and the five upper-air fields at all 13 pressure levels) are reported in Appendix~\ref{app:allvars}. Spectral analysis (Fig.~\ref{fig:sr_spectra}) confirms that the posterior recovers the correct zonal energy spectrum well beyond the observation cutoff, unlike a naive upsampling of the
low-resolution field. The longitudinal traces (Fig.~\ref{fig:sr_traces}) show that the ensemble mean tracks sharp peaks and troughs with an uncertainty band (quantified below). The scatter of reconstructed against true values (Fig.~\ref{fig:sr_scatter}) collapses onto the diagonal at observed pixels and remains tight at unobserved pixels, indicating effective full-state recovery. In time, the per-frame RMSE
(Fig.~\ref{fig:sr_rmse_time}) has a sawtooth structure that dips at each observed frame (shaded orange). We see the error drop after the first frame as the prior's internal dynamics (and two-sided information) help reconstruction. We observe higher errors after the last observed frame, as expected. The following paragraphs explore whether the recovered posterior has calibrated spread and physical consistency.

\begin{figure}[!ht]
\centering
\includegraphics[width=0.9\linewidth]{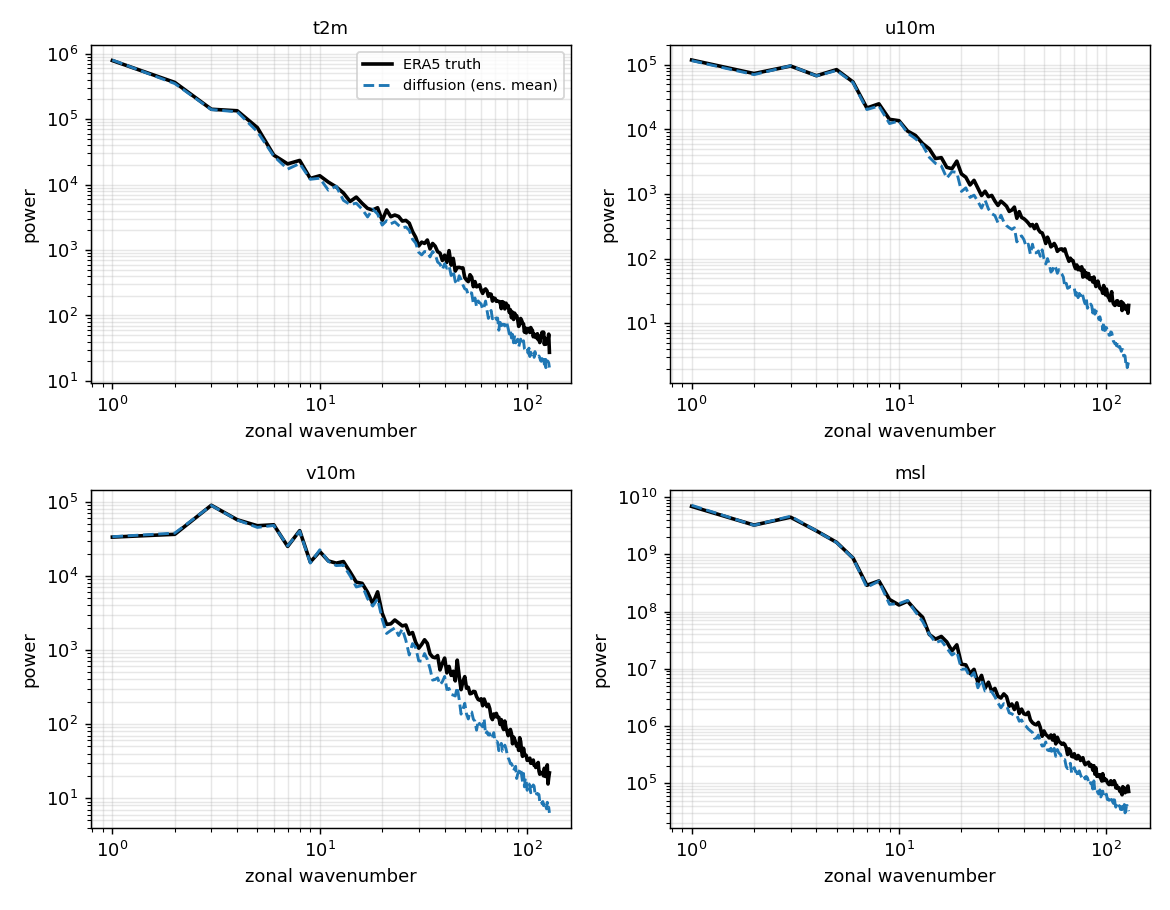}
\caption{Zonal power spectra. The posterior sampling recovers the
correct spectral slope up to high wavenumbers.}
\label{fig:sr_spectra}
\end{figure}

\paragraph{Calibration of the posterior spread.}
An ensemble is useful only if its spread is calibrated, that is, if the mean is wrong by about 1 \, K (skill), then the members should typically differ by 1 \, K (spread). We compute a total spread-skill ratio (SSR) that combines all 69 variables in their standardized and latitude-weighted units (1 is ideal). Table~\ref{tab:calib} reports it for 8-member ensembles of various sampling configurations. As expected, we observe that the setting $\lambda{=}0$ removes the injected noise from the corrector, making the SSR worse. For the sparse radiosonde observations, sampling (using $\lambda{=}1$) is the closest to calibrated (SSR $0.85$). On the densely observed uniform SR task, the sampling configurations are more under-dispersed (SSR $0.72$ to $0.75$), meaning the spread is smaller relative to error. In short, $\lambda{=}1$ remains the choice whenever a calibrated spread is required. In future work, we would need to utilize an inflation-style correction, as operational ensembles routinely apply.

\begin{table}[!ht]
\centering
\caption{Ensemble calibration, 8-member ensembles for every configuration. Total Spread-Skill Ratio (SSR) combines all 69 variables in
standardized units (latitude weighted, scaled by $\sqrt{(d{+}1)/d}$ for a
$d$-member ensemble; $1$ is ideal, and values below $1$ mean the ensemble is
overconfident). Empirical $\pm2\sigma$ coverage has Gaussian target $0.954$.  The unobs.\ restricts to the unobserved frames.}
\label{tab:calib}
\small
\begin{tabular}{llcccc}
\toprule
task & configuration & SSR & Empirical coverage & SSR unobs. & Empirical coverage unobs. \\
\midrule
SR & DPS ($\lambda{=}1$) & 0.72 & 0.772 & 0.74 & 0.777 \\
SR & +corr\,(N30) ($\lambda{=}1$) & 0.72 & 0.766 & 0.74 & 0.770 \\
SR & +corr(all\,$\sigma$) ($\lambda{=}1$) & 0.73 & 0.773 & 0.74 & 0.778 \\
SR & +mom0.5 ($\lambda{=}1$) & 0.75 & 0.778 & 0.76 & 0.784 \\
SR & +corr\,(N30)+$\lambda$0 ($\lambda{=}0$) & 0.52 & 0.565 & 0.53 & 0.570 \\
SR & +corr(all\,$\sigma$)+$\lambda$0 ($\lambda{=}0$) & 0.23 & 0.172 & 0.24 & 0.178 \\
\midrule
IGRA & +corr+DSG ($\lambda{=}1$) & 0.85 & 0.803 & 0.87 & 0.818 \\
IGRA & +corr+DSG+$\lambda$0 ($\lambda{=}0$) & 0.83 & 0.732 & 0.85 & 0.763 \\
\bottomrule
\end{tabular}
\end{table}

\paragraph{Physical consistency.}
A posterior can score well on RMSE while producing physically inconsistent states, so we test the samples against two atmospheric balances (following the physical-consistency checks in prior work \cite{andry2025appa}) (Table~\ref{tab:physcons}). The first is geostrophic balance: in the extratropics, the wind should blow along the geopotential contours. We assess this by comparing the actual 500\,hPa wind with the geostrophic wind computed from the sampled geopotential gradient. The second is hypsometric consistency: the 500-850\,hPa thickness dictates the layer-mean temperature, so the thickness-implied temperature should match the directly sampled one. We observe that the posterior samples are at least as geostrophically balanced as the reanalysis in every configuration. In samples conditioned on IGRA measurements, the hypsometric residual is marginally above the reference ($0.98$-$0.99$\,K versus $0.95$\,K). In general, the samples are dynamically and physically plausible states.

\begin{table}[!ht]
\centering
\caption{Physical consistency of the posterior samples. Geostrophic balance at 500\,hPa in the extratropics ($25^\circ{\le}|\mathrm{lat}|{\le}70^\circ$): cosine similarity between actual and geostrophic wind, and their speed ratio. Hypsometric check: RMSE between the 500-850\,hPa thickness-implied layer-mean
temperature and the directly sampled one.}
\label{tab:physcons}
\small
\setlength{\tabcolsep}{5pt}
\begin{tabular}{llcccc}
\toprule
task & configuration & geo.\ cos & speed ratio & hyps.\ RMSE [K] \\
\midrule
- & ERA5 truth & 0.982 & 1.045 & 0.95 \\
SR & DPS & 0.986 & 1.033 & 0.94 \\
SR & +corr(all\,$\sigma$) & 0.986 & 1.045 & 0.93\\
SR & +corr\,(N30)+$\lambda$0 & 0.987 & 1.040 & 0.94 \\
IGRA & +corr+DSG & 0.986 & 1.047 & 0.98 \\
IGRA & +corr+DSG+$\lambda$0 & 0.988 & 1.051 & 0.99\\
\bottomrule
\end{tabular}
\end{table}

\subsection{Real-observation data fusion}
\label{sec:obs}
\begin{figure}[!ht]
\centering
\includegraphics[width=0.95\linewidth]{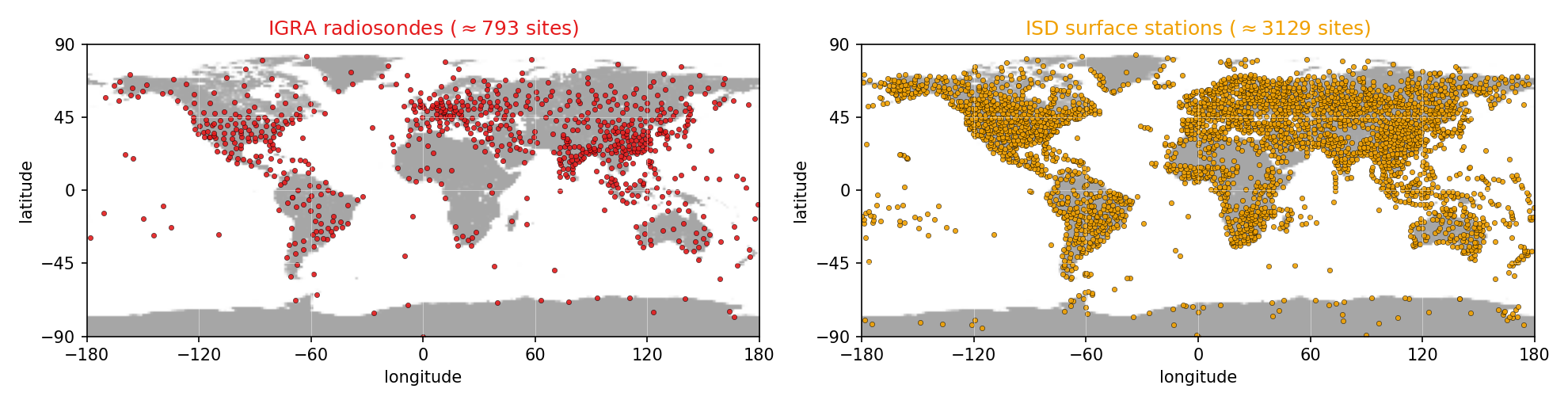}
\caption{The two real observation sources IGRA and ISD (union of reporting sites across the eight observed frames and eight windows).}
\label{fig:station_maps}
\end{figure}

\begin{figure}[!ht]
\centering
\includegraphics[width=0.95\linewidth]{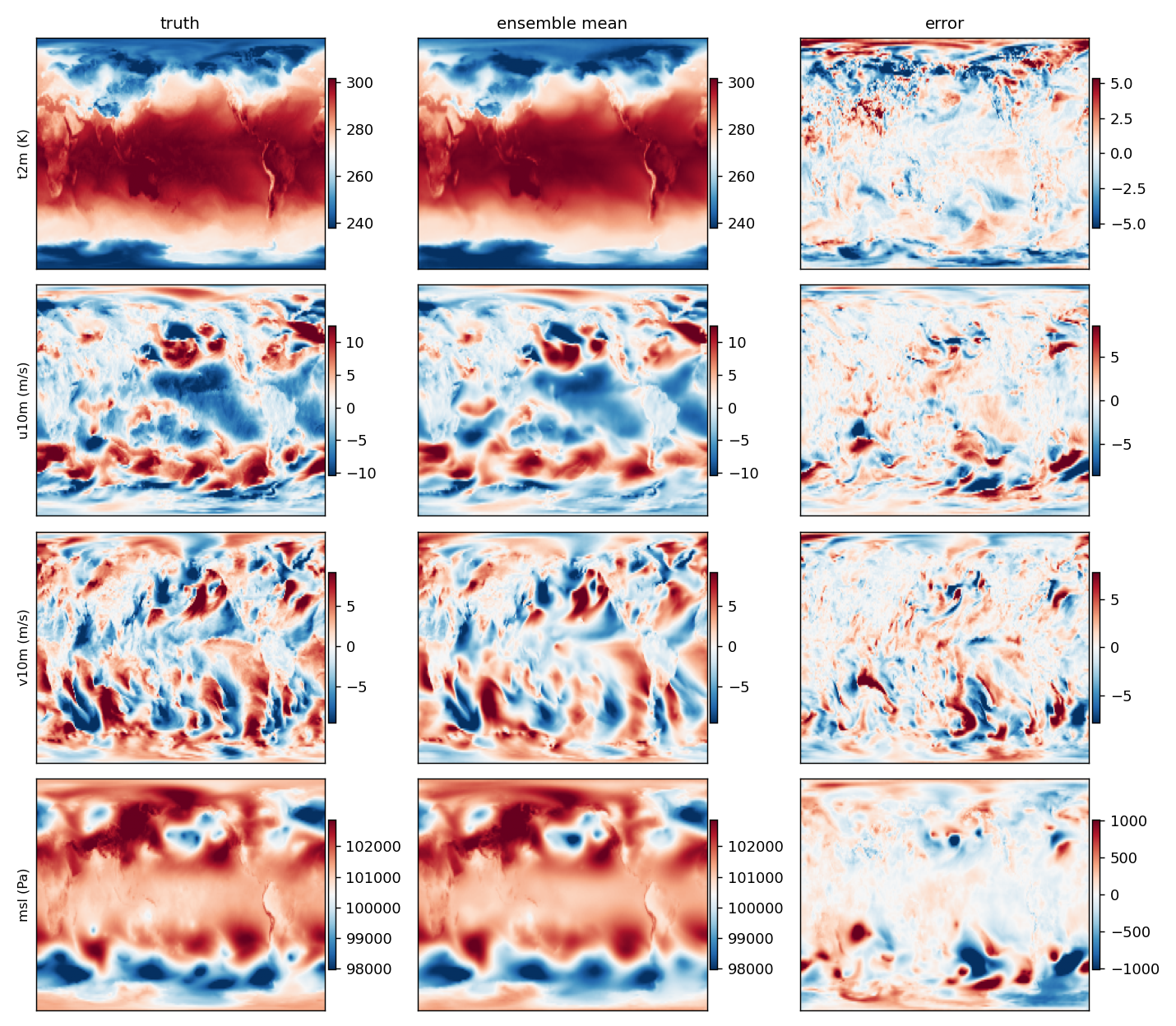}
\caption{IGRA radiosonde data-fusion surface fields: ERA5 truth, posterior ensemble mean, and signed error.}
\label{fig:igra_fields}
\end{figure}

\begin{figure}[!ht]
\centering
\includegraphics[width=0.9\linewidth]{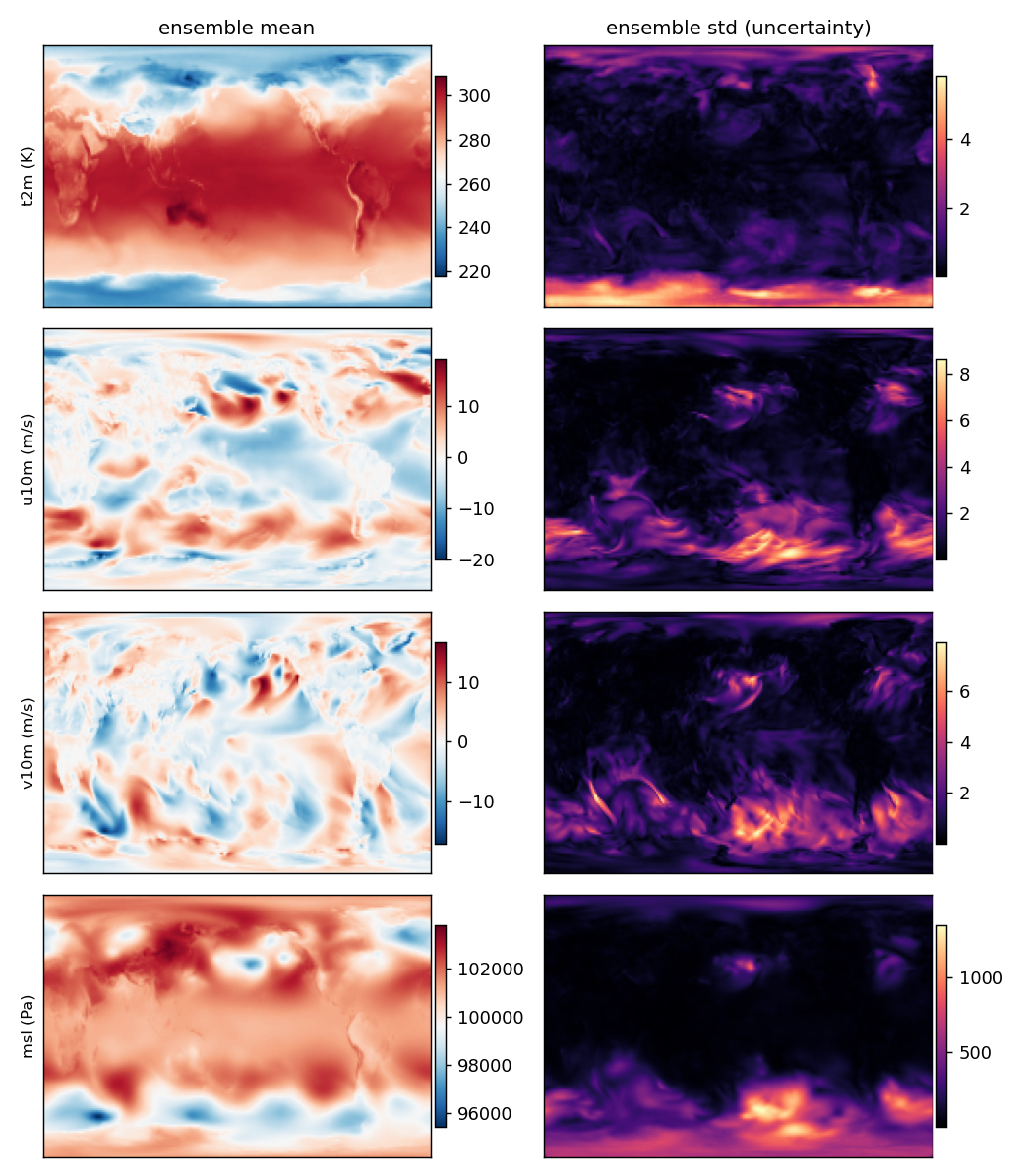}
\caption{IGRA radiosonde data fusion uncertainty quantification: ensemble mean (left) and standard deviation
(right). Spread is lower over the regions with more observation locations.}
\label{fig:igra_uq}
\end{figure}

\begin{figure}[!ht]
\centering
\includegraphics[width=0.95\linewidth]{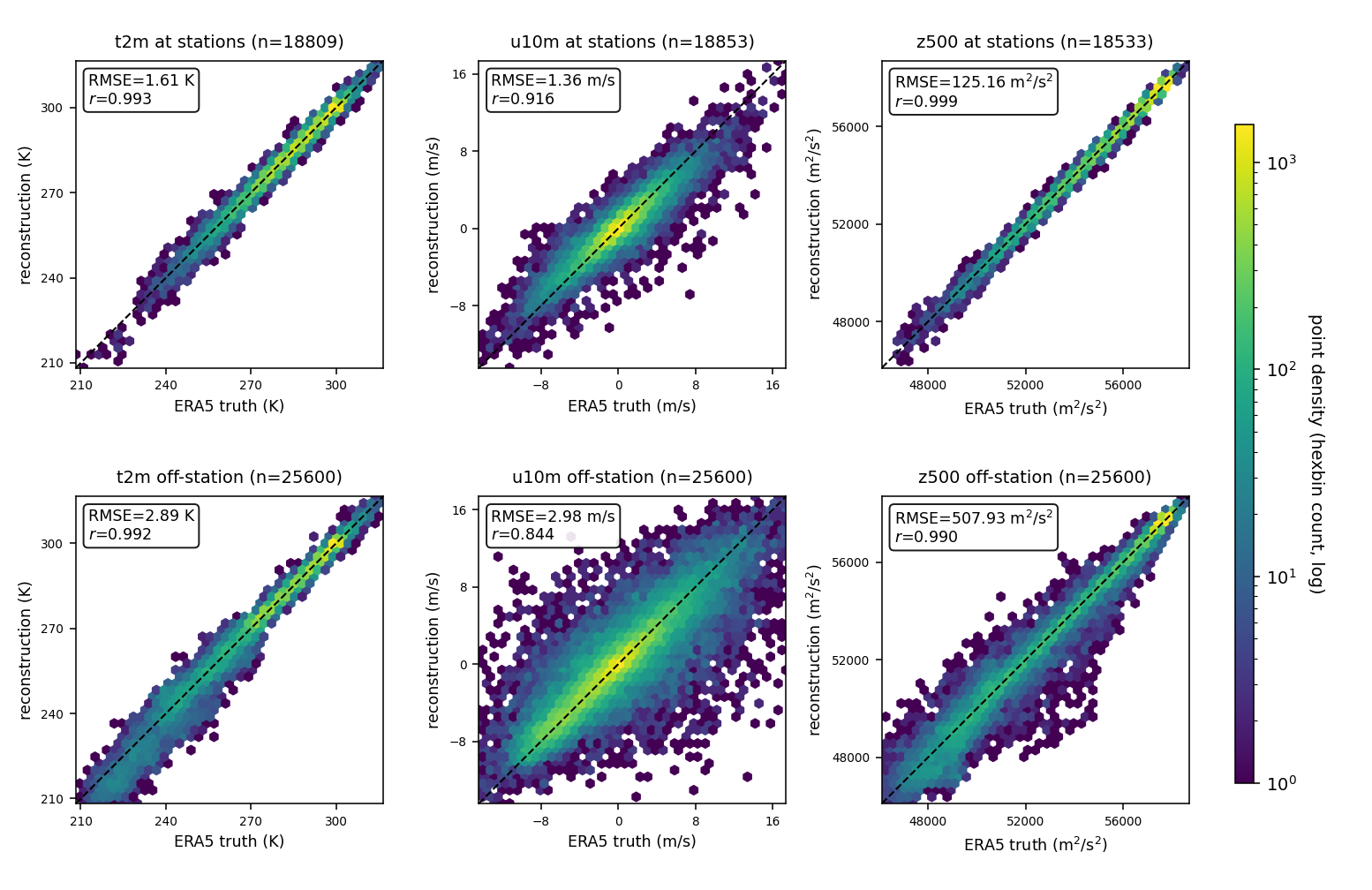}
\caption{IGRA reconstruction vs ERA5 scatter plots for three observed variables (t2m, u10m, z500). \emph{Top}: at the radiosonde station locations; \emph{bottom}: at off-station grid points. Color shows $\log_{10}$ hexbin count; mentioned text gives the RMSE in physical units and Pearson $r$. Similar plots for all observed variables are in Appendix~\ref{app:station_scatter}.}
\label{fig:igra_scatter}
\end{figure}

We now condition on \emph{real}, sparse observations (Fig.~\ref{fig:station_maps}): IGRA measurements and the ISD surface measurements. From these sparse point observations, the prior reconstructs the full 69-channel field (Fig.~\ref{fig:igra_fields}), including the six unobserved channels (mean-sea-level pressure and specific humidity above $300$\,hPa). The ensemble spread of the generated samples (Fig.~\ref{fig:igra_uq}) is lower near the observation stations and higher over the sparse regions. The reconstruction scatter (Fig.~\ref{fig:igra_scatter}) collapses onto the diagonal at the observed station locations (top row), where it largely reflects the data fit, and, more informatively, stays close to the diagonal at off-station grid points (bottom row), quantifying information gain at the soundings. However, some fields, like the winds, are intrinsically small-scale and highly chaotic, which causes higher pixel-wise RMSE error (Tables~\ref{tab:cross},~\ref{tab:igra}). Moreover, the truth here is ERA5, which at the $1.4^\circ$ grid differs considerably from the localized observations. Moreover, even a perfect posterior cannot beat the autoencoder reconstruction floor (Table~\ref{tab:multisample}; e.g.\ u10m $0.53$\,m/s, z500 $37.0$\,m$^2$/s$^2$), so a portion is due to the irreducible decoder error. 

\begin{figure}[!ht]
\centering
\includegraphics[width=0.95\linewidth]{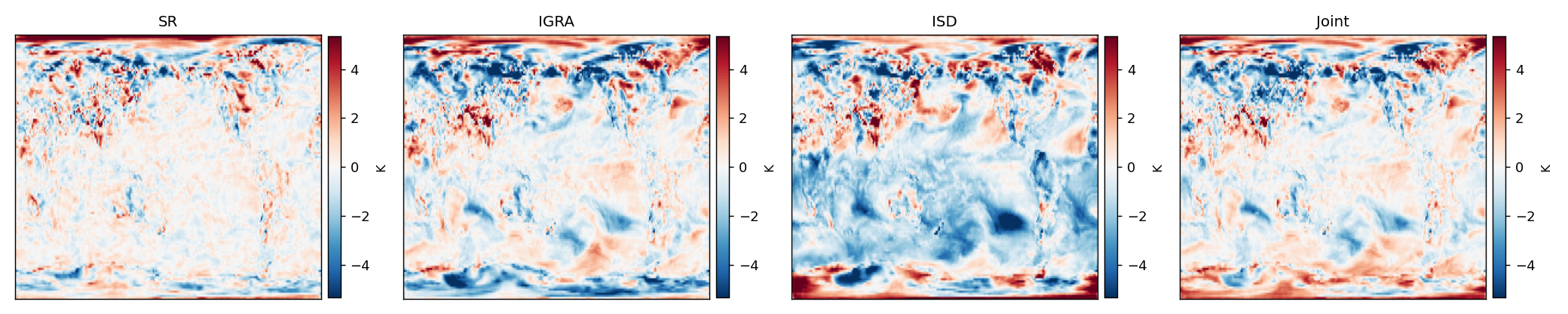}
\caption{2\,m-temperature signed error by assimilation modality. Among the real-observations data-fusion, the joint ISD+IGRA has the lowest error. The SR panel uses structured dense observations and is shown only for reference.}
\label{fig:cross_fields}
\end{figure}

\begin{figure}[!ht]
\centering
\includegraphics[width=0.95\linewidth]{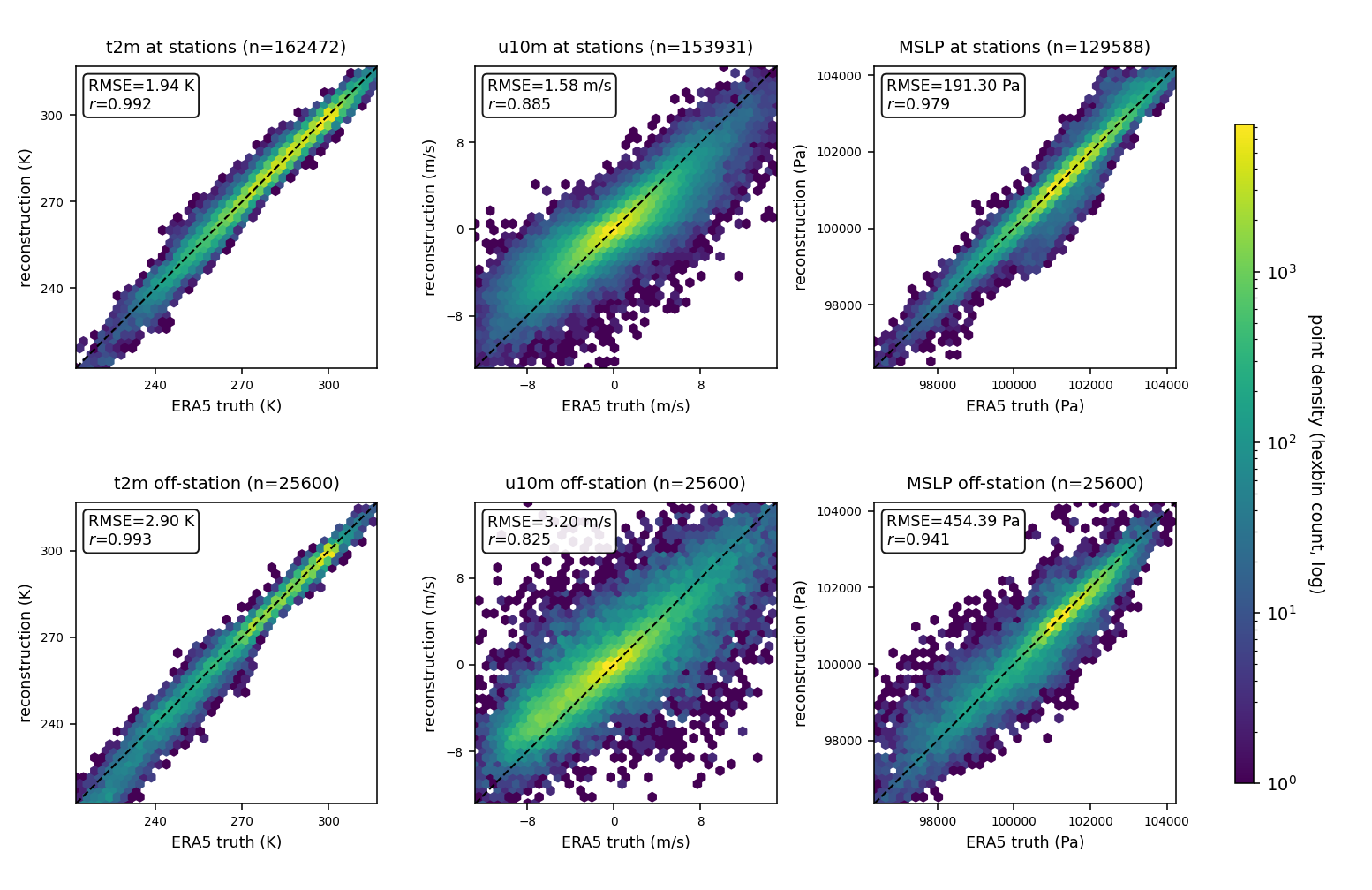}
\caption{ISD surface reconstruction vs ERA5 scatter plots for three observed variables (t2m, u10m, MSLP). \emph{Top}: at the station locations; \emph{bottom}: at off-station grid points (random subsample). Color shows $\log_{10}$ hexbin count; mentioned text gives the RMSE (physical units) and Pearson $r$. All four
observed ISD variables are in Appendix~\ref{app:station_scatter}.}
\label{fig:isd_scatter}
\end{figure}

\paragraph{Multimodal data fusion.}
Intuitively, the addition of more observation points shows lower errors on joint sampling than individually using IGRA and ISD (Table~\ref{tab:cross}). The surface-only ISD is worst despite its density because the prior has to reconstruct every upper-air variable from surface observations alone. The comparison therefore contrasts two whole observing systems: the soundings are sparser on the map but report 63 channels through the depth of the atmosphere against ISD's four. Figure~\ref{fig:cross_fields} visualizes this as the 2\,m-temperature error map across modalities, and Fig.~\ref{fig:isd_scatter} shows the truth-vs-reconstruction scatter for the ISD surface observations.

\begin{table}[!ht]
\centering
\caption{Multimodal data-fusion: per-variable physical RMSE (Units: t2m K, u10m/v10m m/s, q1000 g/kg, z500 m$^2$/s$^2$; bold =
best, underline =
second-best).}
\label{tab:cross}
\small
\begin{tabular}{lccccc}
\toprule
modality & t2m & u10m & v10m & q1000 & z500 \\
\midrule
Joint (IGRA${+}$ISD, 64 var) & \textbf{1.80$\pm$0.11} & \textbf{2.93$\pm$0.14} & \textbf{3.00$\pm$0.17} & \textbf{1.13$\pm$0.04} & \textbf{467.6$\pm$48.5} \\
IGRA (radiosonde, 63 var) & \underline{1.90$\pm$0.13} & \underline{3.01$\pm$0.17} & \underline{3.08$\pm$0.18} & \underline{1.15$\pm$0.04} & \underline{474.9$\pm$50.4} \\
ISD (surface, 4 var) & 2.21$\pm$0.11 & 3.34$\pm$0.18 & 3.33$\pm$0.20 & 1.41$\pm$0.09 & 701.3$\pm$65.8 \\
\bottomrule
\end{tabular}
\end{table}

\paragraph{Matching held-out observations.}
Everything above scores the reconstruction against ERA5, which is not the perfect representation of the ground truth. We therefore verify directly in observation space similar to prior work \cite{alexe2024graphdop}. We withhold a random $20\%$ of stations, assimilate only the remaining $80\%$, and measure the deviation of the reconstruction from the unseen observations. Because station observations are highly localized, a meaningful reference error is the deviation of ERA5 at the same sites. Table~\ref{tab:heldout} shows the result on the observed frames: on the withheld stations the reconstruction is at or below the ERA5 deviations on three of four ISD variables (10\,m winds and pressure) and close for 2\,m temperature. For the IGRA radiosondes, the reconstruction deviation from held-out observations closely tracks that of ERA5. 

\begin{table}[!ht]
\centering
\caption{Observation-space verification at withheld stations. RMSE of the reconstruction versus that of ERA5 at the same sites and times (lower half: withheld IGRA radiosondes; upper half: withheld ISD surface stations). Bold marks where the reconstruction departure is at or below the ERA5 reference.}
\label{tab:heldout}
\small
\setlength{\tabcolsep}{5pt}
\begin{tabular}{llcc}
\toprule
Observation & Variable & RMSE & ERA5 \\
\midrule
ISD & 2\,m temperature [K] & 3.24 & 3.06 \\
ISD & 10\,m $u$-wind [m/s] & \textbf{2.27} & 2.44 \\
ISD & 10\,m $v$-wind [m/s] & \textbf{2.19} & 2.31 \\
ISD & MSLP [hPa] & \textbf{2.16} & 2.50 \\
\midrule
IGRA & 2\,m temperature [K] & 3.16 & 2.91 \\
IGRA & 10\,m $u$-wind [m/s] & \textbf{2.39} & 2.40 \\
IGRA & z500 [m$^2$/s$^2$] & 165 & 155 \\
IGRA & z850 [m$^2$/s$^2$] & 124 & 108 \\
IGRA & t500 [K] & 1.21 & 0.91 \\
IGRA & t850 [K] & 1.58 & 1.23 \\
IGRA & t300 [K] & 1.42 & 1.19 \\
IGRA & u200 [m/s] & 3.77 & 2.95 \\
\bottomrule
\end{tabular}
\end{table}

\subsection{Filtering and smoothing}
\label{sec:filtering}
\begin{figure}[!ht]
\centering
\includegraphics[width=0.85\linewidth]{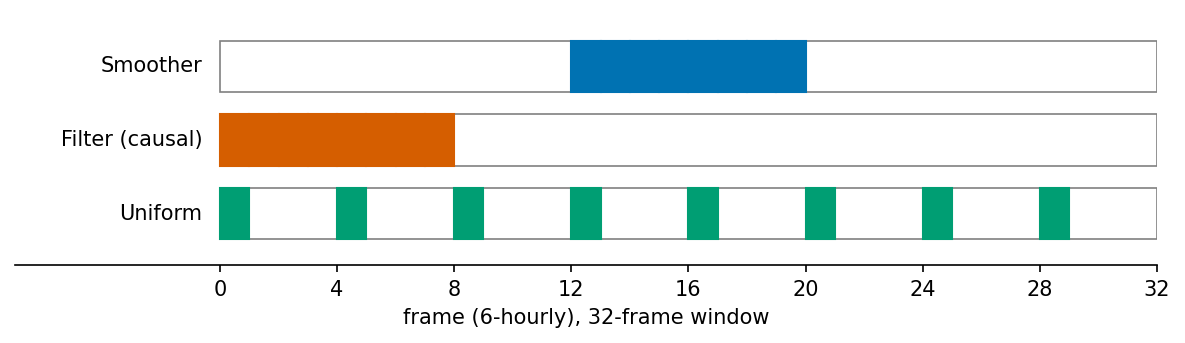}
\caption{The three temporal observation geometries over the 32-frame window, each
observing 8 frames: smoother (two-sided), filter (causal/leading), and the
fixed-interval reference (every fourth frame, the geometry of a reanalysis).
Only \emph{which} frames are observed differs; each realizes one classical DA
regime with the same frozen prior.}
\label{fig:geom}
\end{figure}

We now experiment with the classical estimation regimes of traditional DA \cite{cohn1994fixed} with the same learned prior, changing only which temporal frames are observed. In classical systems, these regimes require distinct frameworks around a numerical forecast model, whereas here, each is just a different observation mask on the same posterior \cite{rozet2023score,andry2025appa}. The smoother observes the middle-8 frames to reconstruct the full 32-frame (8-day) spatio-temporal chunk. The filter observes the leading eight frames, so its unobserved tail is a forecast. We also use a fixed-interval reference which observes one frame per day. A schematic of these tasks is in Fig.~\ref{fig:geom}. We demonstrate these first with the idealized observations (similar to SR), and then with the real IGRA observations. The sampling method is held fixed within each observation source, using the recommended configurations of Table~\ref{tab:config}.

\begin{figure}[!ht]
\centering
\includegraphics[width=0.95\linewidth]{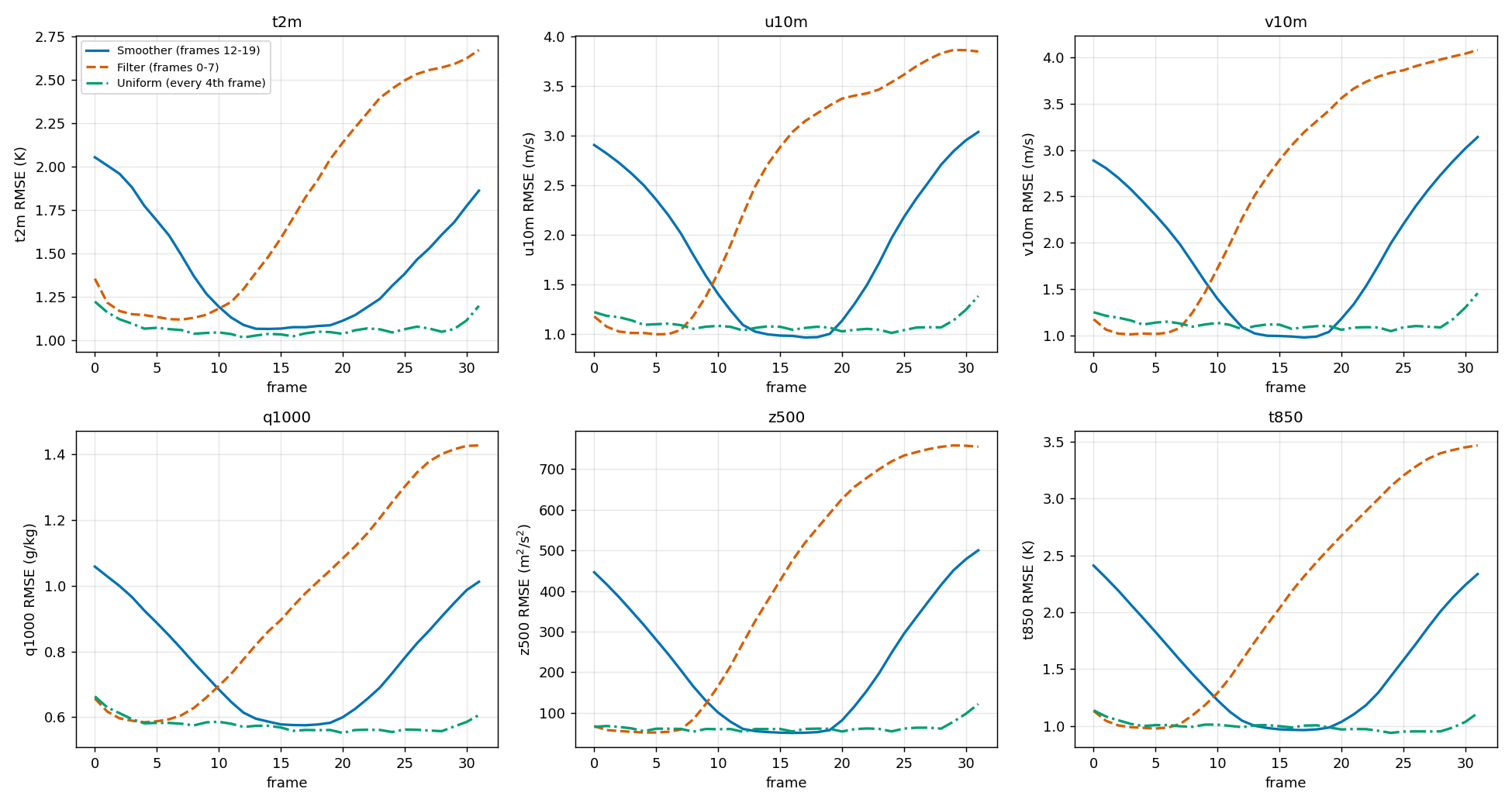}
\caption{DA experiments: per-frame physical RMSE per variable (six panels: t2m, u10m, v10m, q1000, z500, t850) for the three temporal geometries. The filter experiment's error grows forward in time, the smoother is U-shaped, and the fixed-interval reference is flat.}
\label{fig:filtering}
\end{figure}

The motivation of this experiment is to show that a single unconditional prior realizes each regime and shows exactly the behavior classical DA associates with it, in Fig.~\ref{fig:filtering}. Table~\ref{tab:filtering} has the quantitative analysis. The fixed-interval reference flattens the error across the whole window at a low level, as expected when observations constrain the entire trajectory. The smoother gives the familiar U-shape, with the lowest error within the observation window. The filter shows error that is low over the leading observed frames and \emph{grows forward in time} as the prior must forecast beyond the last observation. The error grows smoothly forward, suggesting that the prior learns to propagate the atmospheric fields. Therefore, the prior acts as the learned forecast model, which is the function of a numerical model in classical assimilation. We also run the same three regimes on the \emph{real} IGRA radiosonde observations, the lower half of Table~\ref{tab:filtering}. Together, these show that the same unconditional video prior supports filtering, smoothing, and fixed-interval observing, simply by changing the temporal observation geometry. Classically, each of these would be a separate algorithm wrapped around a forecast model.

\begin{table}[!ht]
\centering
\caption{DA experiments: per-variable physical RMSE over all frames. All geometries observe 8 of the 32 frames; only \emph{which} frames differs. Upper panel: using $8\times$ subsampled observations per axis. Lower panel: using IGRA radiosonde observations; for its fixed-interval row the window starts are aligned to the $00$/$12$\,UTC launch hours. Per-frame observed/unobserved breakdown in Fig.~\ref{fig:filtering} and Appendix~\ref{app:phys}. The frame here refers to the specific timestep of our video window.}
\label{tab:filtering}
\small
\begin{tabular}{lccccc}
\toprule
geometry & t2m & u10m & v10m & q1000 & z500 \\
\midrule
\multicolumn{6}{l}{\emph{SR observations ($8\times$ subsample, sampling configuration is DPS+corr\,(N30))}} \\
\midrule
Smoother (frames 12-19) & 1.40$\pm$0.04 & 1.96$\pm$0.05 & 2.01$\pm$0.07 & 0.85$\pm$0.03 & 232.9$\pm$12.4 \\
Filter (frames 0-7, causal) & 1.85$\pm$0.04 & 2.58$\pm$0.09 & 2.73$\pm$0.07 & 1.05$\pm$0.03 & 412.7$\pm$32.6 \\
Fixed-interval (every 4th frame, ref.) & \textbf{1.03$\pm$0.04} & \textbf{1.10$\pm$0.02} & \textbf{1.15$\pm$0.02} & \textbf{0.63$\pm$0.03} & \textbf{62.8$\pm$1.3} \\
\midrule
\multicolumn{6}{l}{\emph{IGRA radiosondes (real soundings, sampling configuration is DPS+corr+DSG+$\lambda$0)}} \\
\midrule
Smoother (frames 12-19) & 1.90$\pm$0.13 & 3.01$\pm$0.17 & 3.08$\pm$0.18 & 1.15$\pm$0.04 & 474.9$\pm$50.4 \\
Filter (frames 0-7, causal) & 2.13$\pm$0.09 & 3.38$\pm$0.11 & 3.51$\pm$0.10 & 1.26$\pm$0.06 & 621.3$\pm$32.1 \\
Fixed-interval (every 4th frame, ref.) & 1.51$\pm$0.11 & 2.29$\pm$0.10 & 2.28$\pm$0.10 & 0.94$\pm$0.02 & 272.6$\pm$35.8 \\
\bottomrule
\end{tabular}
\end{table}

\subsection{Observations to forecast}
\label{sec:forecast}
The causal filtering geometry of the previous section motivates conditioning on real observations and propagating forward for forecasting, a direct observation-to-forecast \cite{allen2025end,alexe2024graphdop}. We assimilate a global in-situ observation on the leading frames of the window and let the prior forecast the remaining timesteps. Now, since the likelihood is a bilinear interpolation of the decoded field (Sec.~\ref{sec:operators}), it admits only observations that directly measure an ERA5 state variable. This admits three useful in-situ observation sources: IGRA radiosondes, ISD surface stations, and ICOADS ships and buoys (Sec.~\ref{sec:obsdata}; compared to \cite{allen2025end}). It excludes the satellite radiances and scatterometer backscatter, which require a radiative-transfer or geophysical forward
operator. We therefore assimilate the union $\mathrm{ISD}\cup\mathrm{IGRA}\cup\mathrm{ICOADS}$,
a land, ocean, and upper-air network of observations per 6-hourly step.

We therefore apply the grid-cell super-obbing of Sec.~\ref{sec:operators} to the dense ISD and ICOADS observations. This lets the likelihood gradient not be dominated by the over-sampled regions and equalizes each cell's contribution. More details on this are in Appendix~\ref{app:superob}. The sampler which performs best in this case is \textit{DPS+corr+DSG+$\lambda$0} (Table~\ref{tab:variants}: $50$ solver steps, one corrector step at $\sigma{<}3$, $\lambda{=}0$, $76$ NFE). The DSG guidance scale is re-selected on a set of validation windows and applied to a disjoint test set of thirty-two initial conditions spread evenly across 2020, over which we report the mean.

\begin{figure}[!ht]
\centering
\includegraphics[width=0.95\linewidth]{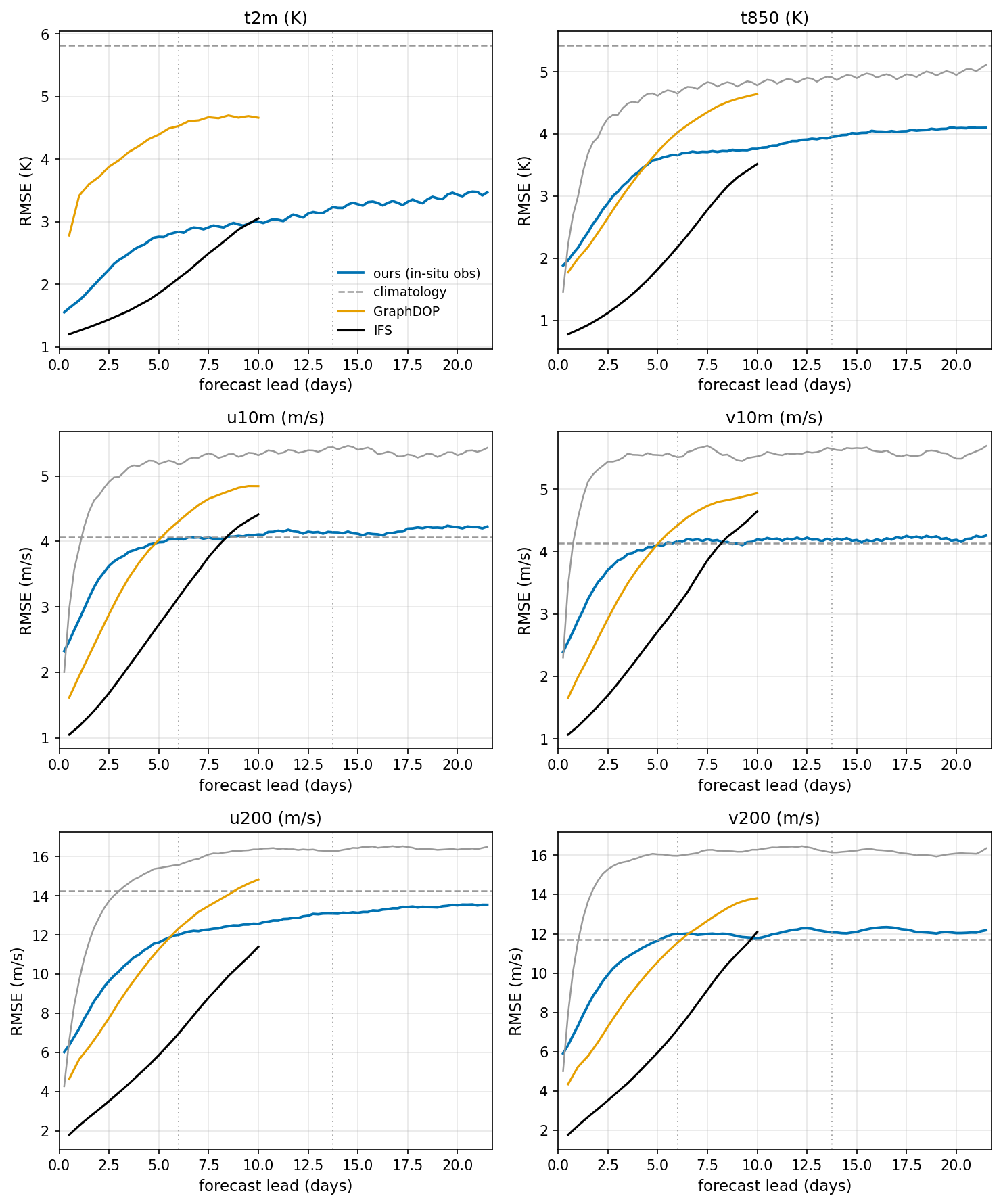}
\caption{Observations-to-forecast skill. Latitude-weighted RMSE against ERA5 versus lead over thirty-two initial conditions spread evenly across 2020, 8-member ensembles. The GraphDOP and IFS curves
\cite{alexe2024graphdop} are extracted from their published figures. The persistence and annual-climatology floors are for reference. Persistence is omitted for 2\,m temperature because of the diurnal cycle.}
\label{fig:obs2forecast_chain}
\end{figure}

\paragraph{Chaining windows into a three-week forecast.}
The same posterior sampler supports autoregressive chaining, an analogue of the cycled reinitialization of operational systems. We take the last generated frame of a window and re-assimilate it as the initial condition of the next window, treating it as a dense observation of that window's first frame. The observation operator for the next window becomes the identity at the first window for every grid cell and all 69 channels. Each additional window extends the forecast by $7$ days and $18$ hours.

Firstly, we observe that the forecast beats persistence of the last observed frame at essentially all leads and stays at or below the annual-climatology error. Figure~\ref{fig:obs2forecast_chain} compares this forecast against GraphDOP \cite{alexe2024graphdop}, the directly comparable observation-only, end-to-end data-driven forecast system, and the operational ECMWF Integrated Forecasting System (IFS), all verified against ERA5. The GraphDOP and IFS skill curves are extracted from the respective figure of \cite{alexe2024graphdop}, every 12\,h through day 10, their full range. The comparison is asymmetric in input information: GraphDOP is trained on a broad slice of the operational observing system, ingesting several satellite measurements alongside the surface, marine, aircraft, and radiosonde measurements \cite{alexe2024graphdop}. However, we still improve on GraphDOP at all leads for 2\,m temperature where our dense surface observation is helpful. For other surface fields, our errors start above GraphDOP and converge to it by days 5 to 7 as forecast error saturates. This suggests that the satellite measurements are very helpful for an accurate forecast, especially at shorter lead times. In longer forecast leads, our methodology shows clear advantages over GraphDOP. We see better stability in the generated forecasts with the metric eventually reaching climatological values.

We perform additional analysis for our observation-to-forecast case. First, we observe that the length of the observed analysis window does not affect the forecast performance. Shortening it from two days to one day, or even to a single analysis frame, leaves forecast skill essentially unchanged (Appendix~\ref{app:superob}). Moreover, in Appendix~\ref{app:aardvark}, we use additional gridded observations from \cite{allen2025end}, where our forecasts perform close to GraphDOP. However, we could perform limited tests due to the availability of only a single observation from \cite{allen2025end}. We leave further tests with operators that can be associated with satellite data to future work.

\begin{figure}[!ht]
\centering
\includegraphics[width=0.8\linewidth]{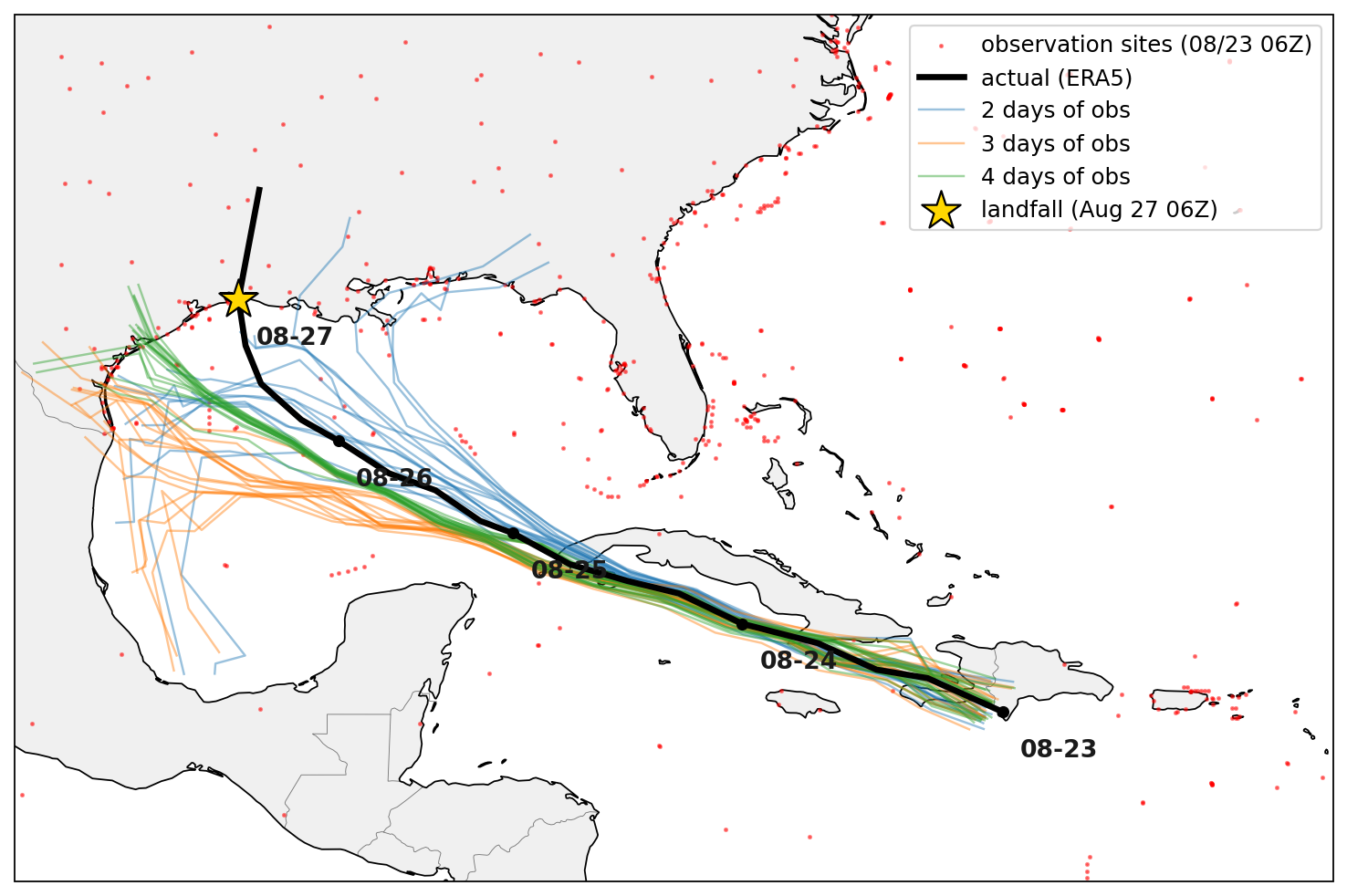}
\caption{Hurricane Laura case study: actual track (ERA5, black; minimum-MSLP) versus the sixteen individual predicted member trajectories for three observed windows starting Aug 23 06Z: two days (blue; landfall at a 54-hour lead), three days (orange), and four days (green). Red dots show the actual observation sites (stations and ships) at the first analysis time; the star marks landfall (Aug 27 06Z,
Cameron, Louisiana).}
\label{fig:laura}
\end{figure}

\begin{figure}[!ht]
\centering
\includegraphics[width=0.8\linewidth]{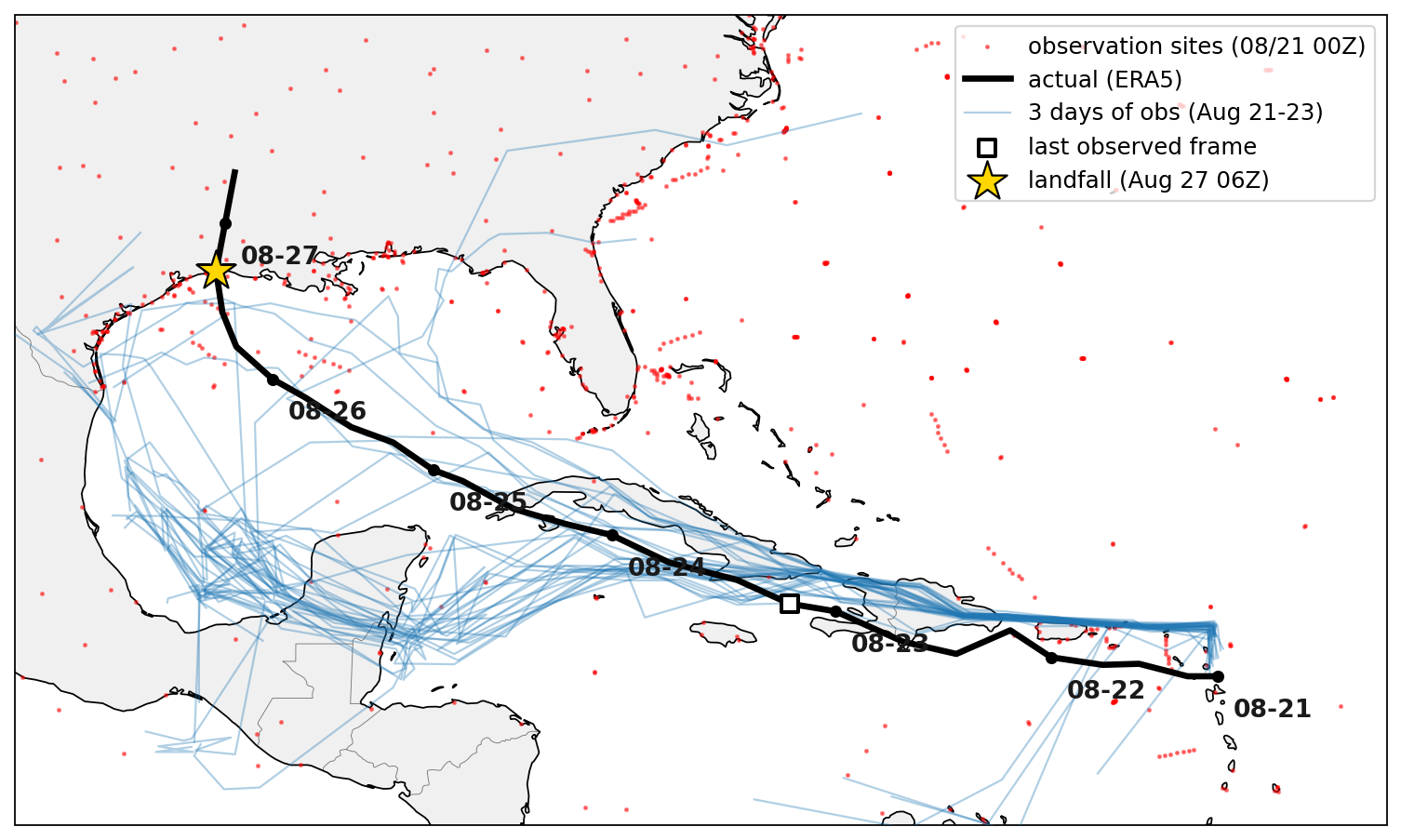}
\caption{Hurricane Laura case study: the window starts Aug 21 00Z and the first three days are observed as Laura crosses the Lesser
Antilles, Puerto Rico, and Hispaniola (last observed frame: open square, Aug 23 18Z); everything afterward is forecast. Thirty-two member trajectories are shown (light blue), and red dots are the observation sites at the first analysis time.}
\label{fig:laura_genesis}
\end{figure}

\begin{figure}[!ht]
\centering
\includegraphics[width=0.72\linewidth]{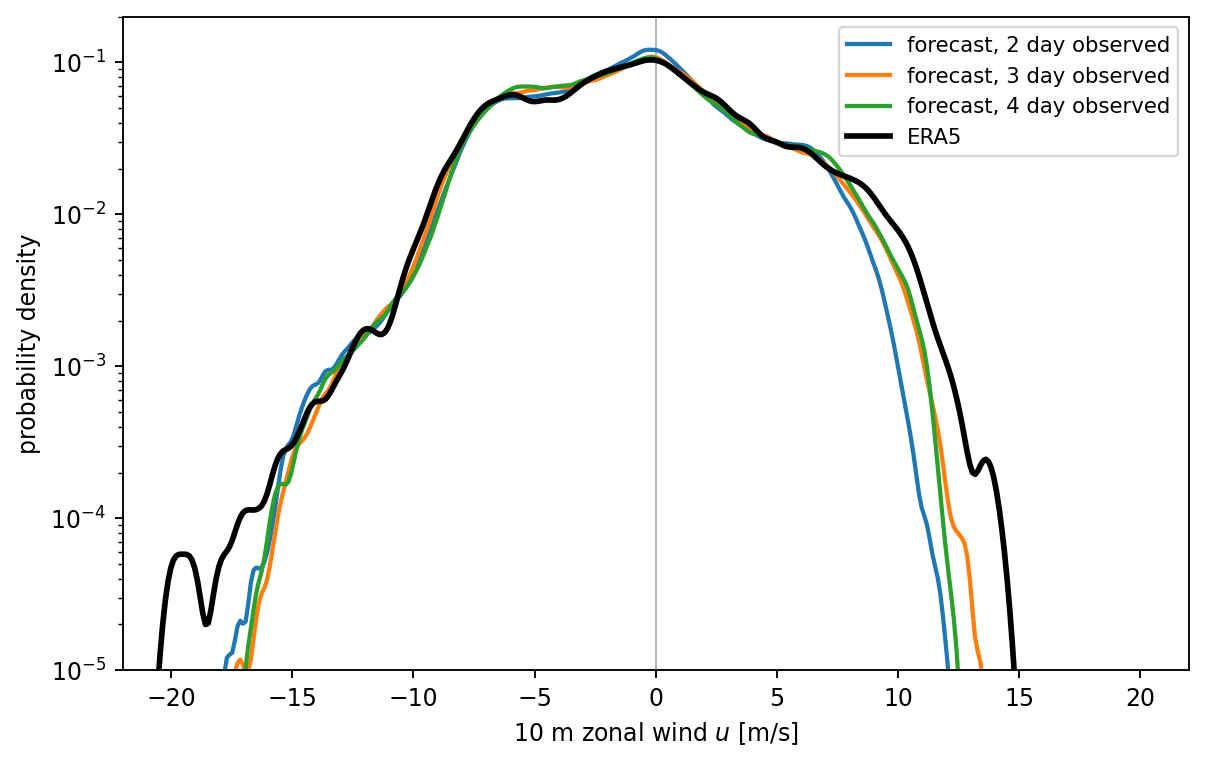}
\caption{Distribution of the $10$\,m zonal wind $u$ near the Laura region: ERA5 (black) against the posterior ensemble for two, three, and four days of observations.}
\label{fig:laura_windpdf}
\end{figure}

\paragraph{Hurricane tracking.}
We finally conduct a case study of Hurricane Laura (in 2020), one of the strongest hurricanes on record to make landfall in the U.S. Starting the window at Aug 23 06Z, Laura passes Hispaniola on its way to Cuba. We condition on the
real in-situ observations ($\mathrm{ISD}\cup\mathrm{IGRA}\cup\mathrm{ICOADS}$ union, exactly the same as the forecast configuration above) over observed windows of increasing length, two, three, and four days of observations, and let the frozen prior forecast the remainder up to when landfall occurs on Aug 27 06Z at Cameron, Louisiana. Figure~\ref{fig:laura} shows sixteen ensemble trajectories for the three cases against ERA5. We observe that the trajectory tightens accordingly, with the forecast with 4 observed days being the closest and showing the least spread. The lower spread here does not directly mean greater accuracy, but that the forecasts become more confident. Firstly, at $1.40625^\circ$ the grid cannot resolve a hurricane. Secondly, there are only a few observations over the Gulf. However, some of the ensemble members track the hurricane accurately. The distribution of $10$\,m wind inside the hurricane region (Fig.~\ref{fig:laura_windpdf}) shows that the posterior matches ERA5 except the extreme tails. We also repeat the experiment from the storm's \emph{early life}: a window starting Aug 21 00Z, observing the three days in which Laura crosses the Lesser Antilles, Puerto Rico, and Hispaniola, followed by a 32-member ensemble forecast in which landfall is a $3.5$-day-lead prediction
(Fig.~\ref{fig:laura_genesis}). The forecast seems to carry the storm into the Caribbean and then spreads, some entering the Gulf. Both these experiments show promising results where our learned prior and posterior sampling methodology can be used for direct obs-to-forecast, given an ideal observation set and the related operator is available.

\begin{figure}[!htbp]
\centering
\includegraphics[width=0.95\linewidth]{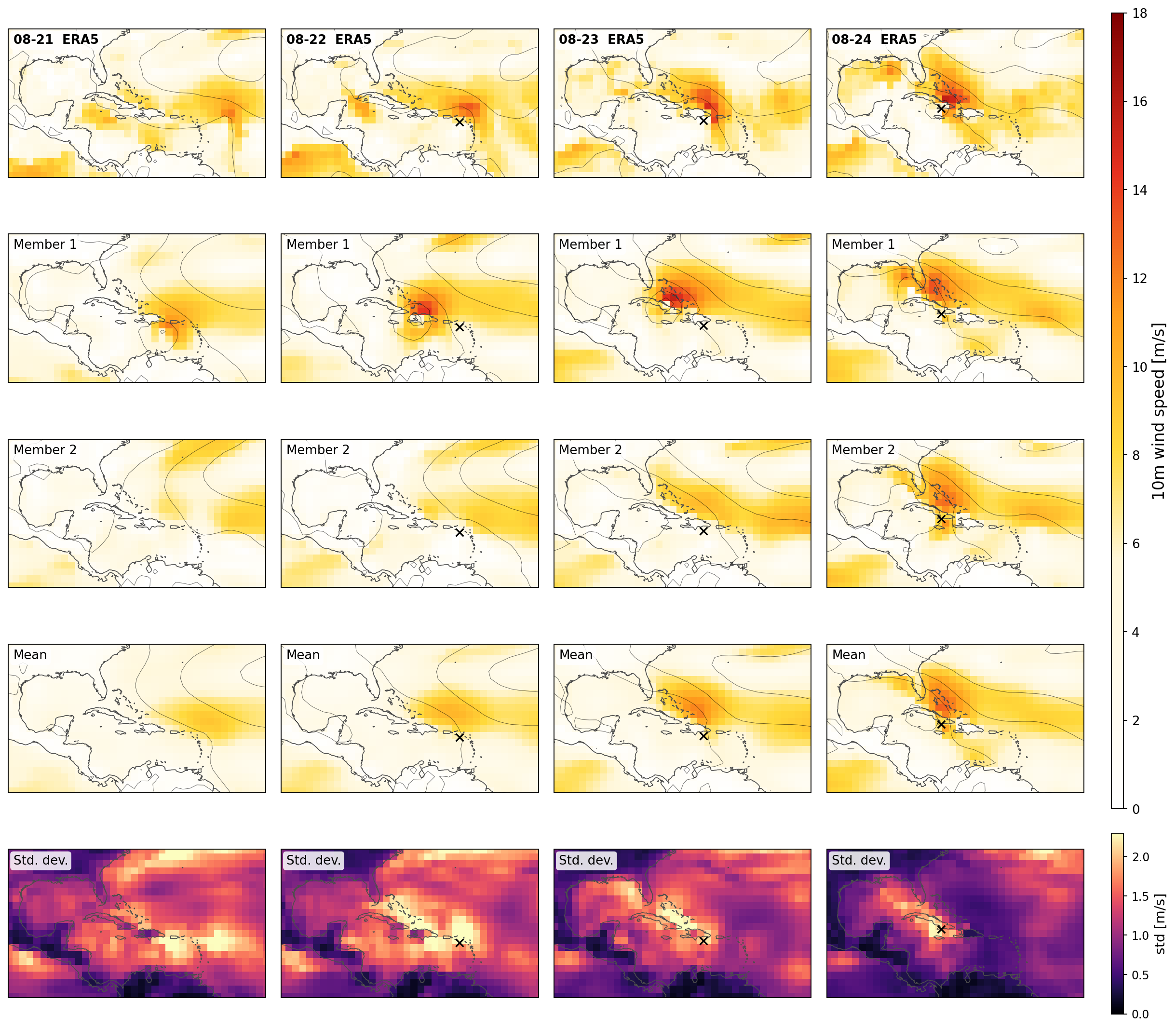}
\caption{Hindcast of Hurricane Laura ($10$\,m wind speed): ERA5 (top row), two individual posterior members, the ensemble mean, and the ensemble spread (standard deviation across members, bottom row) on the four inferred days Aug 21-24 00Z. Black contours are each field's own MSLP; the $\times$ marks the storm position.}
\label{fig:hindcast_wind}
\end{figure}

\begin{figure}[!htbp]
\centering
\includegraphics[width=0.95\linewidth]{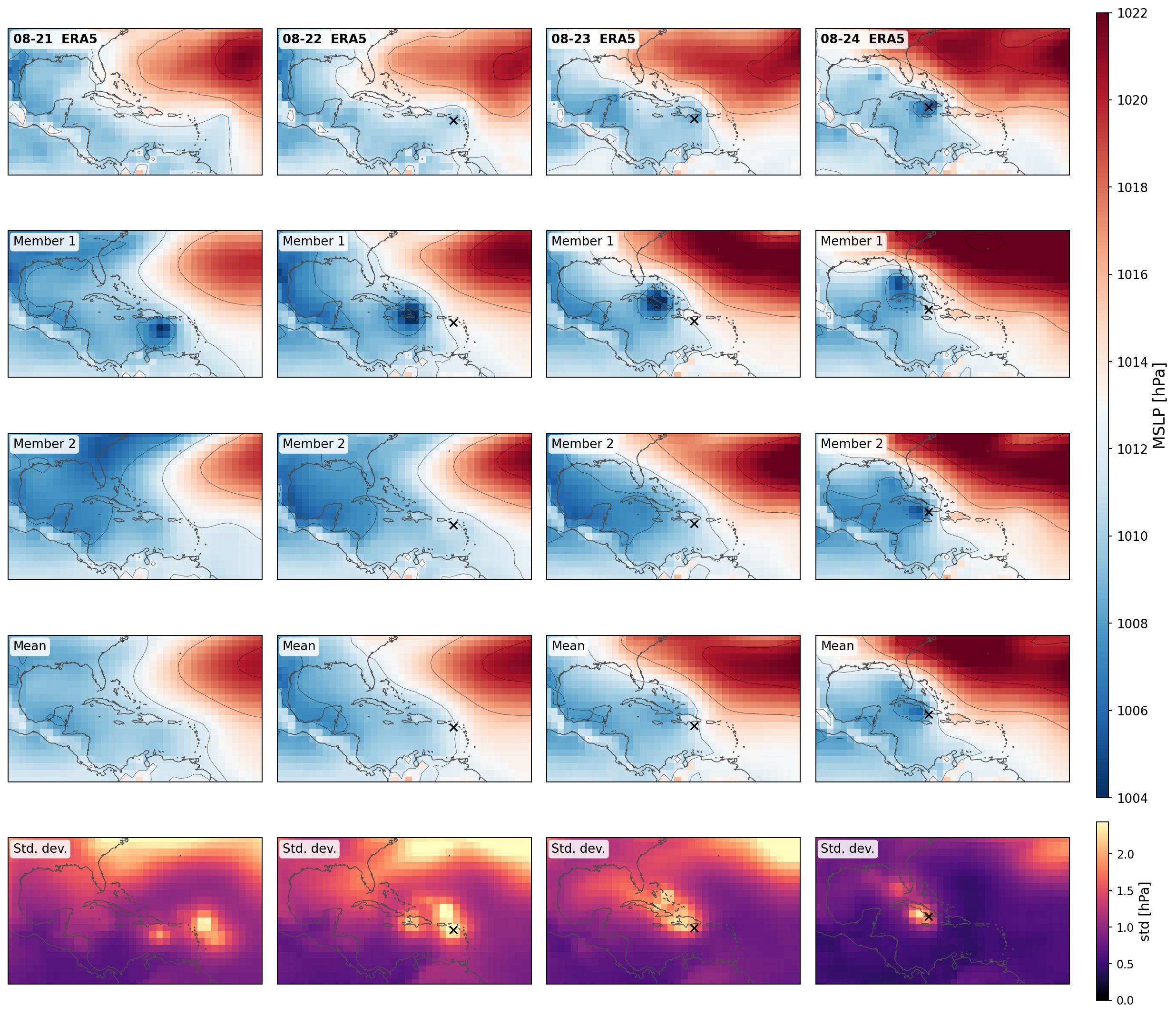}
\caption{Hindcast of Hurricane Laura's precursor (mean sea-level pressure). Individual members develop a closed low that deepens and tracks with the storm's position ($\times$) as Aug 24 approaches, recovering the broad structure of ERA5's developing low.}
\label{fig:hindcast_mslp}
\end{figure}

\begin{figure}[!ht]
\centering
\includegraphics[width=0.72\linewidth]{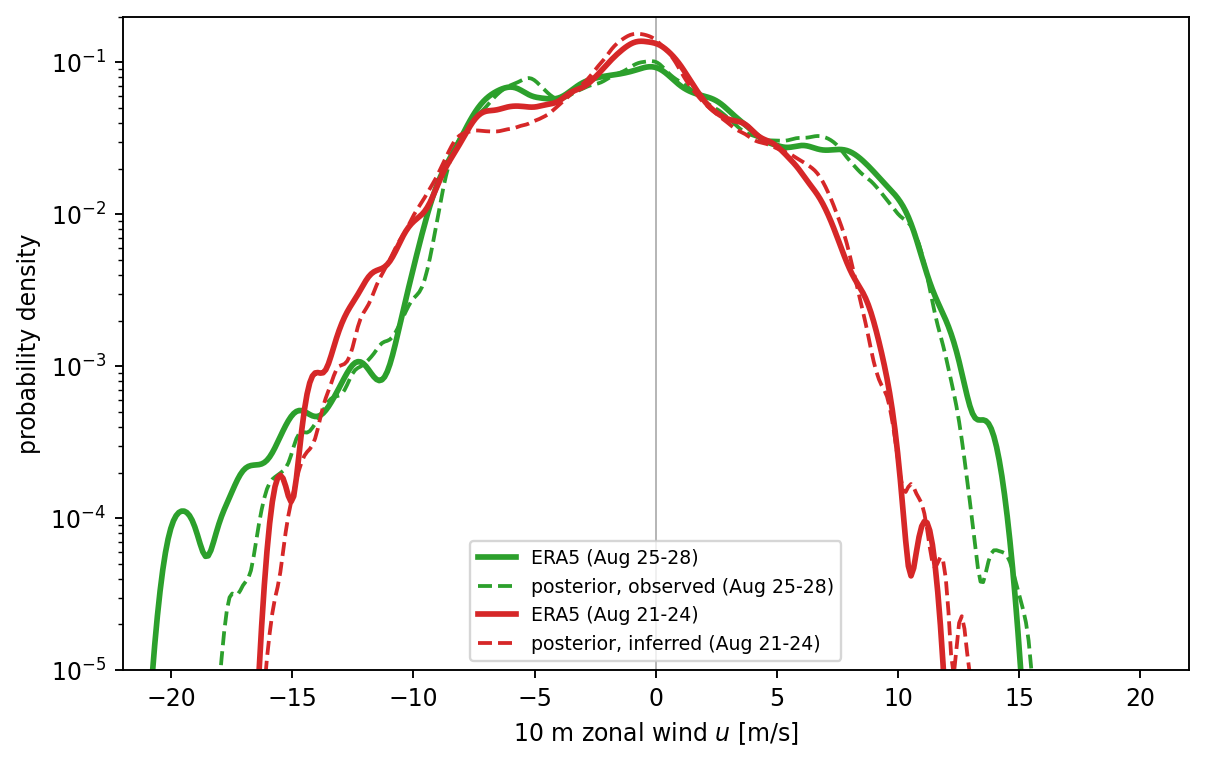}
\caption{Distribution of the $10$\,m zonal wind $u$ inside the box for the hindcast, split into the observed segment (Aug 25-28, green) and the inferred precursor (Aug 21-24, red); ERA5 solid, posterior dashed.}
\label{fig:hindcast_windpdf}
\end{figure}

\paragraph{Hindcast.}
The same framework also runs in reverse. We take the Laura window starting Aug 21 00Z, but now observe the real in-situ observations only over the last four days (Aug 25-28), when the storm runs from maturity to landfall, and hindcast using the prior to infer the first four days (Aug 21-24). This is the initial state that grew into the observed hurricane. We plot individual posterior members, their ensemble mean, and the ensemble spread along with ERA5 (Figs.~\ref{fig:hindcast_wind} and~\ref{fig:hindcast_mslp}). The figures show a coherent structure moving westward across the Caribbean. Information from the observed mature storm propagates backwards in time through the prior to constrain its unobserved genesis, with increasing errors backwards in time. The wind distributions (Fig.~\ref{fig:hindcast_windpdf}) here reproduce the full range of ERA5's $10$\,m wind, as expected when a past state is reconstructed rather than forecasting an extreme event.

\section{Discussion}
In this work, we train a latent generative \textit{video} prior of ERA5 and perform spatio-temporal data assimilation. We perform several experiments starting from simplified super-resolution to data fusion using real observations such as IGRA radiosondes, ISD surface stations, and the ICOADS ocean-atmosphere dataset. We also show that our generated samples are physically consistent in the limited tests that we have performed. Moreover, because the prior is a video, the methodology supports the classical data assimilation experiments without requiring an autoregressive forecast model: observing leading frames yields a causal filter whose error grows forward in time, and observing the middle yields a two-sided smoother. The same posterior sampling framework also yields a direct observation-to-forecast pipeline. Driven only by in-situ observations, with no learned encoder for observations and no separate forecast model, it can produce a prediction competitive with other trained obs-to-forecast systems like GraphDOP \cite{alexe2024graphdop}. We also explore different sampling algorithms, some of which improve efficiency by distributing the sampling budget towards lower noise levels, and others that scale the gradients and noise injection for more accuracy and calibrated spread.

Future work includes extending our sequential cycling across consecutive windows for long-term sub-seasonal forecasts. Moreover, an operator that can inject several observations, especially from satellites, and convert them to the ERA5 grid (like Aardvark \cite{allen2025end} or HealDA \cite{gupta2026healda}) can be used to further improve our pipeline. We present a preliminary experiment using Aardvark in Appendix~\ref{app:aardvark}. Another future direction is to directly include these observations as input to the flow model, with the option to exclude them when unavailable. 

\section*{Acknowledgments}
This research used resources of the Argonne Leadership Computing Facility, a U.S.
Department of Energy Office of Science User Facility operated under contract
DE-AC02-06CH11357. DC and RM acknowledge support from the DOE ASCR award
``Inertial neural surrogates for stable dynamical prediction.'' DC and RM also
acknowledge computing resources through an AI for Science allocation at the National
Energy Research Scientific Computing Center (NERSC) and from the Penn State Institute
for Computational and Data Sciences (ICDS). RM acknowledges the support of a YIP from
the ARO Modeling of Complex Systems program (PM: Robert Martin). We also acknowledge Melissa Adrian for the useful discussions related to this work.

\clearpage
\bibliographystyle{unsrt}
\bibliography{refs}

\clearpage
\appendix

\newcommand{\appitem}[1]{\noindent\textbf{\ref{#1}}\enspace\nameref{#1}\ \dotfill\ \pageref{#1}\par\medskip}
\newcommand{\appsubitem}[1]{\noindent\hspace{1.8em}\ref{#1}\enspace\nameref{#1}\ \dotfill\ \pageref{#1}\par\smallskip}
\section*{Appendix contents}
\appitem{sec:samplers}
\appsubitem{sec:math}
\appsubitem{app:eff}
\appsubitem{sec:improved}
\appsubitem{app:like}
\appsubitem{app:lit}
\appsubitem{app:ibdetail}
\appitem{app:training}
\appitem{app:superob}
\appitem{app:aardvark}
\appitem{app:extended}
\appsubitem{app:phys}
\appsubitem{app:allvars}
\appsubitem{app:station_scatter}

\section{Exploring different sampling techniques}
\label{sec:samplers}
\label{app:ablations}
This appendix collects the design rationale and the sampling study behind the recommended configurations of Table~\ref{tab:config}. Sec.~\ref{sec:math} justifies the central design choices of the guided sampler; the remaining subsections report the experiments. In every experiment the prior is fixed, every run is scored with the same latitude-weighted physical RMSE as the main results, and each method is tested on both observation
types: the dense structured SR grid and the sparse, irregular station observations. Unless stated otherwise, the reported rows are 8-member ensembles over the 8 validation windows disjoint from the test set. Conceptually, the sampler has two independent directions to explore: where the NFE budget is spent (Sec.~\ref{app:eff}) and how the computed gradient and injected noise are used (Sec.~\ref{sec:improved}). Sec.~\ref{app:like} then reports the likelihood-weight sweeps behind the settings used in the paper, and Secs.~\ref{app:lit} and~\ref{app:ibdetail} benchmark existing samplers in literature on the same tasks.

\subsection{Sampler design choices}
\label{sec:math}
\label{sec:des-choices}
We justify the central design choices here.

\paragraph{(a) Low-$\sigma$ corrector helps.}
At high $\sigma$ the rotation mostly transports the prior and observation $\mathbf{y}$ barely constrains the decoded $\xz$. The low-$\sigma$ Langevin corrector targets the posterior score and recovers measurement fidelity only in the low-$\sigma$ band where it can be determined.

\paragraph{(b) Lazy reuse fails.}
DPS requires $\nabla_{\xt}\log p(\mathbf{y}\mid\xt)=(\partial\xz/\partial\xt)^\top \nabla_{\xz}\log p(\mathbf{y}\mid\xz)$. From \eqref{eq:tweedie} the TrigFlow denoiser Jacobian is
\begin{equation}
\frac{\partial\xz}{\partial\xt}=\cos(t)\,I-\sin(t)\,\sd\,\frac{\partial F}{\partial\xt}.
\label{eq:jac}
\end{equation}
Through \eqref{eq:jac} the likelihood gradient depends on $\cos(t)$, $\sin(t)$, and
the changing $\xz$. Therefore, holding it fixed across steps and reusing injects a stale direction which hampers performance in our test cases.

\paragraph{(c) Effect of momentum.}
The per-step likelihood (through a decoder) gradient can be a noisy estimate of the misfit gradient. Momentum averages successive estimates, so the guidance direction has lower variance, similar to several optimizers. We observe empirically in our test cases that it does not help when the Langevin corrector is present: the corrector already re-equilibrates the state at each level, so the accumulated gradient direction adds little. We leave further exploration of the effect of momentum on highly non-linear measurement operators to future work.

\paragraph{(d) DSG normalization.}
The per-timestep coverage of station measurements swings widely (for IGRA it is dense at $00$/$12$Z and sparse at $06$/$18$Z). Therefore, the raw gradient \emph{magnitude} varies erratically and DSG sphere-normalization is preferred.

\paragraph{(e) Second-order update of Algorithm~\ref{alg:sampler}.}
Substituting $\sd F_s=(\cos s\,\xt-\xz)/\sin s$ (from Eq.~\eqref{eq:tweedie}) into the first-order rotation gives the exact exponential-integrator form
\begin{equation}
\mathbf{z}^{(1)}=\frac{\sin t}{\sin s}\,\xt+\frac{\sin\delta}{\sin s}\,\xz,
\end{equation}
the trigonometric analogue of DDIM: the step would be exact if $\xz$ were constant along it. The multistep second-order scheme of DPM-Solver++(2M) \cite{lu2025dpm} replaces $\xz$ by its linear extrapolation in $\log\sigma=\log\tan$ using the previous estimate, $\xz\to\xz+\tfrac{1}{2r}(\hat{\mathbf{z}}_0^{\text{prev}}-\xz)$ with $r$ the log-$\sigma$ step ratio defined in Algorithm~\ref{alg:sampler} (negative under that sign convention). Inserting this yields the correction term $\tfrac{\sin\delta}{2r\sin s}(\hat{\mathbf{z}}_0^{\text{prev}}-\xz)$ of Algorithm~\ref{alg:sampler}.

\subsection{Efficiency: distributing the NFE budget}
\label{app:eff}
\label{app:srdetail}
\begin{figure}[!ht]
\centering
\includegraphics[width=0.95\linewidth]{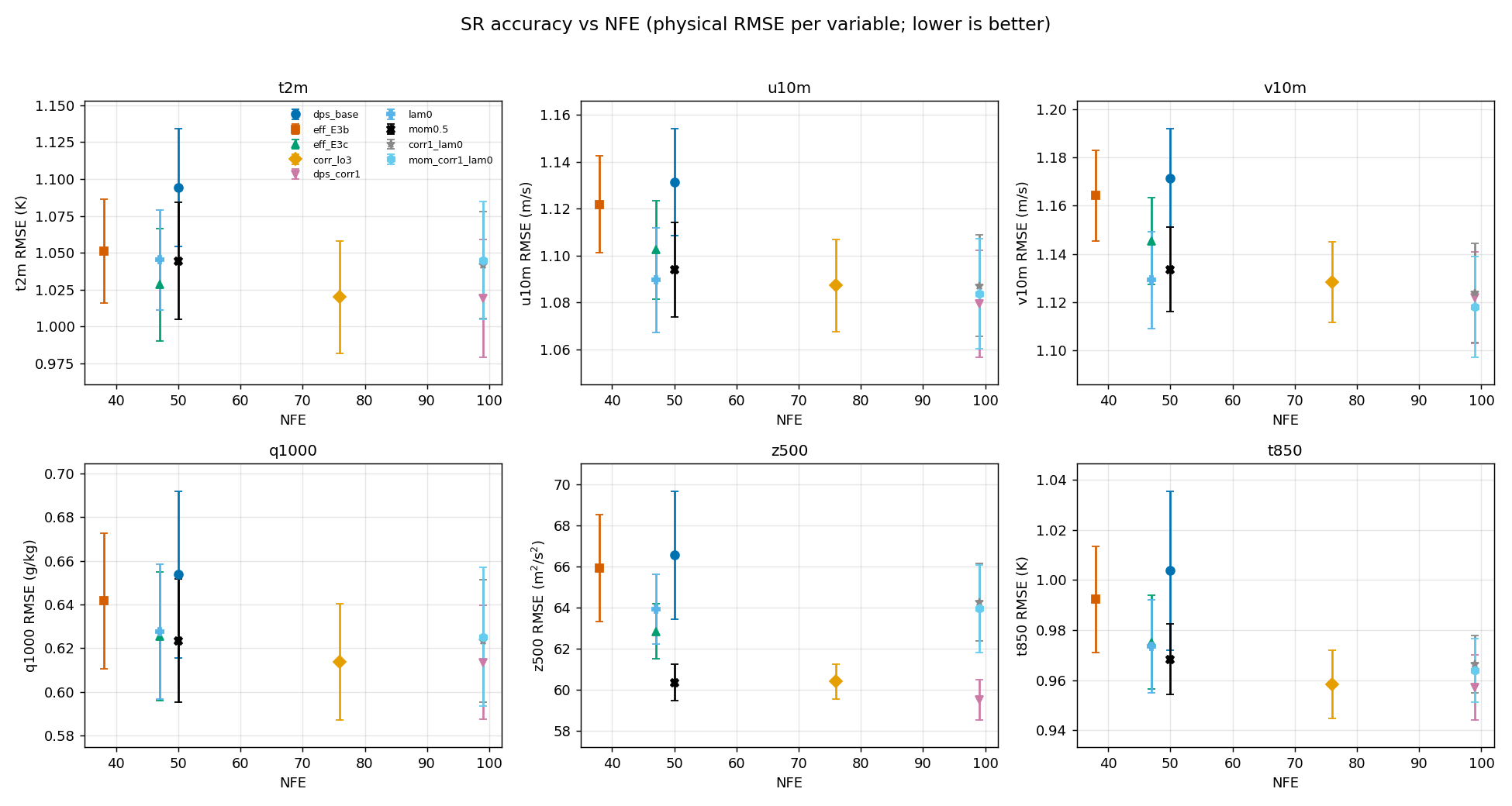}
\caption{SR sampler efficiency: SR physical RMSE per variable (six panels: t2m,
u10m, v10m, q1000, z500, t850) vs.\ NFE; lower is better. Error bars are $\pm1$
standard deviation.}
\label{fig:sr_errorbar}
\end{figure}

\begin{figure}[!ht]
\centering
\includegraphics[width=0.95\linewidth]{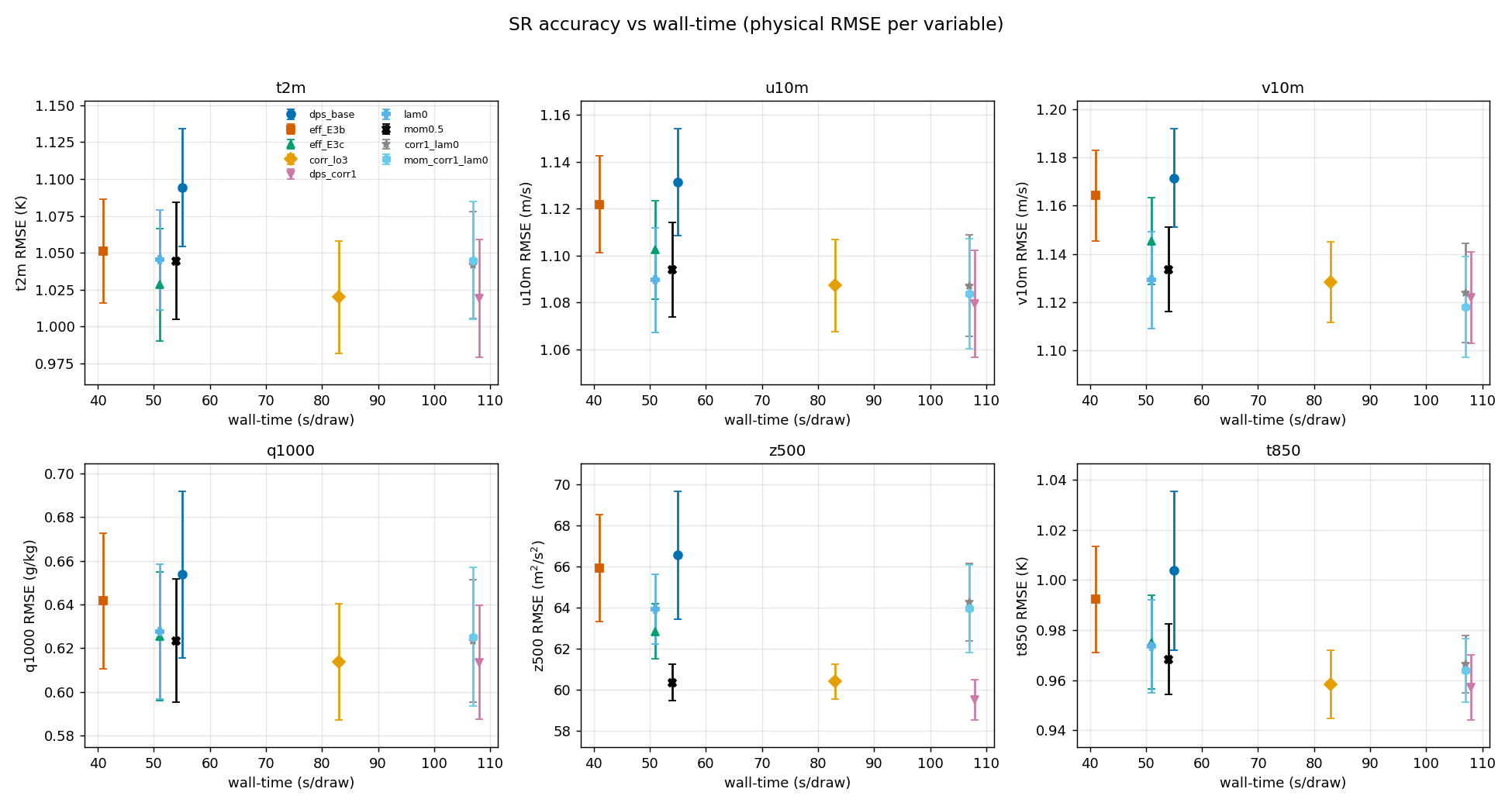}
\caption{Physical RMSE per variable (six panels: t2m, u10m, v10m, q1000, z500, t850)
vs.\ wall-time per draw; markers are the 8-window mean with $\pm1$ standard-deviation
error bars.}
\label{fig:sr_pareto}
\end{figure}

We observe that using NFEs strategically by using the corrector steps at low noise levels helps, as per Table~\ref{tab:multisample} and Fig.~\ref{fig:sr_errorbar}. On SR,
the reduced-step \textit{DPS+corr\,(N25)} matches the 50-step DPS baseline with fewer steps and less
wall time, and adding more corrector steps improves accuracy consistently. The IGRA radiosondes show a similar ordering
(Table~\ref{tab:igra}), except that the 50-step corrector configuration concentrated at low noise performs best. Restricting guidance to late steps \cite{kynkaanniemi2024applying} degrades results the more guidance is removed, and reusing gradients across steps hurts accuracy (Appendix~\ref{sec:math}b). An SDEdit warm start \cite{meng2021sdedit} matches the baseline but does not beat it. The only acceleration that actually helps is the NFE reallocation above. Wall-times per draw are listed alongside NFE in Tables~\ref{tab:multisample} and~\ref{tab:igra}; Fig.~\ref{fig:sr_pareto} shows the SR accuracy against wall time instead of NFE.

\begin{table*}[!htbp]
\centering
\caption{SR sampler ablations: per-variable latitude-weighted \emph{physical} RMSE. Units: t2m K,
u10m/v10m m/s, q1000 g/kg, z500 m$^2$/s$^2$. Bicubic = bicubic upsampling of the observed frames; AE floor = Autoencoder reconstruction error. Bold = best, underline = second-best, per variable.}
\label{tab:multisample}
\small
\setlength{\tabcolsep}{4pt}
\begin{tabular}{lrrccccc}
\toprule
method & NFE & wall (s) & t2m & u10m & v10m & q1000 & z500 \\
\midrule
DPS & 50 & 123 & 1.09$\pm$0.04 & 1.13$\pm$0.02 & 1.17$\pm$0.02 & 0.65$\pm$0.04 & 66.6$\pm$3.1 \\
+corr(all\,$\sigma$) & 99 & 243 & \textbf{1.02$\pm$0.04} & \textbf{1.08$\pm$0.02} & \textbf{1.12$\pm$0.02} & \textbf{0.61$\pm$0.03} & \textbf{59.5$\pm$1.0} \\
+corr & 76 & 186 & \underline{1.02$\pm$0.04} & \underline{1.09$\pm$0.02} & \underline{1.13$\pm$0.02} & \underline{0.61$\pm$0.03} & \underline{60.4$\pm$0.9} \\
+corr\,(N30) & 47 & 116 & 1.03$\pm$0.04 & 1.10$\pm$0.02 & 1.15$\pm$0.02 & 0.63$\pm$0.03 & 62.8$\pm$1.3 \\
+corr\,(N25) & 38 & 94 & 1.05$\pm$0.04 & 1.12$\pm$0.02 & 1.16$\pm$0.02 & 0.64$\pm$0.03 & 65.9$\pm$2.6 \\
\midrule
bicubic (no prior) & - & - & 5.64$\pm$0.30 & 4.29$\pm$0.10 & 3.97$\pm$0.14 & 2.44$\pm$0.09 & 794.7$\pm$30.0 \\
AE floor & - & - & 0.84$\pm$0.03 & 0.53$\pm$0.01 & 0.48$\pm$0.01 & 0.42$\pm$0.02 & 37.0$\pm$1.9 \\
\bottomrule
\end{tabular}
\end{table*}

\begin{table*}[!htbp]
\centering
\caption{IGRA data-fusion sampler ablations: per-variable physical RMSE (mean$\pm$std; units as Table~\ref{tab:multisample}; bold = best, underline = second-best).}
\label{tab:igra}
\footnotesize
\setlength{\tabcolsep}{4pt}
\begin{tabular}{lrrccccc}
\toprule
method & NFE & wall (s) & t2m & u10m & v10m & q1000 & z500 \\
\midrule
DPS (50 steps) & 50 & 142 & \underline{2.23$\pm$0.21} & \underline{3.32$\pm$0.11} & \underline{3.49$\pm$0.14} & \underline{1.25$\pm$0.03} & 567.5$\pm$29.6 \\
+corr (50 st, corr $\sigma{<}3$) & 76 & 215 & \textbf{2.16$\pm$0.20} & \textbf{3.30$\pm$0.11} & \textbf{3.47$\pm$0.14} & \textbf{1.23$\pm$0.03} & \textbf{560.7$\pm$30.5} \\
+corr\,(N30) (30 st, corr $\sigma{<}5$) & 47 & 134 & 2.27$\pm$0.25 & \underline{3.32$\pm$0.11} & \underline{3.49$\pm$0.14} & 1.27$\pm$0.03 & \underline{567.4$\pm$22.4} \\
+corr\,(N25) (25 st, corr $\sigma{<}3$) & 38 & 109 & 2.33$\pm$0.26 & 3.35$\pm$0.11 & 3.51$\pm$0.14 & 1.28$\pm$0.03 & 573.1$\pm$22.5 \\
\bottomrule
\end{tabular}
\end{table*}

\subsection{Reshaping gradients and noise}
\label{sec:improved}
Three modifications change how the sampler uses the computed gradient and noise (Sec.~\ref{sec:design-choices}). First, on the sparse-station tasks, use $\lambda{=}1$ when the ensemble
spread itself matters. On SR task, the $\lambda{=}0$ shows worse performance (Tables~\ref{tab:improved_sr},~\ref{tab:improved_igra}). Second, DSG normalization \cite{yang2024guidance} shows significant performance for station data because the misfit gradient varies with the per-frame report schedule and DSG normalization is needed to keep the updates stable. DSG outperforms on the sensor measurements on most of the variables (Table~\ref{tab:obs_bakeoff}). Third,
momentum guidance is safe only at small decay; it is roughly neutral up to $m{=}0.3$ and degrades beyond that. All headline configurations keep $m{=}0$. We hypothesize that it may be helpful for highly non-linear or noisy operators.

\begin{table}[!ht]
\centering
\caption{SR sampler noise, normalization, and momentum ablations: per-variable physical RMSE (8-window mean of
8-member ensembles; units as Table~\ref{tab:multisample}; bold = best, underline = second-best per variable). This
table is complementary to Table~\ref{tab:multisample}, which gives the efficiency ablations.}
\label{tab:improved_sr}
\footnotesize
\setlength{\tabcolsep}{4pt}
\begin{tabular}{lrccccc}
\toprule
method & NFE & t2m & u10m & v10m & q1000 & z500 \\
\midrule
DPS & 50 & 1.09$\pm$0.04 & 1.13$\pm$0.02 & 1.17$\pm$0.02 & 0.65$\pm$0.04 & 66.6$\pm$3.1 \\
+corr\,(N30)+$\lambda$0 (low-NFE) & 47 & 1.05$\pm$0.03 & 1.09$\pm$0.02 & 1.13$\pm$0.02 & 0.63$\pm$0.03 & 63.9$\pm$1.7 \\
+mom0.5 ($\lambda{=}1$) & 50 & 1.04$\pm$0.04 & 1.09$\pm$0.02 & 1.13$\pm$0.02 & 0.62$\pm$0.03 & \underline{60.3$\pm$0.9} \\
+corr(all\,$\sigma$) (prior best) & 99 & \textbf{1.02$\pm$0.04} & \textbf{1.08$\pm$0.02} & \underline{1.12$\pm$0.02} & \textbf{0.61$\pm$0.03} & \textbf{59.5$\pm$1.0} \\
+corr(all\,$\sigma$)+$\lambda$0 & 99 & \underline{1.04$\pm$0.04} & 1.09$\pm$0.02 & 1.12$\pm$0.02 & \underline{0.62$\pm$0.03} & 64.3$\pm$1.9 \\
+corr(all\,$\sigma$)+$\lambda$0+mom & 99 & 1.04$\pm$0.04 & \underline{1.08$\pm$0.02} & \textbf{1.12$\pm$0.02} & 0.63$\pm$0.03 & 64.0$\pm$2.1 \\
\midrule
AE floor & - & 0.84$\pm$0.03 & 0.53$\pm$0.01 & 0.48$\pm$0.01 & 0.42$\pm$0.02 & 37.0$\pm$1.9 \\
\bottomrule
\end{tabular}
\end{table}

\begin{table}[!ht]
\centering
\caption{IGRA sampler noise, normalization, and momentum ablations: per-variable physical RMSE (units as Table~\ref{tab:multisample}; bold = best, underline = second-best). This
table is complementary to Table~\ref{tab:igra}, which gives the efficiency ablations.}
\label{tab:improved_igra}
\footnotesize
\setlength{\tabcolsep}{4pt}
\begin{tabular}{lrccccc}
\toprule
method & NFE & t2m & u10m & v10m & q1000 & z500 \\
\midrule
DPS+corr+DSG+$\lambda$0 (scale 0.5) & 76 & \textbf{1.90$\pm$0.13} & \textbf{3.01$\pm$0.17} & \textbf{3.08$\pm$0.18} & \underline{1.15$\pm$0.04} & \textbf{474.9$\pm$50.4} \\
DPS+corr+DSG ($\lambda{=}1$, scale 0.5) & 76 & \underline{1.91$\pm$0.11} & \underline{3.03$\pm$0.17} & \underline{3.11$\pm$0.20} & \textbf{1.15$\pm$0.03} & \underline{480.0$\pm$50.8} \\
DPS+corr+$\lambda$0+mom & 76 & 2.07$\pm$0.19 & 3.23$\pm$0.12 & 3.39$\pm$0.14 & 1.21$\pm$0.02 & 539.4$\pm$31.8 \\
DPS+corr+$\lambda$0 & 76 & 2.14$\pm$0.21 & 3.29$\pm$0.11 & 3.45$\pm$0.14 & 1.23$\pm$0.02 & 561.5$\pm$32.3 \\
DPS+corr  & 76 & 2.16$\pm$0.20 & 3.30$\pm$0.11 & 3.47$\pm$0.14 & 1.23$\pm$0.03 & 560.7$\pm$30.5 \\
\bottomrule
\end{tabular}
\end{table}

\begin{table}[!ht]
\centering
\caption{Sampler ablations on the ISD and joint (IGRA+ISD) observations (per-variable physical RMSE, units as Table~\ref{tab:multisample}; bold = best, underline = second-best). Full set in Appendix~\ref{app:phys}.}
\label{tab:obs_bakeoff}
\footnotesize
\setlength{\tabcolsep}{4pt}
\begin{tabular}{llccccc}
\toprule
Observation & method & t2m & u10m & v10m & q1000 & z500 \\
\midrule
ISD (surface) & DPS & \textbf{2.10$\pm$0.08} & 3.39$\pm$0.12 & 3.55$\pm$0.19 & 1.44$\pm$0.06 & \textbf{641.1$\pm$34.9} \\
~ & +corr+DSG & \underline{2.15$\pm$0.10} & \underline{3.36$\pm$0.19} & \underline{3.35$\pm$0.22} & \textbf{1.39$\pm$0.09} & \underline{669.7$\pm$59.8} \\
~ & +corr+DSG+$\lambda$0 & 2.21$\pm$0.11 & \textbf{3.34$\pm$0.18} & \textbf{3.33$\pm$0.20} & \underline{1.41$\pm$0.09} & 701.3$\pm$65.8 \\
\midrule
Joint (both) & DPS & 2.09$\pm$0.17 & 3.27$\pm$0.11 & 3.43$\pm$0.13 & 1.23$\pm$0.02 & 557.3$\pm$27.4 \\
~ & +corr+DSG & \underline{1.81$\pm$0.10} & \underline{2.95$\pm$0.14} & \underline{3.01$\pm$0.17} & \underline{1.13$\pm$0.04} & \textbf{467.2$\pm$52.8} \\
~ & +corr+DSG+$\lambda$0 & \textbf{1.80$\pm$0.11} & \textbf{2.93$\pm$0.14} & \textbf{3.00$\pm$0.17} & \textbf{1.13$\pm$0.04} & \underline{467.6$\pm$48.5} \\
\bottomrule
\end{tabular}
\end{table}

\subsection{Likelihood weighting}
\label{app:like}
Table~\ref{tab:abl_sr_like} sweeps the SR likelihood weight (guidance scale and the
annealing coefficient $\gamma$) for the base DPS sampler. Table~\ref{tab:abl_station}
contrasts DPS guidance at three scales against DSG-normalized guidance across six
scales on the IGRA network. SR uses scale~4.0, $\gamma{=}0.1$ and all station tasks use DSG normalization at scale~0.5. The structured SR misfit is \emph{summed} over the observed pixels, whereas the station misfit is \emph{averaged} over points, channels, and observed frames. Under the averaged reduction, the guidance scale acts as a per-channel weight, so it is re-selected whenever the observed channel set changes, and the variable-splitting baselines require the same retuning.

\begin{table}[!ht]
\centering
\caption{SR likelihood-weight ablation (base DPS, 50 steps, 4 tuning windows disjoint from the test set): per-variable
physical RMSE (units as Table~\ref{tab:multisample}; bold = best within each sweep).
Guidance scale swept at $\gamma{=}0.1$; $\gamma$ swept at scale~2.5. The scale trend
flattens near $3$-$3.5$; the production SR runs use scale $4.0$, $\gamma{=}0.1$.}
\label{tab:abl_sr_like}
\footnotesize
\setlength{\tabcolsep}{4pt}
\begin{tabular}{lccccc}
\toprule
setting & t2m & u10m & v10m & q1000 & z500 \\
\midrule
\multicolumn{6}{l}{\emph{guidance scale} ($\gamma{=}0.1$)} \\
scale 1.5 & 1.78$\pm$1.02 & 1.57$\pm$0.09 & 1.63$\pm$0.08 & 0.82$\pm$0.09 & 97.8$\pm$5.3 \\
scale 2.0 & 1.39$\pm$0.40 & 1.48$\pm$0.05 & 1.55$\pm$0.05 & 0.77$\pm$0.05 & 90.6$\pm$5.9 \\
scale 2.5 & 1.14$\pm$0.02 & 1.41$\pm$0.01 & 1.48$\pm$0.02 & 0.74$\pm$0.02 & 83.3$\pm$2.1 \\
scale 3.0 & \textbf{1.13$\pm$0.02} & 1.39$\pm$0.02 & 1.45$\pm$0.02 & \textbf{0.73$\pm$0.01} & 80.7$\pm$1.8 \\
scale 3.5 & 1.14$\pm$0.01 & 1.37$\pm$0.01 & 1.44$\pm$0.02 & 0.73$\pm$0.02 & \textbf{79.4$\pm$2.1} \\
scale 4.0 & 1.15$\pm$0.02 & \textbf{1.36$\pm$0.02} & \textbf{1.43$\pm$0.02} & 0.74$\pm$0.02 & 79.5$\pm$2.4 \\
\midrule
\multicolumn{6}{l}{\emph{annealing} $\gamma$ (scale 2.5)} \\
$\gamma{=}10^{-2}$ & 1.80$\pm$0.09 & 1.60$\pm$0.03 & 1.65$\pm$0.03 & 0.93$\pm$0.03 & 110.0$\pm$6.2 \\
$\gamma{=}3\times10^{-2}$ & 1.37$\pm$0.07 & \textbf{1.38$\pm$0.02} & \textbf{1.44$\pm$0.03} & 0.81$\pm$0.03 & 86.2$\pm$3.9 \\
$\gamma{=}10^{-1}$ & \textbf{1.14$\pm$0.02} & 1.41$\pm$0.01 & 1.48$\pm$0.02 & \textbf{0.74$\pm$0.02} & \textbf{83.3$\pm$2.1} \\
$\gamma{=}3\times10^{-1}$ & 1.76$\pm$0.55 & 1.76$\pm$0.09 & 1.83$\pm$0.08 & 0.87$\pm$0.05 & 129.2$\pm$18.9 \\
\bottomrule
\end{tabular}
\end{table}

\begin{table}[!ht]
\centering
\caption{Station likelihood ablation on IGRA (4 validation windows, single seed): DPS guidance scale vs.\
DSG-normalized guidance, per-variable physical RMSE (units as
Table~\ref{tab:multisample}; bold = best).}
\label{tab:abl_station}
\small
\begin{tabular}{lccccc}
\toprule
setting & t2m & u10m & v10m & q1000 & z500 \\
\midrule
DPS (scale 3) & 3.06$\pm$0.45 & 4.52$\pm$0.20 & 4.83$\pm$0.21 & 1.63$\pm$0.04 & 834.1$\pm$59.6 \\
DPS (scale 5) & 2.73$\pm$0.34 & 4.38$\pm$0.16 & 4.68$\pm$0.17 & 1.57$\pm$0.04 & 773.1$\pm$59.2 \\
DPS (scale 7) & 3.31$\pm$0.62 & 4.44$\pm$0.13 & 4.63$\pm$0.09 & 1.63$\pm$0.10 & 791.8$\pm$28.9 \\
DSG scale 0.05 & 3.06$\pm$0.41 & 4.35$\pm$0.22 & 4.53$\pm$0.15 & 1.66$\pm$0.22 & 793.5$\pm$63.1 \\
DSG scale 0.15 & 2.91$\pm$0.47 & 4.04$\pm$0.22 & 4.13$\pm$0.17 & 1.52$\pm$0.21 & 687.1$\pm$66.0 \\
DSG scale 0.3 & 2.44$\pm$0.13 & 3.94$\pm$0.30 & 4.10$\pm$0.27 & 1.42$\pm$0.13 & 674.2$\pm$114.9 \\
DSG scale 0.5 & \textbf{2.43$\pm$0.09} & \textbf{3.83$\pm$0.25} & 3.92$\pm$0.29 & 1.40$\pm$0.11 & \textbf{626.9$\pm$79.3} \\
DSG scale 0.7 & 2.64$\pm$0.26 & 3.93$\pm$0.28 & 4.01$\pm$0.34 & 1.41$\pm$0.14 & 701.5$\pm$135.8 \\
DSG scale 1.0 & 2.90$\pm$0.45 & 4.09$\pm$0.44 & \textbf{3.83$\pm$0.20} & 1.48$\pm$0.12 & 717.7$\pm$163.7 \\
\bottomrule
\end{tabular}
\end{table}

\subsection{Comparison with existing samplers}
\label{app:lit}
We benchmark against existing inverse-problem samplers. We port them from their respective paper and tune them for our case, on both observation experiments in Table~\ref{tab:litbaselines}. The compared methods include the leading variable-splitting
methods of the \textsc{InverseBench} benchmark \cite{zheng2025inversebench}: \textit{DAPS} \cite{zhang2025improving} and \textit{PnP-DM} \cite{wu2024principled}, its most accurate methods on the closed-form operators, together with the
efficient \textit{DiffPIR} \cite{zhu2023denoising}, alongside RED-diff \cite{mardani2024variational}, STeP's HMC sampler \cite{zhang2025step}, guided Restart \cite{xu2023restart}, and a back-projection/DDNM port \cite{wang2022zero}. The table also has \emph{graded corrector}, our own ablation that adds a second batch of Langevin steps in a lowest-$\sigma$ band on top of the single-band corrector of Alg.~\ref{alg:effe3}. The two experiments show different outcomes. On the uniformly observed SR task, several of those methods are genuinely competitive with 8-member ensembles: guided Restart is the most accurate on the winds, and our ablated graded corrector on temperature and geopotential. RED-diff, \textit{DAPS}, and \textit{PnP-DM} show poor performance, the latter two at an order of magnitude more cost per draw, because their decoupled annealing runs an inner reverse-diffusion and Langevin chain at \emph{every} noise level. On the sparse radiosondes, in contrast, every other sampling technique performs inferior to our sampler on every variable, and the variable-splitting methods fail badly (we did not pursue some methods that showed clear inferiority). We conjecture that the decoupled-annealing robustness that makes these methods accurate on nonlinear closed-form operators offers little advantage for our smooth, differentiable operator, where gradient guidance is already stable.

A major issue for the station observations is how the data term is scaled. The strength hyperparameters used for
SR case are far too weak for the sub-steps of the variable-splitting samplers (Table~\ref{tab:abl_diffpir}). Once that strength is re-tuned for the station likelihood (\textit{DiffPIR} $\sigma_n{=}5\times10^{-4}$,\textit{PnP-DM} $\tau{=}10^{-4}$, matching the \textit{DAPS} retune), the gap shrinks considerably and \textit{DiffPIR} comes close to the guided samplers, but none of them beats our sampler on any variable. The remaining gap can be accounted for by the changes in the number of station locations from frame to frame, which keeps the data sub-step poorly scaled. The performance of these methods reflects our implementations rather than the methods in general.

\begin{table*}[!htbp]
\centering
\caption{Comparison with other existing samplers, on both observation geometries: per-variable physical RMSE
(8-window mean of 8-member ensemble means; units as Table~\ref{tab:multisample};
bold = best, underline = second-best per panel). NFE counts denoiser evaluations per draw; $^\dagger$
marks methods whose inner loops use additional compute.}
\label{tab:litbaselines}
\small
\setlength{\tabcolsep}{4pt}
\begin{tabular}{lcccccc}
\toprule
method & NFE & t2m & u10m & v10m & q1000 & z500 \\
\midrule
\multicolumn{7}{l}{\emph{Super-resolution (uniform space-time subsampling)}} \\
\midrule
RED-diff \cite{mardani2024variational} & 150 & 1.30$\pm$0.13 & 1.65$\pm$0.06 & 1.70$\pm$0.05 & 0.78$\pm$0.02 & 104.9$\pm$4.6 \\
STeP-HMC \cite{zhang2025step} & 30$^\dagger$ & 1.32$\pm$0.16 & 1.17$\pm$0.03 & 1.20$\pm$0.03 & 0.75$\pm$0.08 & 86.9$\pm$13.0 \\
guided Restart \cite{xu2023restart} & 85 & 1.05$\pm$0.04 & \textbf{1.07$\pm$0.02} & \textbf{1.10$\pm$0.02} & 0.63$\pm$0.03 & 66.1$\pm$2.5 \\
back-projection / DDNM \cite{wang2022zero} & 76 & 1.08$\pm$0.05 & 1.10$\pm$0.02 & 1.14$\pm$0.02 & \underline{0.63$\pm$0.03} & 69.3$\pm$2.8 \\
graded corrector (ours, ablated) & 71 & \textbf{1.02$\pm$0.04} & \underline{1.09$\pm$0.02} & 1.13$\pm$0.02 & \textbf{0.62$\pm$0.03} & \textbf{60.7$\pm$1.1} \\
DAPS \cite{zhang2025improving} & 120$^\dagger$ & 1.20$\pm$0.08 & 1.28$\pm$0.04 & 1.34$\pm$0.04 & 0.67$\pm$0.03 & 78.6$\pm$3.0 \\
DiffPIR \cite{zhu2023denoising} & 50 & 1.08$\pm$0.09 & 1.16$\pm$0.02 & 1.20$\pm$0.02 & 0.66$\pm$0.03 & 70.1$\pm$3.6 \\
PnP-DM \cite{wu2024principled} & 400$^\dagger$ & 1.56$\pm$0.34 & 1.50$\pm$0.06 & 1.53$\pm$0.06 & 0.77$\pm$0.04 & 101.2$\pm$7.8 \\
\emph{ours} (DPS+corr\,(N30)+$\lambda$0) & 47 & \underline{1.05$\pm$0.03} & 1.09$\pm$0.02 & \underline{1.13$\pm$0.02} & 0.63$\pm$0.03 & \underline{63.9$\pm$1.7} \\
\midrule
\multicolumn{7}{l}{\emph{IGRA radiosondes (real soundings)}} \\
\midrule
STeP-HMC \cite{zhang2025step} & 50$^\dagger$ & 2.35$\pm$0.23 & 3.43$\pm$0.10 & 3.60$\pm$0.14 & 1.29$\pm$0.03 & 613.9$\pm$29.8 \\
guided Restart \cite{xu2023restart} & 95 & \underline{2.25$\pm$0.20} & 3.38$\pm$0.10 & 3.55$\pm$0.14 & \underline{1.27$\pm$0.03} & \underline{590.2$\pm$28.4} \\
graded corrector (ours, ablated) & 110 & 2.27$\pm$0.21 & 3.40$\pm$0.10 & 3.57$\pm$0.14 & 1.27$\pm$0.03 & 597.1$\pm$29.3 \\
DAPS \cite{zhang2025improving} & 120$^\dagger$ & 3.43$\pm$0.61 & 3.86$\pm$0.14 & 3.94$\pm$0.13 & 1.57$\pm$0.11 & 767.9$\pm$69.1 \\
DiffPIR \cite{zhu2023denoising} & 50 & 2.53$\pm$0.29 & \underline{3.32$\pm$0.21} & \underline{3.39$\pm$0.18} & 1.28$\pm$0.06 & 600.0$\pm$95.5 \\
PnP-DM \cite{wu2024principled} & 400$^\dagger$ & 4.13$\pm$0.58 & 4.18$\pm$0.10 & 4.06$\pm$0.14 & 1.89$\pm$0.18 & 857.9$\pm$37.5 \\
\emph{ours} (DPS+corr+DSG+$\lambda$0) & 76 & \textbf{1.90$\pm$0.13} & \textbf{3.01$\pm$0.17} & \textbf{3.08$\pm$0.18} & \textbf{1.15$\pm$0.04} & \textbf{474.9$\pm$50.4} \\
\bottomrule
\end{tabular}
\end{table*}

\subsection{Variable-splitting samplers: further ablations}
\label{app:ibdetail}
Tables~\ref{tab:abl_daps_sr}-\ref{tab:abl_daps_igra} give further hyperparameter ablations behind the \textit{DAPS} rows of Table~\ref{tab:litbaselines}, and Table~\ref{tab:abl_diffpir} those for
\textit{DiffPIR} and \textit{PnP-DM}. The likelihood weight $\tau$ (the assumed measurement-noise scale in the data term of the inner Langevin) controls data-fitting strength. On SR a small $\tau$ performs well, and on the station data a much smaller $\tau$ recovers a usable performance. All \textit{DAPS} runs use $40$ annealing levels with $3$ inner reverse-diffusion
and $30$ Langevin steps each, chosen for reasonable compute in comparison to other methods. This is
roughly $1/8$ of the \textsc{InverseBench} default denoiser budget. Even at this budget \textit{DAPS}
costs an order of magnitude more per draw than our sampler.

\begin{table}[!ht]
\centering
\caption{\textit{DAPS} ablation on SR (8-window mean; 40 anneal steps, 3 inner
reverse-diffusion + 30 Langevin steps each): per-variable physical RMSE (units as
Table~\ref{tab:multisample}; bold = best).}
\label{tab:abl_daps_sr}
\small
\begin{tabular}{lccccc}
\toprule
setting & t2m & u10m & v10m & q1000 & z500 \\
\midrule
$\tau{=}0.02$ & 2.16$\pm$0.80 & 1.76$\pm$0.08 & 1.82$\pm$0.07 & 0.87$\pm$0.07 & 123.0$\pm$16.1 \\
$\tau{=}0.01$ & \textbf{1.20$\pm$0.08} & \textbf{1.28$\pm$0.04} & \textbf{1.34$\pm$0.04} & \textbf{0.67$\pm$0.03} & \textbf{78.6$\pm$3.0} \\
$\tau{=}0.01$ + $\lambda=0$ & 1.50$\pm$0.33 & 1.56$\pm$0.06 & 1.64$\pm$0.07 & 0.76$\pm$0.03 & 96.5$\pm$5.5 \\
$\tau{=}0.01$ + DSG (lr~$10^{-2}$) & 2.52$\pm$0.46 & 2.94$\pm$0.21 & 3.08$\pm$0.15 & 1.31$\pm$0.15 & 258.0$\pm$34.2 \\
$\tau{=}0.01$ + DSG + $(\lambda=0)$ & 2.48$\pm$0.32 & 2.92$\pm$0.22 & 3.03$\pm$0.17 & 1.31$\pm$0.13 & 250.6$\pm$21.7 \\
\bottomrule
\end{tabular}
\end{table}

\begin{table}[!ht]
\centering
\caption{\textit{DAPS} ablation on IGRA: per-variable
physical RMSE (units as Table~\ref{tab:multisample}; bold = best per column). The row carried into
the IGRA panel of Table~\ref{tab:litbaselines} is $\tau{=}10^{-4}$. The SR-tuned $\tau$
($0.005$-$0.01$) badly underfits (note the large t2m errors); $\tau{\approx}10^{-4}$
or the scale-invariant DSG step is needed. The $\tau\in\{0.01,0.005\}$ rows use
Langevin lr $2{\times}10^{-4}$ and the $\tau\in\{10^{-4},3{\times}10^{-5}\}$ rows lr
$10^{-4}$.}
\label{tab:abl_daps_igra}
\small
\begin{tabular}{lccccc}
\toprule
setting & t2m & u10m & v10m & q1000 & z500 \\
\midrule
$\tau{=}0.01$ & 7.93$\pm$3.17 & 5.69$\pm$0.28 & 5.92$\pm$0.27 & 3.13$\pm$0.97 & 1530.9$\pm$358.4 \\
$\tau{=}0.005$ & 8.84$\pm$4.17 & 5.62$\pm$0.39 & 5.77$\pm$0.29 & 3.39$\pm$1.25 & 1586.2$\pm$486.5 \\
$\tau{=}3\times10^{-5}$ & 10.22$\pm$1.03 & 5.32$\pm$0.31 & 4.23$\pm$0.15 & 2.51$\pm$0.11 & 2118.5$\pm$136.9 \\
$\tau{=}10^{-4}$ (8-member) & \textbf{3.43$\pm$0.61} & \textbf{3.86$\pm$0.14} & \textbf{3.94$\pm$0.13} & \textbf{1.57$\pm$0.11} & \textbf{767.9$\pm$69.1} \\
$\tau{=}10^{-4}$ + $(\lambda=0)$ & 4.24$\pm$0.85 & 4.97$\pm$0.15 & 5.13$\pm$0.25 & 2.12$\pm$0.29 & 988.9$\pm$100.0 \\
DSG (lr~$5\times10^{-3}$) & 5.75$\pm$1.64 & 5.37$\pm$0.43 & 5.50$\pm$0.30 & 2.52$\pm$0.59 & 1191.6$\pm$252.5 \\
DSG (lr~$2\times10^{-2}$) & 5.02$\pm$0.75 & 5.17$\pm$0.29 & 5.47$\pm$0.22 & 2.41$\pm$0.41 & 1064.7$\pm$118.1 \\
DSG (lr~$10^{-2}$) & 5.03$\pm$1.01 & 5.25$\pm$0.23 & 5.45$\pm$0.23 & 2.36$\pm$0.32 & 1118.8$\pm$127.9 \\
DSG (lr~$10^{-2}$) + $(\lambda=0)$ & 4.55$\pm$0.81 & 5.12$\pm$0.26 & 5.29$\pm$0.14 & 2.29$\pm$0.33 & 1032.7$\pm$123.5 \\
\bottomrule
\end{tabular}
\end{table}

We also perform ablation studies on other methods. Because the decoder operator is nonlinear, we replace DiffPIR's closed-form proximal data solve (Zhu et al.) by a single clamped gradient step. Its prior-trust weight $w$ is swept, where a
larger $w$ \emph{weakens} the data step since the step size is $\propto 1/w$
(data clamp $0.3$ at $w{=}100$, $0.2$ at $w{=}300$). DiffPIR runs $50$ steps
with $\xi{=}0.1$ and uses our stabilized re-noising variant, with the effective
noise computed from the pre-correction estimate. PnP-DM is the split-Gibbs sampler which uses a likelihood Langevin step anchored on the previous clean sample
(coupling $\rho$ set to the annealed $\sigma$) alternates with a stochastic reverse-flow prior step, at $40$ annealing levels with $10$ inner ancestral and $30$ Langevin steps each.

\begin{table}[!ht]
\centering
\caption{\textit{DiffPIR} and \textit{PnP-DM} ablation, per-variable physical RMSE
(units as Table~\ref{tab:multisample}). Top block: SR; bottom: IGRA.}
\label{tab:abl_diffpir}
\small
\begin{tabular}{lccccc}
\toprule
setting & t2m & u10m & v10m & q1000 & z500 \\
\midrule
\multicolumn{6}{l}{\emph{SR}} \\
DiffPIR ($w{=}100$) & \textbf{1.08$\pm$0.09} & \textbf{1.16$\pm$0.02} & \textbf{1.20$\pm$0.02} & \textbf{0.66$\pm$0.03} & \textbf{70.1$\pm$3.6} \\
DiffPIR ($w{=}300$) & 1.20$\pm$0.06 & 1.49$\pm$0.04 & 1.56$\pm$0.03 & 0.76$\pm$0.03 & 94.5$\pm$3.3 \\
PnP-DM ($\tau{=}0.02$) & 2.32$\pm$1.01 & 1.93$\pm$0.09 & 1.95$\pm$0.08 & 1.03$\pm$0.08 & 158.3$\pm$12.4 \\
PnP-DM ($\tau{=}0.01$) & 1.56$\pm$0.34 & 1.50$\pm$0.06 & 1.53$\pm$0.06 & 0.77$\pm$0.04 & 101.2$\pm$7.8 \\
PnP-DM ($\tau{=}0.005$) & 1.55$\pm$0.30 & 1.48$\pm$0.04 & 1.54$\pm$0.03 & 0.76$\pm$0.03 & 95.5$\pm$6.5 \\
\midrule
\multicolumn{6}{l}{\emph{IGRA}} \\
DiffPIR ($\sigma_n{=}5\times10^{-4}$, 8-member) & \textbf{2.53$\pm$0.29} & \textbf{3.32$\pm$0.21} & \textbf{3.39$\pm$0.18} & \textbf{1.28$\pm$0.06} & \textbf{600.0$\pm$95.5} \\
DiffPIR ($\sigma_n{=}5\times10^{-4}$, $w{=}300$) & 3.43$\pm$0.68 & 4.55$\pm$0.18 & 4.75$\pm$0.22 & 1.74$\pm$0.12 & 835.3$\pm$52.6 \\
DiffPIR ($\sigma_n{=}5\times10^{-3}$) & 3.48$\pm$0.82 & 4.72$\pm$0.21 & 4.84$\pm$0.27 & 1.86$\pm$0.19 & 877.2$\pm$63.8 \\
DiffPIR ($\sigma_n{=}0.05$) & 6.73$\pm$2.62 & 5.56$\pm$0.28 & 5.68$\pm$0.23 & 2.84$\pm$0.82 & 1360.5$\pm$285.7 \\
PnP-DM ($\tau{=}3\times10^{-5}$) & 14.56$\pm$1.87 & 7.51$\pm$0.58 & 4.69$\pm$0.23 & 3.39$\pm$0.29 & 2643.2$\pm$212.0 \\
PnP-DM ($\tau{=}10^{-4}$, 8-member) & 4.13$\pm$0.58 & 4.18$\pm$0.10 & 4.06$\pm$0.14 & 1.89$\pm$0.18 & 857.9$\pm$37.5 \\
PnP-DM ($\tau{=}0.005$) & 7.76$\pm$2.90 & 6.74$\pm$0.59 & 4.68$\pm$0.12 & 2.95$\pm$0.86 & 2210.0$\pm$283.1 \\
\bottomrule
\end{tabular}
\end{table}

\section{Model and training details}
\label{app:training}
Tables~\ref{tab:ae_details} and~\ref{tab:prior_details} collect the complete
architecture and training configuration of the two trained components, the
autoencoder and the TrigFlow prior; everything else in the paper runs zero-shot
on these frozen weights.

\begin{table}[!htbp]
\centering
\caption{Autoencoder: architecture and training details.}
\label{tab:ae_details}
\small
\setlength{\tabcolsep}{5pt}
\begin{tabular}{l l}
\toprule
\multicolumn{2}{l}{\emph{Architecture}} \\
\midrule

parameters & $451.1$\,M (encoder $188.6$\,M, decoder $262.4$\,M) \\
widths / blocks & $[96,192,384,768]$; $3$ residual blocks per level ($+1$ per decoder level) \\
level strides & $[(1,2,2),(2,2,2),(2,1,1)]$ \\
latent saturation & \textit{softclip2}, scale $10$ \\
surface conditioning & land-sea mask + orography, area-resampled per decoder level, \\
 & \quad injected through zero-initialized $3{\times}3$ convolutions \\
dropout & $0.05$ \\
\midrule
\multicolumn{2}{l}{\emph{Optimization}} \\
\midrule
objective & latitude-weighted, variance-normalized MSE (variance floor $10^{-2}$) \\
 & \quad $+\,0.05\times$(spatial-gradient $+$ Laplacian $+$ temporal-difference) \\
optimizer & Adam, $\beta{=}(0.9,0.95)$, weight decay $0$ \\
learning rate & $5\times10^{-4}$, cosine-annealed; gradient clip $5$ \\
batch & $=128$\\
latent regularizer & Gaussian latent noise, std $0.02$, before the decoder \\
epochs & $155$ \\
\midrule
\multicolumn{2}{l}{\emph{Data and hardware}} \\
\midrule
training data & 32-frame ERA5 windows, $1979$-$2018$ ($2019$ validation) \\
hardware & $16\times$ NVIDIA A100-40GB\\
training time & $\approx9$ days \\
\bottomrule
\end{tabular}
\end{table}

\begin{table}[!htbp]
\centering
\caption{TrigFlow prior (DiT3D): architecture and training details.}
\label{tab:prior_details}
\small
\setlength{\tabcolsep}{5pt}
\begin{tabular}{l l}
\toprule
\multicolumn{2}{l}{\emph{Architecture}} \\
\midrule
parameters & $525.4$\,M\\
input & standardized AE latents $(128,8,32,64)$ \\
tokenization & patch $(1,2,2)$ $\to$ $4096$ tokens \\
width / depth / heads & $1536$ / $12$ / $24$; MLP ratio $4$ \\
timestep embedding & sinusoidal, size $384$, input scale $1000$ \\
\midrule
\multicolumn{2}{l}{\emph{Optimization}} \\
\midrule
objective & TrigFlow velocity matching, $\sd{=}1$, $\ln\sigma\sim\mathcal N(0,1.5^2)$, \\
 & \quad adaptive log-variance weighting; $\sigma\in[0.002,80]$ \\
optimizer & AdamW, lr $2\times10^{-4}$, $\beta{=}(0.9,0.999)$, weight decay $0.01$; gradient clip $1$ \\
lr schedule & $500$-kimg linear ramp-up, floor at $0.05\times$ peak \\
batch size & $256$\\
EMA & half-life $500$\,kimg, ramp-up ratio $0.05$ \\
precision & bf16 autocast (TF32 matmul) \\
training length & $19{,}901$\,kilo images \\
\midrule
\multicolumn{2}{l}{\emph{Data and hardware}} \\
\midrule
training data & AE latents of the ERA5 training windows ($1979$-$2018$) \\
hardware & $8\times$ NVIDIA A100-40GB \\
Training time & $\approx5$ days \\
\bottomrule
\end{tabular}
\end{table}

\section{Grid-cell super-obbing and the observed-window length}\label{app:superob}

\begin{figure}[!ht]
\centering
\includegraphics[width=0.95\linewidth]{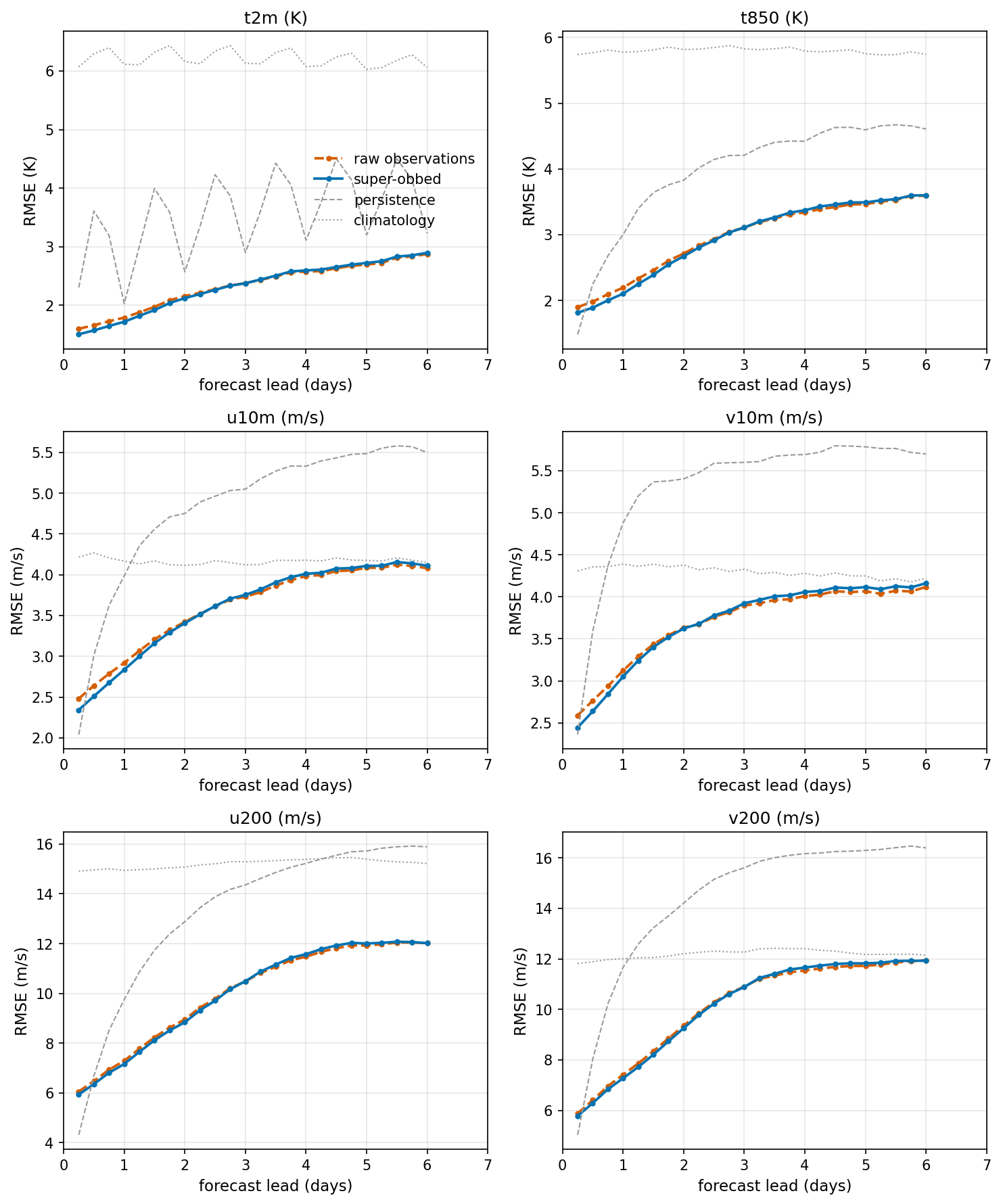}
\caption{Effect of grid-cell super-obbing on forecast skill: the raw clustered observations versus one super-observation per cell, both compared against ERA5 and averaged over the eight
test initial conditions (8-member ensembles). Gray reference lines mark persistence of the last observed frame (dashed) and the annual climatology (dotted).}
\label{fig:superob}
\end{figure}

Figure~\ref{fig:superob} shows the effect of the super-obbing step of Sec.~\ref{sec:forecast}.
This process collapses the clustered raw observations to one weighted super-observation per occupied grid cell. We observe that this lowers latitude-weighted forecast RMSE in the first forecast timesteps. Beyond day two the raw and super-obbed forecasts are essentially similar.
However, we use super-obbing in our results following previous DA works \cite{hoffman2018effect}.

\begin{figure}[!ht]
\centering
\includegraphics[width=0.95\linewidth]{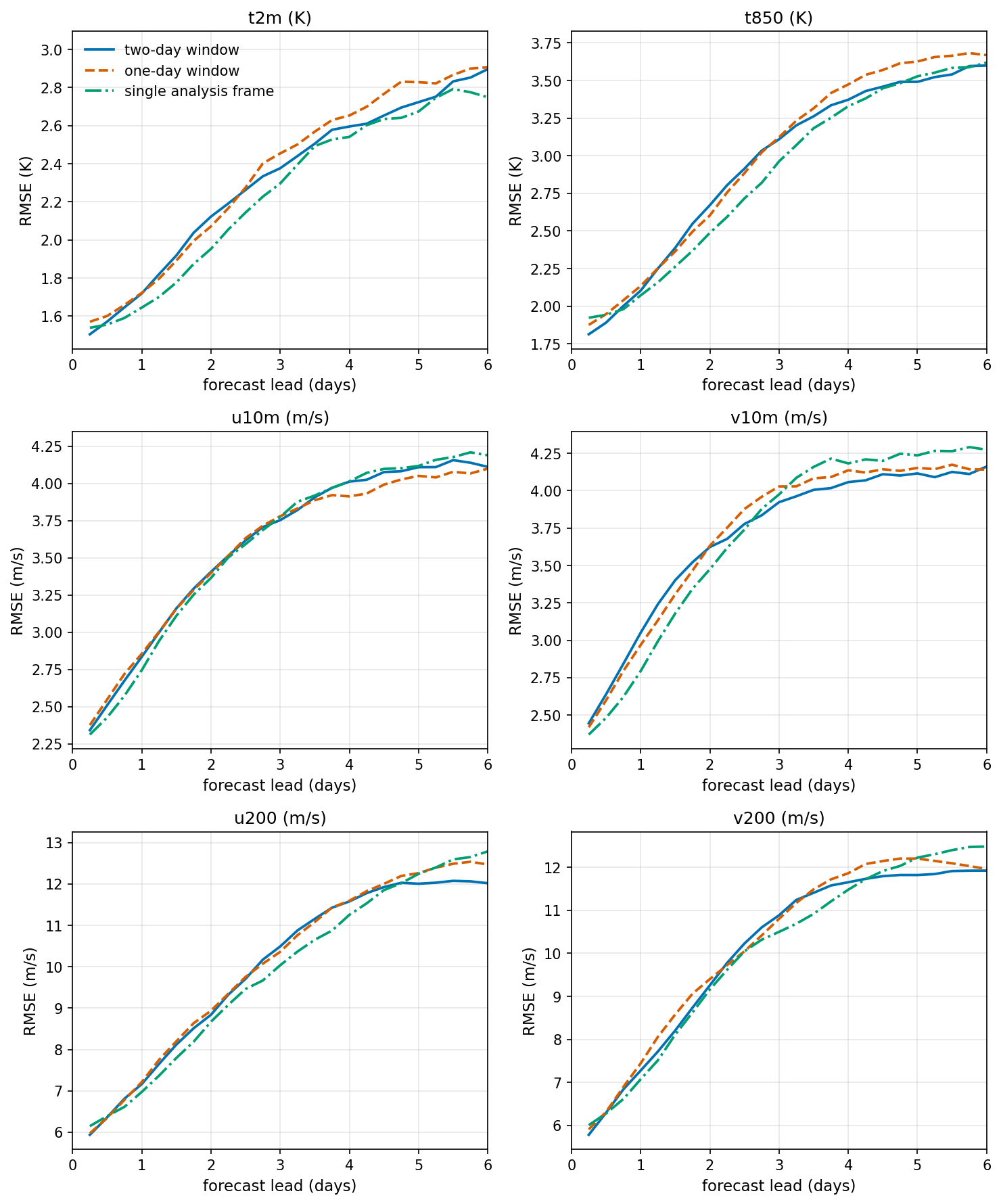}
\caption{Observed-window ablation for the in-situ obs$\to$forecast task of
Sec.~\ref{sec:forecast}: conditioning on a two-day window, a one-day window, and a single
analysis frame. The plots show Latitude-weighted RMSE against ERA5 versus lead.}
\label{fig:obs2forecast_windows}
\end{figure}

The length of the observed window does not drastically change the outcomes either
(Fig.~\ref{fig:obs2forecast_windows}). Shortening it from two days (eight frames) to one day leaves day-1 to day-3 skill essentially unchanged and even conditioning on a single analysis frame only slightly worsens the first forecast timestep. This result is consistent with previous work on AI based weather forecasting
\cite{lam2023learning}.

\section{Forecasting from the Aardvark latent observations}
\label{app:aardvark}
\begin{figure}[!ht]
\centering
\includegraphics[width=0.95\linewidth]{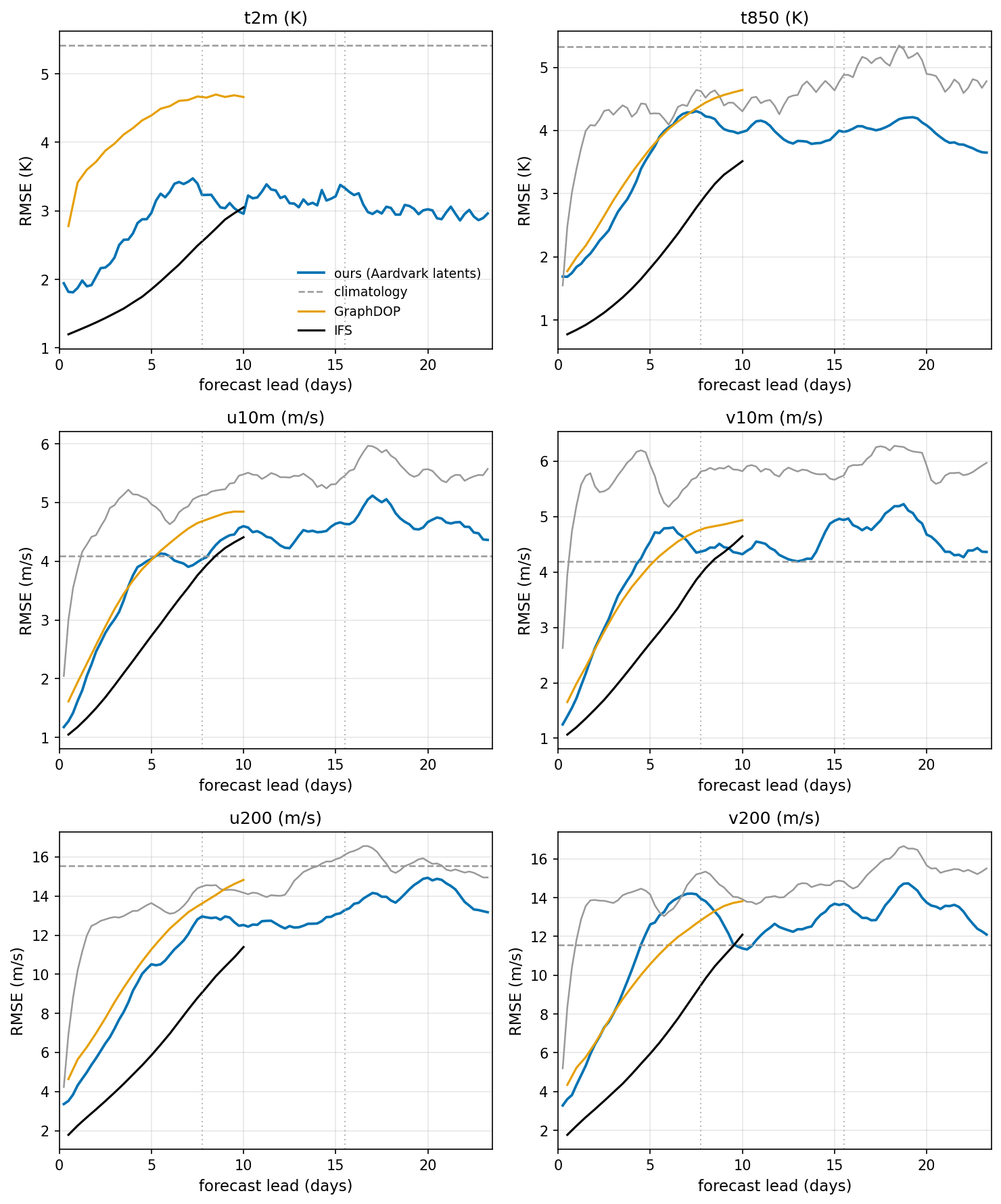}
\caption{Forecast using a single (train-period) latent observation of Aardvark (2018-08-18). GraphDOP and IFS as reported in \cite{alexe2024graphdop} through day 10. The plots show Latitude-weighted RMSE against ERA5 versus lead.}
\label{fig:aardvark_chain}
\end{figure}

Sec.~\ref{sec:forecast} uses only the fraction of the global observing system that our interpolation operator ingests directly. Especially fields at lower pressure levels are weak because a few hundred radiosonde measurements sample it sparsely and the satellite observations are excluded. To experiment with richer observations, we use the Aardvark latent observations \cite{allen2025end}, whose encoder maps
several observation modalities (including the satellite radiances, scatterometer, and geostationary imagery) onto a regular grid of 24 ERA5 variables. We use the Aardvark latent observation to forecast a $23.25$-day lead
(Fig.~\ref{fig:aardvark_chain}). This experiment is just an illustration, not a rigorous test case benchmark. The public Aardvark release contains only one observation data point, on 2018-08-18, (which falls within our training period). A held-out, multiple initial conditions test would require an extended observation-assembly pipeline, which we leave for future work. In this simplified experiment, we observe that the Aardvark latent lowers the forecast errors, with some variables performing better than GraphDOP.

\section{Extended results}
\label{app:extended}
This section expands the five-variable tables and the selected-variable figures of the main text across the full channel set: the complete per-variable RMSE tables for every experiment (Sec.~\ref{app:phys}), the per-channel super-resolution reconstructions (Sec.~\ref{app:allvars}), and the reconstruction-versus-observation scatter for every observed variable (Sec.~\ref{app:station_scatter}).

\subsection{Full per-variable physical RMSE}
\label{app:phys}
The main-text tables report five representative variables; here we give the
latitude-weighted physical RMSE of \emph{all} 69 output channels. Tables~\ref{tab:phys_surface}-\ref{tab:phys_specific} are divided by variables and pressure levels and columns have the different tasks: super-resolution and the IGRA, ISD, and joint measurements, each with its recommended sampler configuration from Table~\ref{tab:config} (mean$\pm$std).

\begin{table}[!htbp]\centering\footnotesize\setlength{\tabcolsep}{4pt}
\caption{Full per-variable physical RMSE, Surface (SR: \textit{DPS+corr\,(N30)+$\lambda$0}; stations: \textit{DPS+corr+DSG+$\lambda$0}).}
\label{tab:phys_surface}
\begin{tabular}{lcccc}\toprule
variable & SR & IGRA & ISD & Joint \\\midrule
2m temperature (K) & 1.05$\pm$0.03 & 1.90$\pm$0.13 & 2.21$\pm$0.11 & 1.80$\pm$0.11 \\
10m $u$-wind (m/s) & 1.09$\pm$0.02 & 3.01$\pm$0.17 & 3.34$\pm$0.18 & 2.93$\pm$0.14 \\
10m $v$-wind (m/s) & 1.13$\pm$0.02 & 3.08$\pm$0.18 & 3.33$\pm$0.20 & 3.00$\pm$0.17 \\
MSLP (Pa) & 77.6$\pm$2.1 & 469.3$\pm$44.4 & 492.8$\pm$43.1 & 447.7$\pm$38.1 \\
\bottomrule\end{tabular}\end{table}

\begin{table}[!htbp]\centering\footnotesize\setlength{\tabcolsep}{4pt}
\caption{Full per-variable physical RMSE, Geopotential (m$^2$/s$^2$) (SR: \textit{DPS+corr\,(N30)+$\lambda$0}; stations: \textit{DPS+corr+DSG+$\lambda$0}).}
\label{tab:phys_geopotential}
\begin{tabular}{lcccc}\toprule
variable & SR & IGRA & ISD & Joint \\\midrule
50\,hPa & 109.6$\pm$4.1 & 857.8$\pm$327.3 & 2461.9$\pm$466.1 & 839.2$\pm$319.1 \\
100\,hPa & 93.6$\pm$8.5 & 709.6$\pm$168.8 & 2202.2$\pm$490.6 & 703.0$\pm$164.4 \\
150\,hPa & 82.7$\pm$6.0 & 678.0$\pm$109.0 & 1969.2$\pm$407.9 & 670.2$\pm$109.8 \\
200\,hPa & 80.3$\pm$5.3 & 684.6$\pm$84.0 & 1721.5$\pm$319.8 & 676.1$\pm$83.1 \\
250\,hPa & 85.1$\pm$6.0 & 694.6$\pm$73.7 & 1495.6$\pm$237.1 & 685.8$\pm$71.4 \\
300\,hPa & 81.8$\pm$4.4 & 668.1$\pm$67.4 & 1287.7$\pm$179.8 & 659.5$\pm$64.6 \\
400\,hPa & 73.9$\pm$2.9 & 566.4$\pm$57.4 & 942.0$\pm$108.5 & 558.7$\pm$54.7 \\
500\,hPa & 63.9$\pm$1.7 & 474.9$\pm$50.4 & 701.3$\pm$65.8 & 467.6$\pm$48.5 \\
600\,hPa & 56.9$\pm$1.3 & 410.1$\pm$45.2 & 540.8$\pm$38.3 & 402.2$\pm$43.4 \\
700\,hPa & 54.7$\pm$1.7 & 370.7$\pm$40.5 & 443.5$\pm$25.5 & 361.2$\pm$38.3 \\
850\,hPa & 54.0$\pm$1.7 & 349.0$\pm$34.9 & 383.0$\pm$29.8 & 336.8$\pm$31.4 \\
925\,hPa & 56.4$\pm$1.9 & 355.6$\pm$34.0 & 383.3$\pm$32.8 & 341.9$\pm$29.8 \\
1000\,hPa & 61.0$\pm$1.8 & 369.2$\pm$34.5 & 392.9$\pm$34.7 & 353.5$\pm$29.9 \\
\bottomrule\end{tabular}\end{table}

\begin{table}[!htbp]\centering\footnotesize\setlength{\tabcolsep}{4pt}
\caption{Full per-variable physical RMSE, Zonal wind $u$ (m/s) (SR: \textit{DPS+corr\,(N30)+$\lambda$0}; stations: \textit{DPS+corr+DSG+$\lambda$0}).}
\label{tab:phys_zonal}
\begin{tabular}{lcccc}\toprule
variable & SR & IGRA & ISD & Joint \\\midrule
50\,hPa & 1.78$\pm$0.05 & 4.47$\pm$0.92 & 10.29$\pm$1.38 & 4.43$\pm$0.93 \\
100\,hPa & 2.10$\pm$0.06 & 4.91$\pm$0.25 & 7.59$\pm$0.73 & 4.85$\pm$0.26 \\
150\,hPa & 2.40$\pm$0.06 & 6.10$\pm$0.35 & 9.35$\pm$0.64 & 6.02$\pm$0.36 \\
200\,hPa & 2.57$\pm$0.05 & 7.46$\pm$0.47 & 10.60$\pm$0.59 & 7.35$\pm$0.45 \\
250\,hPa & 2.72$\pm$0.05 & 8.26$\pm$0.49 & 11.06$\pm$0.51 & 8.16$\pm$0.45 \\
300\,hPa & 2.73$\pm$0.05 & 8.19$\pm$0.42 & 10.56$\pm$0.46 & 8.10$\pm$0.40 \\
400\,hPa & 2.50$\pm$0.06 & 6.95$\pm$0.33 & 8.63$\pm$0.38 & 6.88$\pm$0.30 \\
500\,hPa & 2.19$\pm$0.04 & 5.74$\pm$0.29 & 6.96$\pm$0.34 & 5.68$\pm$0.27 \\
600\,hPa & 2.00$\pm$0.04 & 5.01$\pm$0.27 & 5.90$\pm$0.30 & 4.94$\pm$0.25 \\
700\,hPa & 1.87$\pm$0.04 & 4.57$\pm$0.24 & 5.31$\pm$0.28 & 4.50$\pm$0.22 \\
850\,hPa & 1.68$\pm$0.03 & 4.27$\pm$0.23 & 4.86$\pm$0.26 & 4.18$\pm$0.20 \\
925\,hPa & 1.54$\pm$0.04 & 4.28$\pm$0.22 & 4.79$\pm$0.25 & 4.18$\pm$0.18 \\
1000\,hPa & 1.20$\pm$0.03 & 3.33$\pm$0.19 & 3.75$\pm$0.22 & 3.25$\pm$0.16 \\
\bottomrule\end{tabular}\end{table}

\begin{table}[!htbp]\centering\footnotesize\setlength{\tabcolsep}{4pt}
\caption{Full per-variable physical RMSE, Meridional wind $v$ (m/s) (SR: \textit{DPS+corr\,(N30)+$\lambda$0}; stations: \textit{DPS+corr+DSG+$\lambda$0}).}
\label{tab:phys_meridional}
\begin{tabular}{lcccc}\toprule
variable & SR & IGRA & ISD & Joint \\\midrule
50\,hPa & 1.73$\pm$0.04 & 3.39$\pm$0.57 & 5.10$\pm$0.66 & 3.37$\pm$0.58 \\
100\,hPa & 2.07$\pm$0.07 & 4.34$\pm$0.29 & 5.76$\pm$0.13 & 4.30$\pm$0.26 \\
150\,hPa & 2.35$\pm$0.07 & 5.79$\pm$0.42 & 7.74$\pm$0.31 & 5.72$\pm$0.39 \\
200\,hPa & 2.53$\pm$0.07 & 7.37$\pm$0.57 & 9.57$\pm$0.44 & 7.24$\pm$0.55 \\
250\,hPa & 2.70$\pm$0.05 & 8.39$\pm$0.62 & 10.42$\pm$0.50 & 8.25$\pm$0.61 \\
300\,hPa & 2.73$\pm$0.04 & 8.38$\pm$0.53 & 10.07$\pm$0.38 & 8.25$\pm$0.52 \\
400\,hPa & 2.49$\pm$0.05 & 7.06$\pm$0.42 & 8.21$\pm$0.28 & 6.95$\pm$0.40 \\
500\,hPa & 2.17$\pm$0.04 & 5.77$\pm$0.39 & 6.60$\pm$0.27 & 5.67$\pm$0.37 \\
600\,hPa & 1.97$\pm$0.03 & 4.94$\pm$0.37 & 5.55$\pm$0.28 & 4.84$\pm$0.35 \\
700\,hPa & 1.84$\pm$0.03 & 4.42$\pm$0.32 & 4.89$\pm$0.26 & 4.33$\pm$0.30 \\
850\,hPa & 1.67$\pm$0.04 & 4.13$\pm$0.29 & 4.45$\pm$0.26 & 4.03$\pm$0.27 \\
925\,hPa & 1.57$\pm$0.04 & 4.29$\pm$0.27 & 4.62$\pm$0.27 & 4.18$\pm$0.26 \\
1000\,hPa & 1.25$\pm$0.02 & 3.44$\pm$0.20 & 3.75$\pm$0.25 & 3.35$\pm$0.20 \\
\bottomrule\end{tabular}\end{table}

\begin{table}[!htbp]\centering\footnotesize\setlength{\tabcolsep}{4pt}
\caption{Full per-variable physical RMSE, Temperature (K) (SR: \textit{DPS+corr\,(N30)+$\lambda$0}; stations: \textit{DPS+corr+DSG+$\lambda$0}).}
\label{tab:phys_temperature}
\begin{tabular}{lcccc}\toprule
variable & SR & IGRA & ISD & Joint \\\midrule
50\,hPa & 0.92$\pm$0.03 & 2.47$\pm$0.63 & 5.91$\pm$1.05 & 2.48$\pm$0.60 \\
100\,hPa & 0.86$\pm$0.03 & 2.29$\pm$0.41 & 4.60$\pm$0.68 & 2.26$\pm$0.43 \\
150\,hPa & 0.70$\pm$0.02 & 2.25$\pm$0.27 & 4.57$\pm$0.92 & 2.23$\pm$0.26 \\
200\,hPa & 0.72$\pm$0.02 & 2.68$\pm$0.24 & 5.53$\pm$0.98 & 2.65$\pm$0.21 \\
250\,hPa & 0.73$\pm$0.02 & 2.43$\pm$0.11 & 5.38$\pm$1.06 & 2.42$\pm$0.13 \\
300\,hPa & 0.71$\pm$0.02 & 2.15$\pm$0.14 & 5.03$\pm$0.92 & 2.12$\pm$0.15 \\
400\,hPa & 0.71$\pm$0.02 & 2.27$\pm$0.15 & 4.79$\pm$0.72 & 2.21$\pm$0.13 \\
500\,hPa & 0.72$\pm$0.01 & 2.26$\pm$0.14 & 4.27$\pm$0.54 & 2.20$\pm$0.10 \\
600\,hPa & 0.74$\pm$0.01 & 2.23$\pm$0.13 & 3.84$\pm$0.43 & 2.16$\pm$0.09 \\
700\,hPa & 0.77$\pm$0.01 & 2.19$\pm$0.14 & 3.49$\pm$0.37 & 2.12$\pm$0.10 \\
850\,hPa & 0.97$\pm$0.02 & 2.37$\pm$0.15 & 3.00$\pm$0.13 & 2.29$\pm$0.14 \\
925\,hPa & 0.93$\pm$0.03 & 2.19$\pm$0.18 & 2.62$\pm$0.13 & 2.09$\pm$0.15 \\
1000\,hPa & 0.93$\pm$0.03 & 1.96$\pm$0.17 & 2.31$\pm$0.11 & 1.86$\pm$0.14 \\
\bottomrule\end{tabular}\end{table}

\begin{table}[!htbp]\centering\footnotesize\setlength{\tabcolsep}{4pt}
\caption{Full per-variable physical RMSE, Specific humidity (g/kg) (SR: \textit{DPS+corr\,(N30)+$\lambda$0}; stations: \textit{DPS+corr+DSG+$\lambda$0}).}
\label{tab:phys_specific}
\begin{tabular}{lcccc}\toprule
variable & SR & IGRA & ISD & Joint \\\midrule
50\,hPa & 0.000$\pm$0.000 & 0.000$\pm$0.000 & 0.000$\pm$0.000 & 0.000$\pm$0.000 \\
100\,hPa & 0.000$\pm$0.000 & 0.001$\pm$0.000 & 0.001$\pm$0.000 & 0.001$\pm$0.000 \\
150\,hPa & 0.002$\pm$0.000 & 0.002$\pm$0.000 & 0.003$\pm$0.000 & 0.002$\pm$0.000 \\
200\,hPa & 0.010$\pm$0.001 & 0.014$\pm$0.001 & 0.019$\pm$0.002 & 0.014$\pm$0.001 \\
250\,hPa & 0.034$\pm$0.002 & 0.050$\pm$0.004 & 0.063$\pm$0.004 & 0.050$\pm$0.003 \\
300\,hPa & 0.078$\pm$0.003 & 0.118$\pm$0.008 & 0.148$\pm$0.008 & 0.117$\pm$0.008 \\
400\,hPa & 0.232$\pm$0.007 & 0.361$\pm$0.021 & 0.463$\pm$0.022 & 0.358$\pm$0.021 \\
500\,hPa & 0.447$\pm$0.014 & 0.710$\pm$0.022 & 0.886$\pm$0.043 & 0.703$\pm$0.022 \\
600\,hPa & 0.649$\pm$0.012 & 1.061$\pm$0.017 & 1.248$\pm$0.063 & 1.050$\pm$0.015 \\
700\,hPa & 0.873$\pm$0.022 & 1.419$\pm$0.048 & 1.645$\pm$0.087 & 1.402$\pm$0.044 \\
850\,hPa & 1.064$\pm$0.038 & 1.622$\pm$0.057 & 1.859$\pm$0.110 & 1.603$\pm$0.053 \\
925\,hPa & 0.794$\pm$0.027 & 1.220$\pm$0.055 & 1.497$\pm$0.112 & 1.206$\pm$0.054 \\
1000\,hPa & 0.627$\pm$0.031 & 1.151$\pm$0.036 & 1.411$\pm$0.094 & 1.132$\pm$0.037 \\
\bottomrule\end{tabular}\end{table}

\subsection{Per-channel super-resolution reconstructions}
\label{app:allvars}
\begin{figure}[p]\centering
\includegraphics[width=0.95\linewidth]{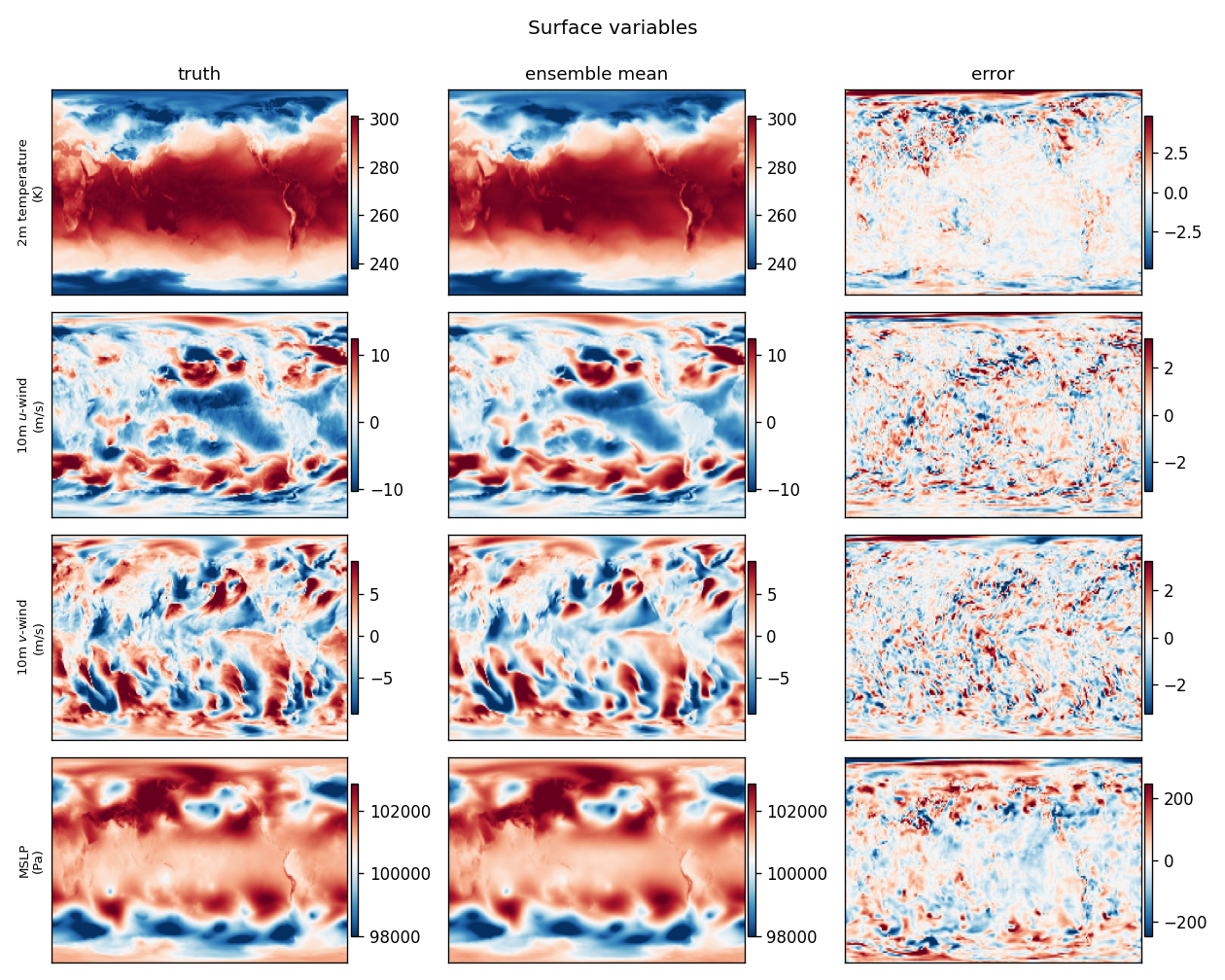}
\caption{Surface variables: ERA5 truth, 8-member ensemble mean, and signed error (mean $-$ truth; red positive, blue negative) for the super-resolution task. The following per-level figures use the same panel layout.}
\label{fig:app_surface}
\end{figure}

\begin{figure}[p]\centering
\includegraphics[width=0.95\linewidth]{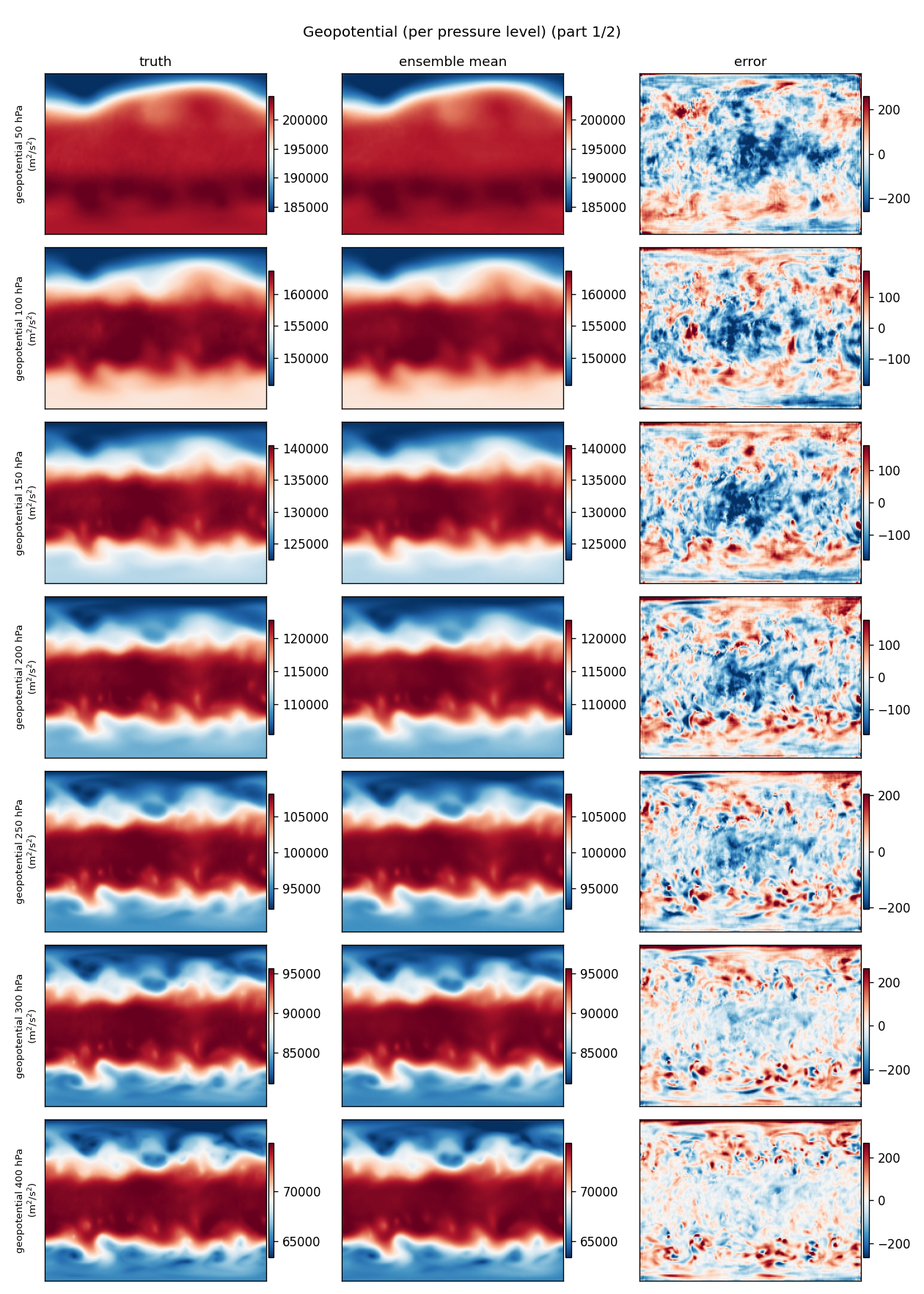}
\caption{Geopotential (per pressure level), part 1/2; same panel layout as Fig.~\ref{fig:app_surface}.}
\label{fig:app_geopotential1}
\end{figure}

\begin{figure}[p]\centering
\includegraphics[width=0.95\linewidth]{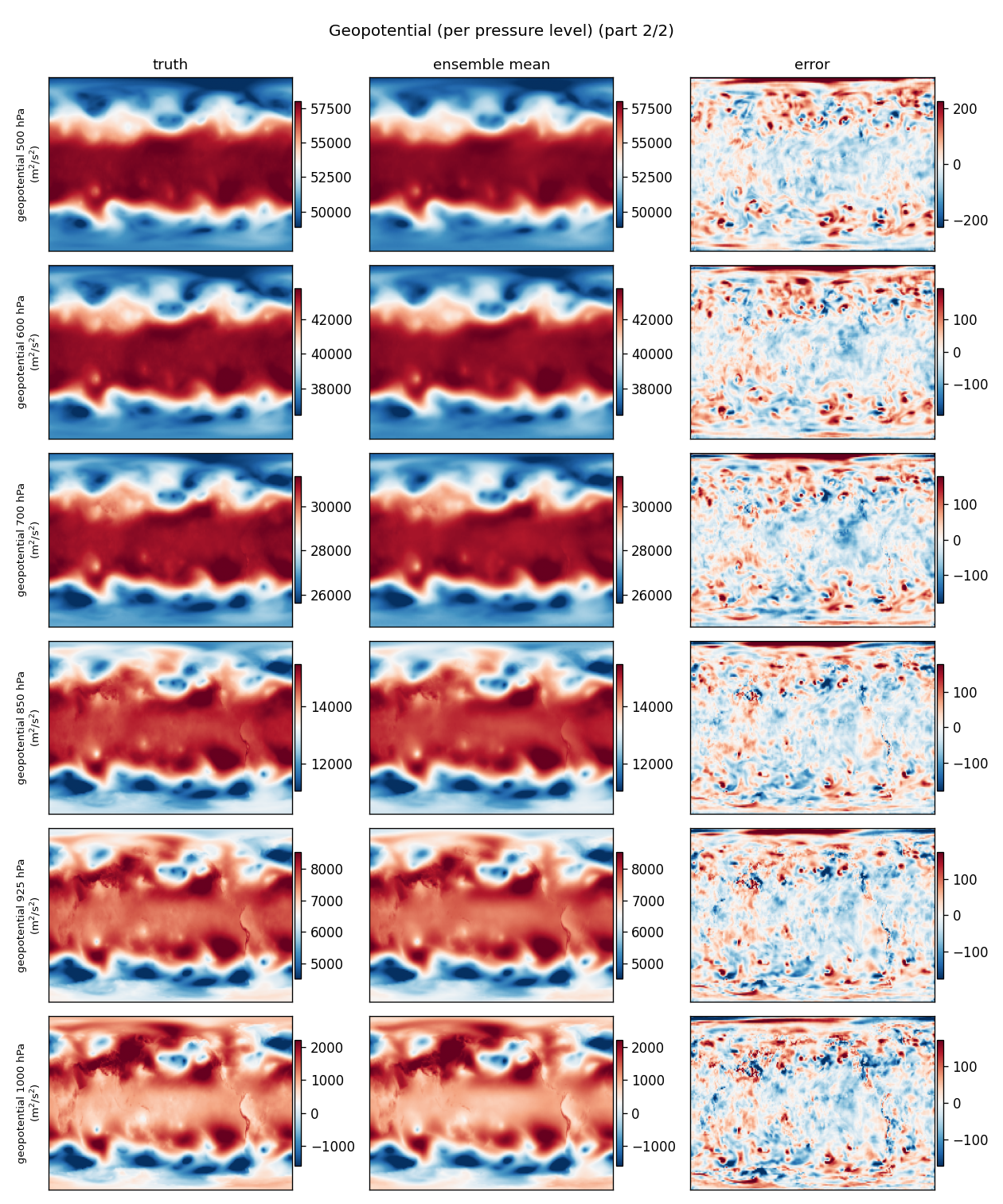}
\caption{Geopotential (per pressure level), part 2/2; same panel layout as Fig.~\ref{fig:app_surface}.}
\label{fig:app_geopotential2}
\end{figure}

\begin{figure}[p]\centering
\includegraphics[width=0.95\linewidth]{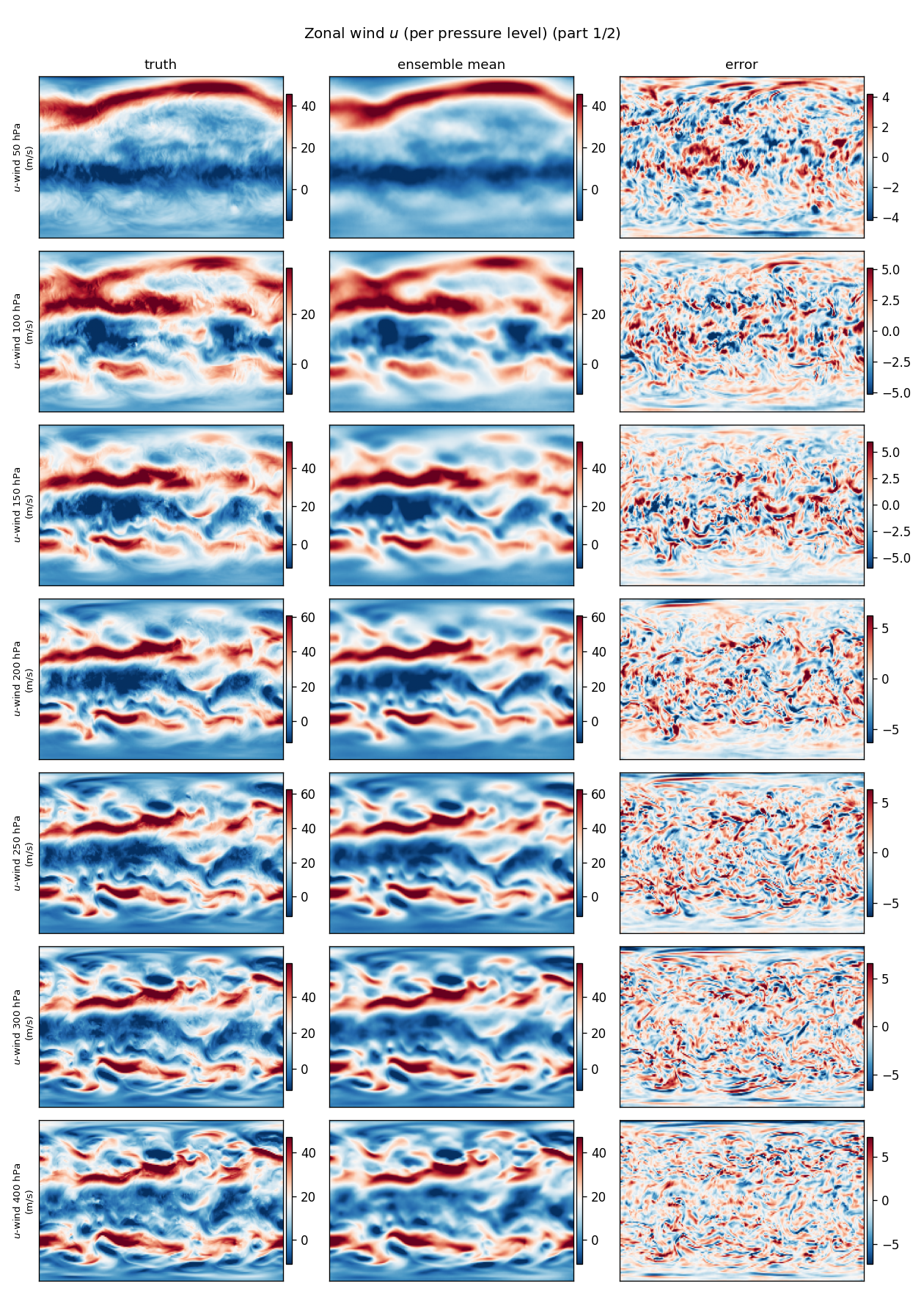}
\caption{Zonal wind $u$ (per pressure level), part 1/2; same panel layout as Fig.~\ref{fig:app_surface}.}
\label{fig:app_uwind1}
\end{figure}

\begin{figure}[p]\centering
\includegraphics[width=0.95\linewidth]{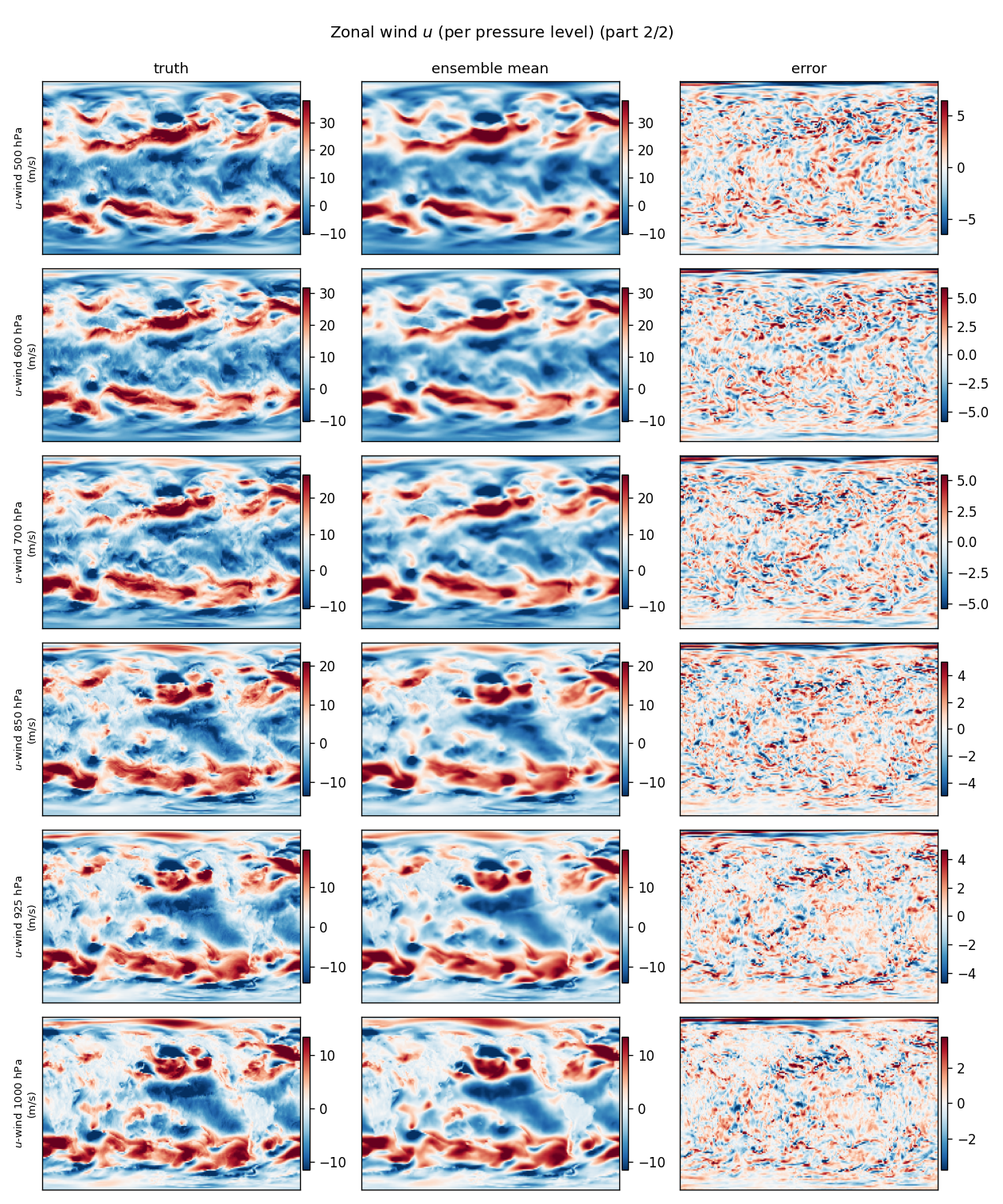}
\caption{Zonal wind $u$ (per pressure level), part 2/2; same panel layout as Fig.~\ref{fig:app_surface}.}
\label{fig:app_uwind2}
\end{figure}

\begin{figure}[p]\centering
\includegraphics[width=0.95\linewidth]{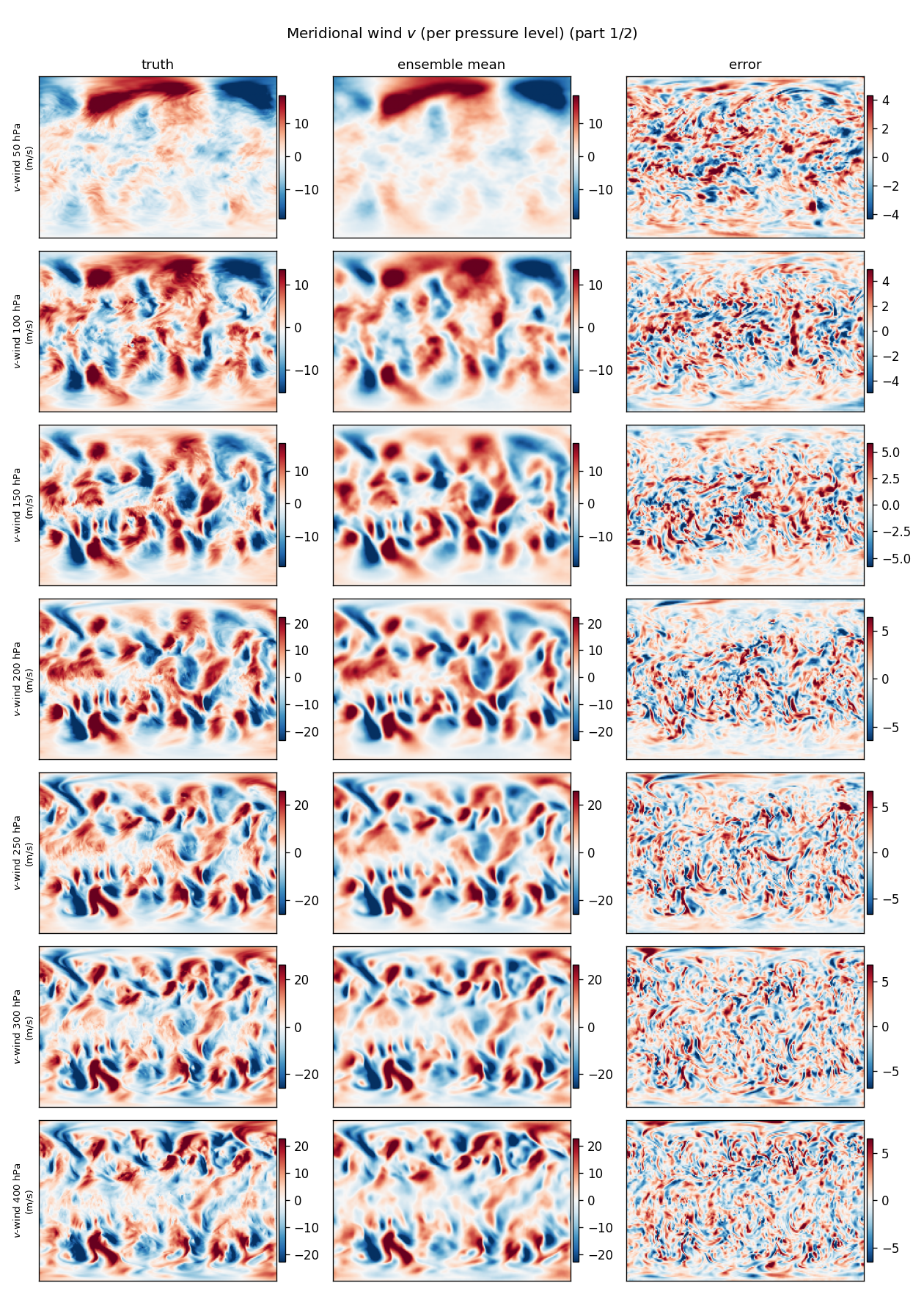}
\caption{Meridional wind $v$ (per pressure level), part 1/2; same panel layout as Fig.~\ref{fig:app_surface}.}
\label{fig:app_vwind1}
\end{figure}

\begin{figure}[p]\centering
\includegraphics[width=0.95\linewidth]{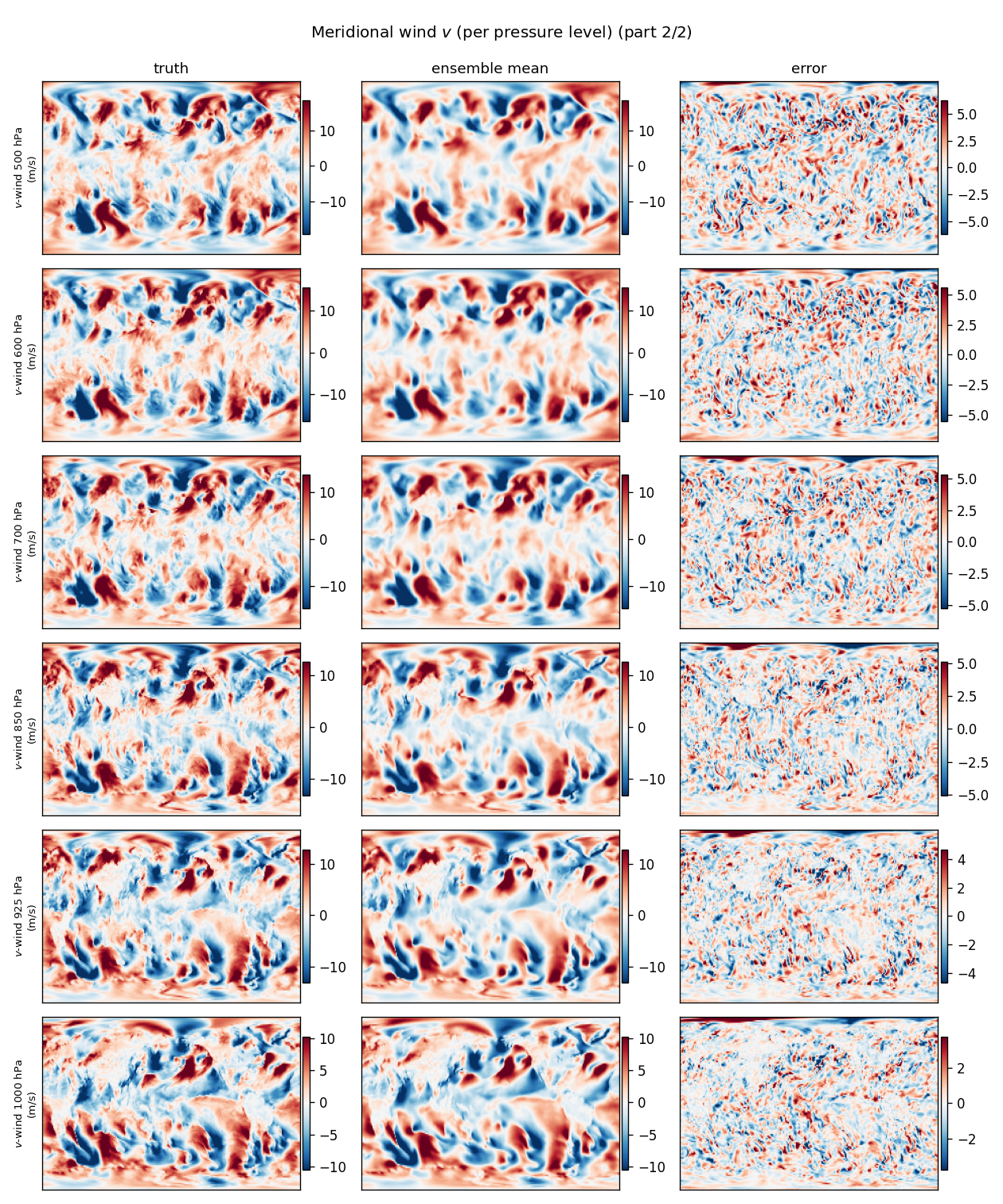}
\caption{Meridional wind $v$ (per pressure level), part 2/2; same panel layout as Fig.~\ref{fig:app_surface}.}
\label{fig:app_vwind2}
\end{figure}

\begin{figure}[p]\centering
\includegraphics[width=0.95\linewidth]{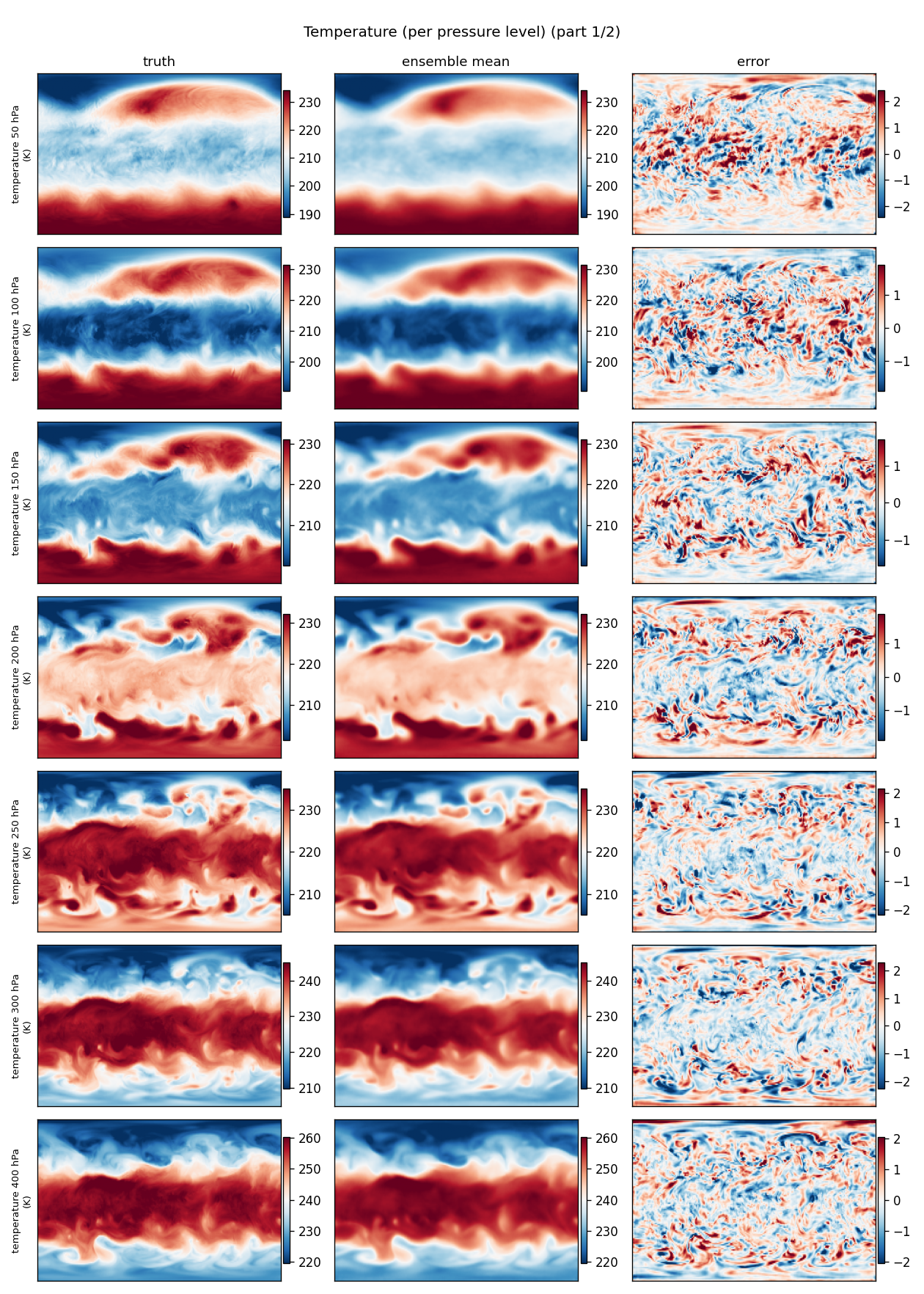}
\caption{Temperature (per pressure level), part 1/2; same panel layout as Fig.~\ref{fig:app_surface}.}
\label{fig:app_temperature1}
\end{figure}

\begin{figure}[p]\centering
\includegraphics[width=0.95\linewidth]{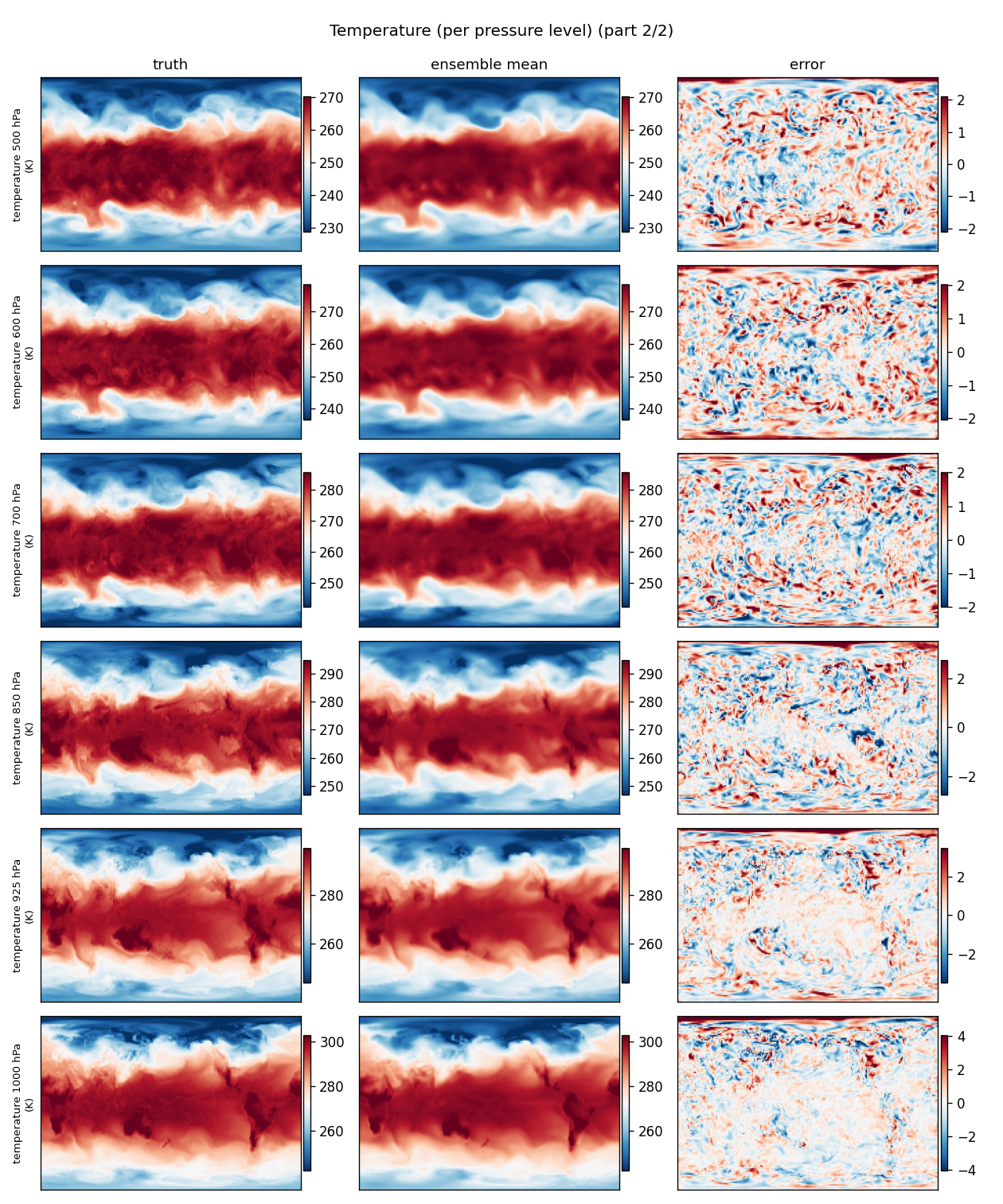}
\caption{Temperature (per pressure level), part 2/2; same panel layout as Fig.~\ref{fig:app_surface}.}
\label{fig:app_temperature2}
\end{figure}

\begin{figure}[p]\centering
\includegraphics[width=0.95\linewidth]{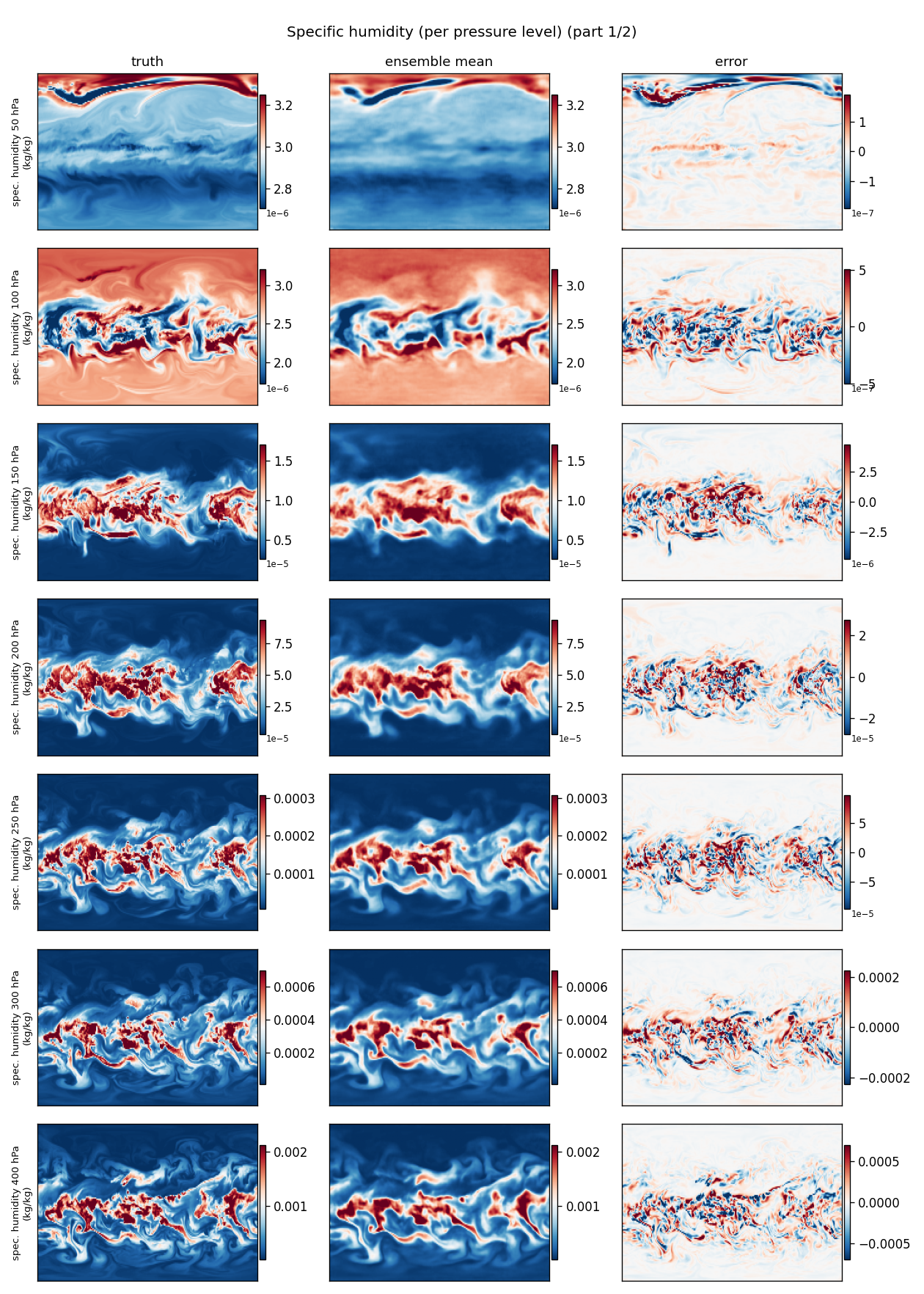}
\caption{Specific humidity (per pressure level), part 1/2; same panel layout as Fig.~\ref{fig:app_surface}.}
\label{fig:app_humidity1}
\end{figure}

\begin{figure}[p]\centering
\includegraphics[width=0.95\linewidth]{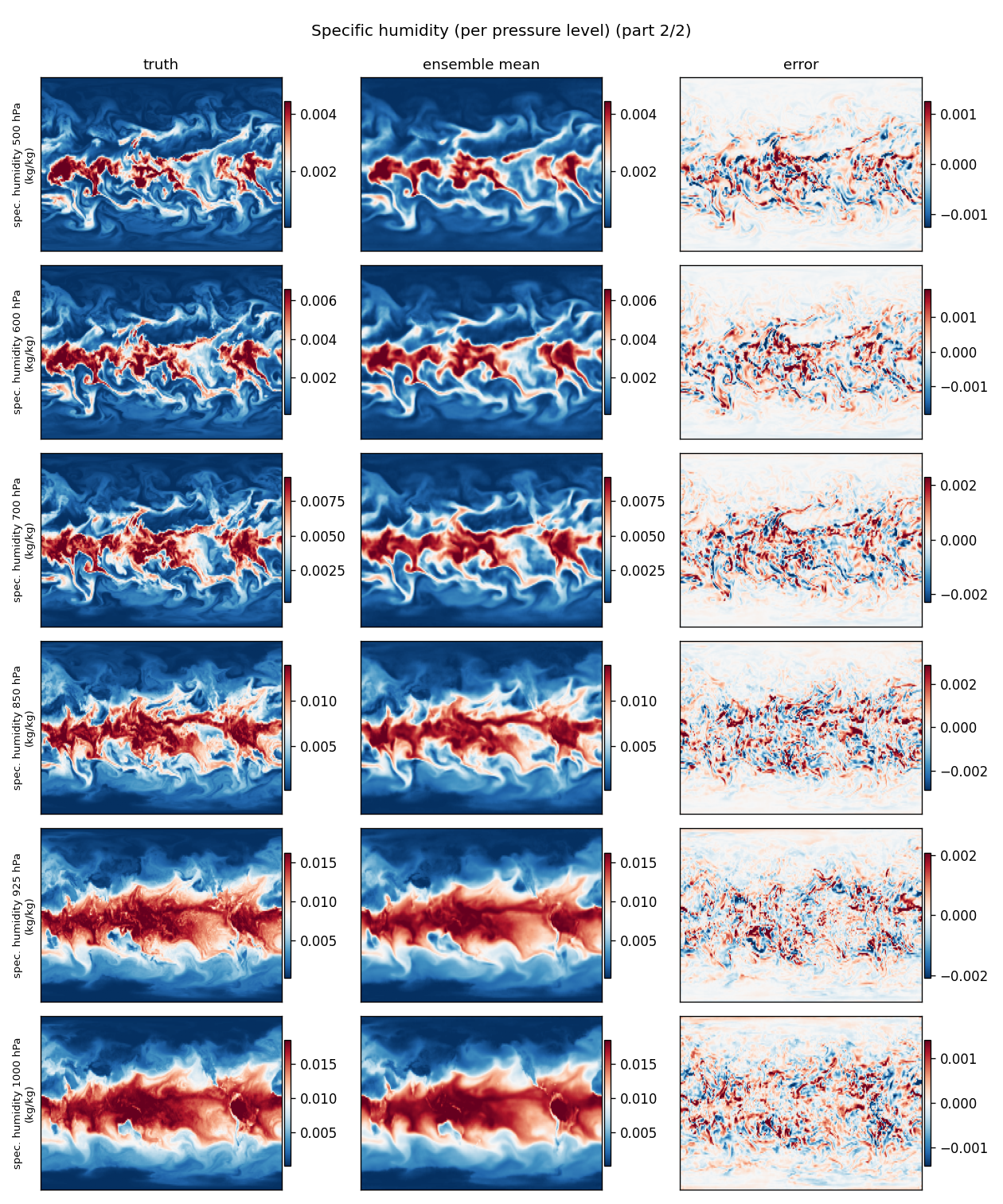}
\caption{Specific humidity (per pressure level), part 2/2; same panel layout as Fig.~\ref{fig:app_surface}.}
\label{fig:app_humidity2}
\end{figure}

The qualitative figures in the main text show a small set of the variables. This appendix shows the reconstruction of \emph{every} one of the 69 output channels: the four surface fields and the five upper-air fields (geopotential, zonal and meridional wind, temperature, and specific humidity), each at
all 13 pressure levels (50-1000\,hPa;
Figs.~\ref{fig:app_surface}-\ref{fig:app_humidity2}). Every panel shows the ERA5 truth, the
posterior ensemble mean (the same 8-member protocol as the quantitative tables),
and the signed error (mean $-$ truth)
for the super-resolution task ($8\times$ spatial subsample on every fourth
frame, \textit{DPS+corr\,(N30)} sampler) at a representative observed frame.

\subsection{Per-variable observation scatter}
\label{app:station_scatter}
The main text shows truth-vs-reconstruction scatter for three variables per task
(Figs.~\ref{fig:igra_scatter} and~\ref{fig:isd_scatter}). Here we give the record of all remaining variables, at the station locations and at off-station
grid points, pooled over the eight observed frames and eight windows. 

\begin{figure}[!ht]
\centering
\includegraphics[width=0.95\linewidth]{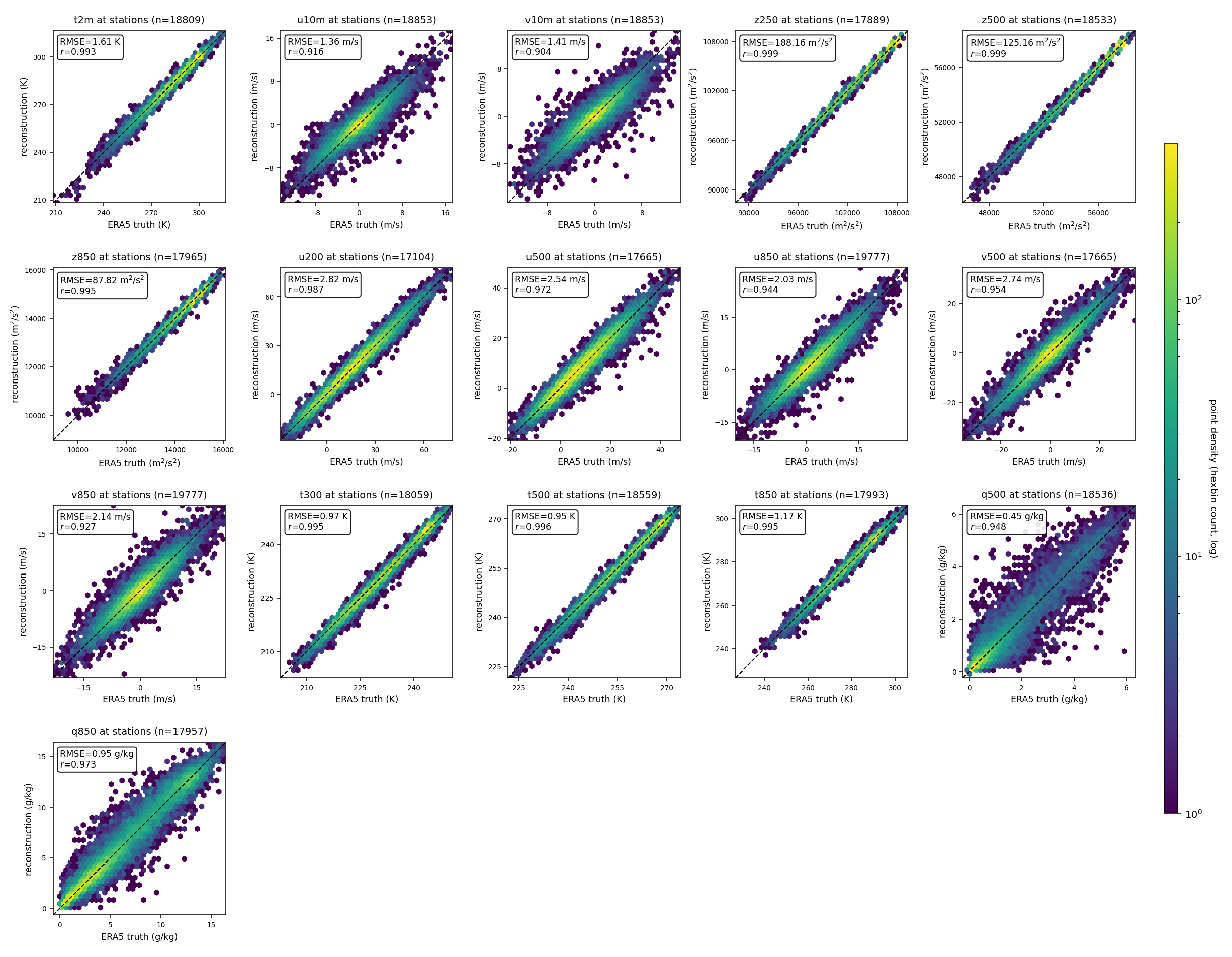}
\caption{IGRA: ERA5 truth vs.\ reconstruction at the radiosonde station locations, for a
representative subset of the 63 observed variables (the surface fields, and each upper-air
field sampled at an upper, middle and lower level).}
\label{fig:igra_app_st}
\end{figure}

\begin{figure}[!ht]
\centering
\includegraphics[width=0.95\linewidth]{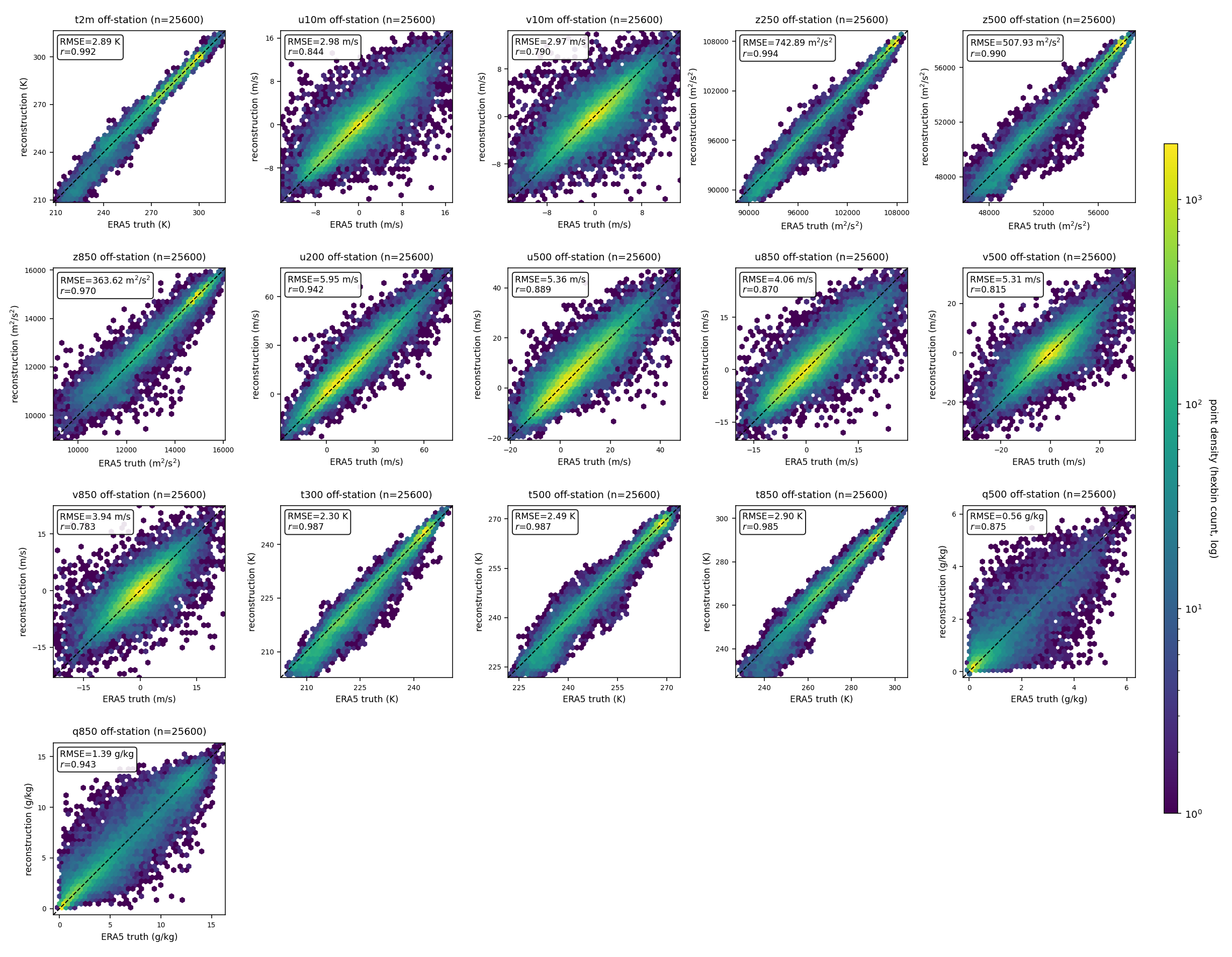}
\caption{IGRA: ERA5 truth vs.\ reconstruction at off-station grid points, for the same
representative subset of the 63 observed variables. Winds and humidity show the largest
off-station scatter.}
\label{fig:igra_app_off}
\end{figure}

\begin{figure}[!ht]
\centering
\includegraphics[width=0.9\linewidth]{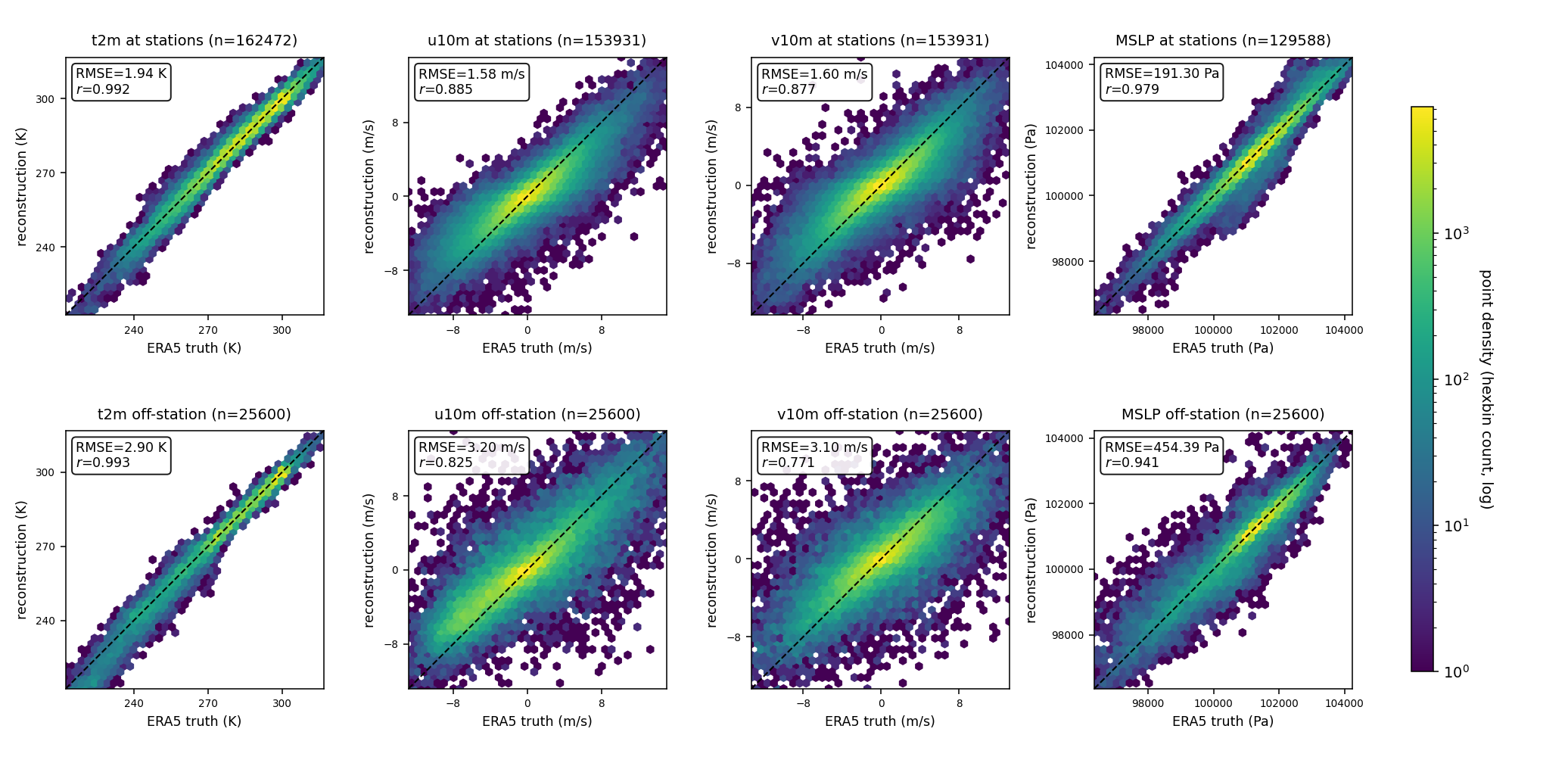}
\caption{ISD surface observations: ERA5 truth vs.\ reconstruction at stations (top) and
off-station grid points (bottom), all four observed variables.}
\label{fig:isd_app}
\end{figure}

\end{document}